\documentclass{article}

\usepackage{microtype}
\usepackage{graphicx}
\usepackage{subfigure}
\usepackage{booktabs}
\usepackage{multirow}
\usepackage{hyperref}
\usepackage{xurl}
\usepackage{amsmath}
\usepackage{amssymb}
\usepackage{enumitem}
\usepackage{tikz}
\usepackage{placeins}
\usepackage{dblfloatfix}
\usetikzlibrary{shapes,arrows,positioning,fit,backgrounds}

\usepackage[accepted]{mlsys2025}
\makeatletter
\renewcommand{\Notice@String}{}
\renewcommand{\printAffiliationsAndNotice}[1]{%
\stepcounter{@affiliationcounter}%
{\let\thefootnote\relax\footnotetext{\hspace*{-\footnotesep}\ifdefined\isaccepted #1\fi%
\forloop{@affilnum}{1}{\value{@affilnum} < \value{@affiliationcounter}}{
\textsuperscript{\arabic{@affilnum}}\ifcsname @affilname\the@affilnum\endcsname%
\csname @affilname\the@affilnum\endcsname%
\else
{\bf AUTHORERR: Missing \textbackslash{}mlsysaffiliation.}
\fi
}.
\ifdefined\mlsyscorrespondingauthor@text
Correspondence to: \mlsyscorrespondingauthor@text.
\else
{\bf AUTHORERR: Missing \textbackslash{}mlsyscorrespondingauthor.}
\fi
}
}
}
\makeatother

\mlsystitlerunning{ParetoBandit: Budget-Paced Adaptive LLM Routing}

\begin{document}


\newcommand{\bpAlpha}{0.01}
\newcommand{\bpNeff}{1163.9}
\newcommand{\bpGamma}{0.997}
\newcommand{\bpFixedLlamaReward}{0.793}
\newcommand{\bpFixedLlamaCost}{\$2.9\ensuremath{{\times}10^{-5}}}
\newcommand{\bpFixedMistralReward}{0.923}
\newcommand{\bpFixedMistralCost}{\$5.3\ensuremath{{\times}10^{-4}}}
\newcommand{\bpFixedGeminiReward}{0.932}
\newcommand{\bpFixedGeminiCost}{\$1.5\ensuremath{{\times}10^{-2}}}
\newcommand{\bpPacerTightestReward}{0.793 \pm 0.00004}
\newcommand{\bpPacerTightestRewardSE}{0.793 \pm 0.00004}
\newcommand{\bpPacerTightestCost}{\$2.9\ensuremath{{\times}10^{-5}}}
\newcommand{\bpPacerTightestUtil}{0.98\times}
\newcommand{\bpPacerTightestFinalLambda}{1.99}
\newcommand{\bpPacerTightestLambdaMedian}{2.86}
\newcommand{\bpPacerTightestRewardCILo}{0.793}
\newcommand{\bpPacerTightestRewardCIHi}{0.793}
\newcommand{\bpPacerTightestCostCILo}{0.000029}
\newcommand{\bpPacerTightestCostCIHi}{0.000029}
\newcommand{\bpPacerTightestUtilCILo}{0.98}
\newcommand{\bpPacerTightestUtilCIHi}{0.99}
\newcommand{\bpPacerSecondReward}{0.813 \pm 0.0005}
\newcommand{\bpPacerSecondRewardSE}{0.813 \pm 0.0005}
\newcommand{\bpPacerSecondCost}{\$8.3\ensuremath{{\times}10^{-5}}}
\newcommand{\bpPacerSecondUtil}{1.00\times}
\newcommand{\bpPacerSecondFinalLambda}{0.68}
\newcommand{\bpPacerSecondLambdaMedian}{0.64}
\newcommand{\bpPacerSecondRewardCILo}{0.812}
\newcommand{\bpPacerSecondRewardCIHi}{0.814}
\newcommand{\bpPacerSecondCostCILo}{0.000083}
\newcommand{\bpPacerSecondCostCIHi}{0.000084}
\newcommand{\bpPacerSecondUtilCILo}{1.00}
\newcommand{\bpPacerSecondUtilCIHi}{1.02}
\newcommand{\bpPacerThirdReward}{0.859 \pm 0.0008}
\newcommand{\bpPacerThirdRewardSE}{0.859 \pm 0.0008}
\newcommand{\bpPacerThirdCost}{\$2.3\ensuremath{{\times}10^{-4}}}
\newcommand{\bpPacerThirdUtil}{1.00\times}
\newcommand{\bpPacerThirdFinalLambda}{0.53}
\newcommand{\bpPacerThirdLambdaMedian}{0.46}
\newcommand{\bpPacerThirdRewardCILo}{0.857}
\newcommand{\bpPacerThirdRewardCIHi}{0.860}
\newcommand{\bpPacerThirdCostCILo}{0.000234}
\newcommand{\bpPacerThirdCostCIHi}{0.000236}
\newcommand{\bpPacerThirdUtilCILo}{1.00}
\newcommand{\bpPacerThirdUtilCIHi}{1.00}
\newcommand{\bpPacerFourthReward}{0.898 \pm 0.002}
\newcommand{\bpPacerFourthRewardSE}{0.898 \pm 0.002}
\newcommand{\bpPacerFourthCost}{\$6.5\ensuremath{{\times}10^{-4}}}
\newcommand{\bpPacerFourthUtil}{0.98\times}
\newcommand{\bpPacerFourthFinalLambda}{0.29}
\newcommand{\bpPacerFourthLambdaMedian}{0.17}
\newcommand{\bpPacerFourthRewardCILo}{0.894}
\newcommand{\bpPacerFourthRewardCIHi}{0.901}
\newcommand{\bpPacerFourthCostCILo}{0.000647}
\newcommand{\bpPacerFourthCostCIHi}{0.000653}
\newcommand{\bpPacerFourthUtilCILo}{0.98}
\newcommand{\bpPacerFourthUtilCIHi}{0.98}
\newcommand{\bpPacerFifthReward}{0.917 \pm 0.0008}
\newcommand{\bpPacerFifthRewardSE}{0.917 \pm 0.0008}
\newcommand{\bpPacerFifthCost}{\$1.7\ensuremath{{\times}10^{-3}}}
\newcommand{\bpPacerFifthUtil}{0.91\times}
\newcommand{\bpPacerFifthFinalLambda}{0.01}
\newcommand{\bpPacerFifthLambdaMedian}{0.01}
\newcommand{\bpPacerFifthRewardCILo}{0.916}
\newcommand{\bpPacerFifthRewardCIHi}{0.919}
\newcommand{\bpPacerFifthCostCILo}{0.001685}
\newcommand{\bpPacerFifthCostCIHi}{0.001730}
\newcommand{\bpPacerFifthUtilCILo}{0.90}
\newcommand{\bpPacerFifthUtilCIHi}{0.92}
\newcommand{\bpPacerSixthReward}{0.924 \pm 0.0003}
\newcommand{\bpPacerSixthRewardSE}{0.924 \pm 0.0003}
\newcommand{\bpPacerSixthCost}{\$4.7\ensuremath{{\times}10^{-3}}}
\newcommand{\bpPacerSixthUtil}{0.89\times}
\newcommand{\bpPacerSixthFinalLambda}{0.01}
\newcommand{\bpPacerSixthLambdaMedian}{0.00}
\newcommand{\bpPacerSixthRewardCILo}{0.924}
\newcommand{\bpPacerSixthRewardCIHi}{0.925}
\newcommand{\bpPacerSixthCostCILo}{0.004610}
\newcommand{\bpPacerSixthCostCIHi}{0.004786}
\newcommand{\bpPacerSixthUtilCILo}{0.87}
\newcommand{\bpPacerSixthUtilCIHi}{0.90}
\newcommand{\bpPacerLoosestReward}{0.929 \pm 0.0003}
\newcommand{\bpPacerLoosestRewardSE}{0.929 \pm 0.0003}
\newcommand{\bpPacerLoosestCost}{\$1.1\ensuremath{{\times}10^{-2}}}
\newcommand{\bpPacerLoosestUtil}{0.74\times}
\newcommand{\bpPacerLoosestFinalLambda}{0.00}
\newcommand{\bpPacerLoosestLambdaMedian}{0.00}
\newcommand{\bpPacerLoosestRewardCILo}{0.928}
\newcommand{\bpPacerLoosestRewardCIHi}{0.930}
\newcommand{\bpPacerLoosestCostCILo}{0.010780}
\newcommand{\bpPacerLoosestCostCIHi}{0.011354}
\newcommand{\bpPacerLoosestUtilCILo}{0.72}
\newcommand{\bpPacerLoosestUtilCIHi}{0.76}
\newcommand{\bpCheapestModelCost}{\$2.9\ensuremath{{\times}10^{-5}}}
\newcommand{\bpExpensiveModelCost}{\$1.5\ensuremath{{\times}10^{-2}}}
\newcommand{\bpPriceRangeX}{530}
\newcommand{\bpCeilingOraclePct}{96.4}
\newcommand{\bpCeilingOraclePctCILo}{96.4}
\newcommand{\bpCeilingOraclePctCIHi}{96.5}
\newcommand{\bpOracleReward}{0.963}
\newcommand{\bpOracleCost}{\$5.0\ensuremath{{\times}10^{-3}}}
\newcommand{\bpUtilBindingLow}{0.98\times}
\newcommand{\bpUtilBindingHigh}{1.00\times}
\newcommand{\bpLoosestUtil}{0.74\times}
\newcommand{\bpAnnotQualPct}{92}
\newcommand{\bpAnnotCostPct}{2}
\newcommand{\bpAnnotCost}{\$2.3\ensuremath{{\times}10^{-4}}}
\newcommand{\bpAnnotBudget}{\$2.3\ensuremath{{\times}10^{-4}}}
\newcommand{\bpAnnotSavingPctGemini}{98}
\newcommand{\bpAnnotQualPctCILo}{92}
\newcommand{\bpAnnotQualPctCIHi}{92}
\newcommand{\bpAnnotSavingPctGeminiCILo}{98}
\newcommand{\bpAnnotSavingPctGeminiCIHi}{98}
\newcommand{\bpAnnotLlamaFrac}{56}
\newcommand{\bpAnnotMistralFrac}{44}
\newcommand{\bpAnnotGeminiFrac}{0.0}
\newcommand{\bpPacerTightestQualPctGemini}{85.1}
\newcommand{\bpPacerTightestSavingPctGemini}{99.8}
\newcommand{\bpPacerTightestQualPctGeminiCILo}{85.1}
\newcommand{\bpPacerTightestQualPctGeminiCIHi}{85.1}
\newcommand{\bpPacerTightestSavingPctGeminiCILo}{99.8}
\newcommand{\bpPacerTightestSavingPctGeminiCIHi}{99.8}
\newcommand{\bpPacerSecondQualPctGemini}{87.2}
\newcommand{\bpPacerSecondSavingPctGemini}{99.5}
\newcommand{\bpPacerSecondQualPctGeminiCILo}{87.1}
\newcommand{\bpPacerSecondQualPctGeminiCIHi}{87.3}
\newcommand{\bpPacerSecondSavingPctGeminiCILo}{99.4}
\newcommand{\bpPacerSecondSavingPctGeminiCIHi}{99.5}
\newcommand{\bpPacerThirdQualPctGemini}{92.1}
\newcommand{\bpPacerThirdSavingPctGemini}{98.5}
\newcommand{\bpPacerThirdQualPctGeminiCILo}{92.0}
\newcommand{\bpPacerThirdQualPctGeminiCIHi}{92.3}
\newcommand{\bpPacerThirdSavingPctGeminiCILo}{98.4}
\newcommand{\bpPacerThirdSavingPctGeminiCIHi}{98.5}
\newcommand{\bpPacerFourthQualPctGemini}{96.3}
\newcommand{\bpPacerFourthSavingPctGemini}{95.7}
\newcommand{\bpPacerFourthQualPctGeminiCILo}{95.9}
\newcommand{\bpPacerFourthQualPctGeminiCIHi}{96.6}
\newcommand{\bpPacerFourthSavingPctGeminiCILo}{95.7}
\newcommand{\bpPacerFourthSavingPctGeminiCIHi}{95.7}
\newcommand{\bpPacerFifthQualPctGemini}{98.4}
\newcommand{\bpPacerFifthSavingPctGemini}{88.7}
\newcommand{\bpPacerFifthQualPctGeminiCILo}{98.2}
\newcommand{\bpPacerFifthQualPctGeminiCIHi}{98.5}
\newcommand{\bpPacerFifthSavingPctGeminiCILo}{88.6}
\newcommand{\bpPacerFifthSavingPctGeminiCIHi}{88.9}
\newcommand{\bpPacerSixthQualPctGemini}{99.2}
\newcommand{\bpPacerSixthSavingPctGemini}{69.0}
\newcommand{\bpPacerSixthQualPctGeminiCILo}{99.1}
\newcommand{\bpPacerSixthQualPctGeminiCIHi}{99.2}
\newcommand{\bpPacerSixthSavingPctGeminiCILo}{68.4}
\newcommand{\bpPacerSixthSavingPctGeminiCIHi}{69.6}
\newcommand{\bpPacerLoosestQualPctGemini}{99.7}
\newcommand{\bpPacerLoosestSavingPctGemini}{26.8}
\newcommand{\bpPacerLoosestQualPctGeminiCILo}{99.6}
\newcommand{\bpPacerLoosestQualPctGeminiCIHi}{99.7}
\newcommand{\bpPacerLoosestSavingPctGeminiCILo}{25.1}
\newcommand{\bpPacerLoosestSavingPctGeminiCIHi}{28.9}


\newcommand{\bdNeff}{1164}
\newcommand{\bdAlpha}{0.01}
\newcommand{\bdGamma}{0.997}
\newcommand{\bdNaiveTightPhaseOneRatio}{1.07\times}
\newcommand{\bdNaiveTightPhaseOneCost}{\$3.2e-04}
\newcommand{\bdNaiveTightPhaseOneCostSE}{\$1.2e-06}
\newcommand{\bdNaiveTightPhaseOneRatioSE}{0.00}
\newcommand{\bdNaiveTightPhaseOneRatioCILo}{1.06}
\newcommand{\bdNaiveTightPhaseOneRatioCIHi}{1.07}
\newcommand{\bdNaiveTightPhaseTwoRatio}{0.89\times}
\newcommand{\bdNaiveTightPhaseTwoCost}{\$2.7e-04}
\newcommand{\bdNaiveTightPhaseTwoCostSE}{\$3.6e-08}
\newcommand{\bdNaiveTightPhaseTwoRatioSE}{0.00}
\newcommand{\bdNaiveTightPhaseTwoRatioCILo}{0.89}
\newcommand{\bdNaiveTightPhaseTwoRatioCIHi}{0.89}
\newcommand{\bdNaiveTightPhaseThreeRatio}{1.09\times}
\newcommand{\bdNaiveTightPhaseThreeCost}{\$3.3e-04}
\newcommand{\bdNaiveTightPhaseThreeCostSE}{\$1.2e-06}
\newcommand{\bdNaiveTightPhaseThreeRatioSE}{0.00}
\newcommand{\bdNaiveTightPhaseThreeRatioCILo}{1.08}
\newcommand{\bdNaiveTightPhaseThreeRatioCIHi}{1.10}
\newcommand{\bdRecalTightPhaseOneRatio}{1.07\times}
\newcommand{\bdRecalTightPhaseOneCost}{\$3.2e-04}
\newcommand{\bdRecalTightPhaseOneCostSE}{\$1.2e-06}
\newcommand{\bdRecalTightPhaseOneRatioSE}{0.00}
\newcommand{\bdRecalTightPhaseOneRatioCILo}{1.06}
\newcommand{\bdRecalTightPhaseOneRatioCIHi}{1.07}
\newcommand{\bdRecalTightPhaseTwoRatio}{0.99\times}
\newcommand{\bdRecalTightPhaseTwoCost}{\$3.0e-04}
\newcommand{\bdRecalTightPhaseTwoCostSE}{\$8.5e-07}
\newcommand{\bdRecalTightPhaseTwoRatioSE}{0.00}
\newcommand{\bdRecalTightPhaseTwoRatioCILo}{0.99}
\newcommand{\bdRecalTightPhaseTwoRatioCIHi}{1.00}
\newcommand{\bdRecalTightPhaseThreeRatio}{1.44\times}
\newcommand{\bdRecalTightPhaseThreeCost}{\$4.3e-04}
\newcommand{\bdRecalTightPhaseThreeCostSE}{\$5.7e-06}
\newcommand{\bdRecalTightPhaseThreeRatioSE}{0.02}
\newcommand{\bdRecalTightPhaseThreeRatioCILo}{1.40}
\newcommand{\bdRecalTightPhaseThreeRatioCIHi}{1.47}
\newcommand{\bdForgetTightPhaseOneRatio}{2.62\times}
\newcommand{\bdForgetTightPhaseOneCost}{\$7.9e-04}
\newcommand{\bdForgetTightPhaseOneCostSE}{\$4.9e-05}
\newcommand{\bdForgetTightPhaseOneRatioSE}{0.16}
\newcommand{\bdForgetTightPhaseOneRatioCILo}{2.34}
\newcommand{\bdForgetTightPhaseOneRatioCIHi}{2.97}
\newcommand{\bdForgetTightPhaseTwoRatio}{0.88\times}
\newcommand{\bdForgetTightPhaseTwoCost}{\$2.7e-04}
\newcommand{\bdForgetTightPhaseTwoCostSE}{\$1.6e-06}
\newcommand{\bdForgetTightPhaseTwoRatioSE}{0.01}
\newcommand{\bdForgetTightPhaseTwoRatioCILo}{0.87}
\newcommand{\bdForgetTightPhaseTwoRatioCIHi}{0.89}
\newcommand{\bdForgetTightPhaseThreeRatio}{5.50\times}
\newcommand{\bdForgetTightPhaseThreeCost}{\$1.7e-03}
\newcommand{\bdForgetTightPhaseThreeCostSE}{\$8.4e-05}
\newcommand{\bdForgetTightPhaseThreeRatioSE}{0.28}
\newcommand{\bdForgetTightPhaseThreeRatioCILo}{4.99}
\newcommand{\bdForgetTightPhaseThreeRatioCIHi}{6.05}
\newcommand{\bdParetoBanditTightPhaseOneRatio}{1.01\times}
\newcommand{\bdParetoBanditTightPhaseOneCost}{\$3.0e-04}
\newcommand{\bdParetoBanditTightPhaseOneCostSE}{\$9.1e-07}
\newcommand{\bdParetoBanditTightPhaseOneRatioSE}{0.00}
\newcommand{\bdParetoBanditTightPhaseOneRatioCILo}{1.00}
\newcommand{\bdParetoBanditTightPhaseOneRatioCIHi}{1.01}
\newcommand{\bdParetoBanditTightPhaseTwoRatio}{0.95\times}
\newcommand{\bdParetoBanditTightPhaseTwoCost}{\$2.8e-04}
\newcommand{\bdParetoBanditTightPhaseTwoCostSE}{\$1.1e-06}
\newcommand{\bdParetoBanditTightPhaseTwoRatioSE}{0.00}
\newcommand{\bdParetoBanditTightPhaseTwoRatioCILo}{0.94}
\newcommand{\bdParetoBanditTightPhaseTwoRatioCIHi}{0.95}
\newcommand{\bdParetoBanditTightPhaseThreeRatio}{1.04\times}
\newcommand{\bdParetoBanditTightPhaseThreeCost}{\$3.1e-04}
\newcommand{\bdParetoBanditTightPhaseThreeCostSE}{\$7.4e-06}
\newcommand{\bdParetoBanditTightPhaseThreeRatioSE}{0.02}
\newcommand{\bdParetoBanditTightPhaseThreeRatioCILo}{1.01}
\newcommand{\bdParetoBanditTightPhaseThreeRatioCIHi}{1.09}
\newcommand{\bdParetoBanditTightLambdaPhaseOne}{0.44}
\newcommand{\bdParetoBanditTightGeminiPhaseOne}{0}
\newcommand{\bdParetoBanditTightRewardPhaseOne}{0.8621}
\newcommand{\bdParetoBanditTightRewardPhaseOneSE}{0.0010}
\newcommand{\bdParetoBanditTightRewardPhaseOneCILo}{0.8604}
\newcommand{\bdParetoBanditTightRewardPhaseOneCIHi}{0.8640}
\newcommand{\bdParetoBanditTightLambdaPhaseTwo}{0.08}
\newcommand{\bdParetoBanditTightGeminiPhaseTwo}{81}
\newcommand{\bdParetoBanditTightRewardPhaseTwo}{0.9334}
\newcommand{\bdParetoBanditTightRewardPhaseTwoSE}{0.0011}
\newcommand{\bdParetoBanditTightRewardPhaseTwoCILo}{0.9313}
\newcommand{\bdParetoBanditTightRewardPhaseTwoCIHi}{0.9353}
\newcommand{\bdParetoBanditTightLambdaPhaseThree}{0.59}
\newcommand{\bdParetoBanditTightGeminiPhaseThree}{0}
\newcommand{\bdParetoBanditTightRewardPhaseThree}{0.8543}
\newcommand{\bdParetoBanditTightRewardPhaseThreeSE}{0.0024}
\newcommand{\bdParetoBanditTightRewardPhaseThreeCILo}{0.8492}
\newcommand{\bdParetoBanditTightRewardPhaseThreeCIHi}{0.8581}
\newcommand{\bdParetoBanditTightRewardLift}{0.071}
\newcommand{\bdParetoBanditTightRewardLiftSE}{0.0014}
\newcommand{\bdParetoBanditTightRewardLiftCILo}{0.068}
\newcommand{\bdParetoBanditTightRewardLiftCIHi}{0.074}
\newcommand{\bdParetoBanditTightPhaseOneUtil}{1.01\times}
\newcommand{\bdNaiveModPhaseOneRatio}{0.57\times}
\newcommand{\bdNaiveModPhaseOneCost}{\$3.8e-04}
\newcommand{\bdNaiveModPhaseOneCostSE}{\$6.8e-07}
\newcommand{\bdNaiveModPhaseOneRatioSE}{0.00}
\newcommand{\bdNaiveModPhaseOneRatioCILo}{0.57}
\newcommand{\bdNaiveModPhaseOneRatioCIHi}{0.57}
\newcommand{\bdNaiveModPhaseTwoRatio}{0.40\times}
\newcommand{\bdNaiveModPhaseTwoCost}{\$2.7e-04}
\newcommand{\bdNaiveModPhaseTwoCostSE}{\$3.0e-08}
\newcommand{\bdNaiveModPhaseTwoRatioSE}{0.00}
\newcommand{\bdNaiveModPhaseTwoRatioCILo}{0.40}
\newcommand{\bdNaiveModPhaseTwoRatioCIHi}{0.40}
\newcommand{\bdNaiveModPhaseThreeRatio}{0.61\times}
\newcommand{\bdNaiveModPhaseThreeCost}{\$4.0e-04}
\newcommand{\bdNaiveModPhaseThreeCostSE}{\$4.7e-06}
\newcommand{\bdNaiveModPhaseThreeRatioSE}{0.01}
\newcommand{\bdNaiveModPhaseThreeRatioCILo}{0.59}
\newcommand{\bdNaiveModPhaseThreeRatioCIHi}{0.62}
\newcommand{\bdRecalModPhaseOneRatio}{0.57\times}
\newcommand{\bdRecalModPhaseOneCost}{\$3.8e-04}
\newcommand{\bdRecalModPhaseOneCostSE}{\$6.8e-07}
\newcommand{\bdRecalModPhaseOneRatioSE}{0.00}
\newcommand{\bdRecalModPhaseOneRatioCILo}{0.57}
\newcommand{\bdRecalModPhaseOneRatioCIHi}{0.57}
\newcommand{\bdRecalModPhaseTwoRatio}{0.44\times}
\newcommand{\bdRecalModPhaseTwoCost}{\$2.9e-04}
\newcommand{\bdRecalModPhaseTwoCostSE}{\$6.4e-07}
\newcommand{\bdRecalModPhaseTwoRatioSE}{0.00}
\newcommand{\bdRecalModPhaseTwoRatioCILo}{0.44}
\newcommand{\bdRecalModPhaseTwoRatioCIHi}{0.44}
\newcommand{\bdRecalModPhaseThreeRatio}{2.14\times}
\newcommand{\bdRecalModPhaseThreeCost}{\$1.4e-03}
\newcommand{\bdRecalModPhaseThreeCostSE}{\$2.0e-05}
\newcommand{\bdRecalModPhaseThreeRatioSE}{0.03}
\newcommand{\bdRecalModPhaseThreeRatioCILo}{2.08}
\newcommand{\bdRecalModPhaseThreeRatioCIHi}{2.20}
\newcommand{\bdForgetModPhaseOneRatio}{1.50\times}
\newcommand{\bdForgetModPhaseOneCost}{\$1.0e-03}
\newcommand{\bdForgetModPhaseOneCostSE}{\$4.7e-05}
\newcommand{\bdForgetModPhaseOneRatioSE}{0.07}
\newcommand{\bdForgetModPhaseOneRatioCILo}{1.39}
\newcommand{\bdForgetModPhaseOneRatioCIHi}{1.65}
\newcommand{\bdForgetModPhaseTwoRatio}{0.43\times}
\newcommand{\bdForgetModPhaseTwoCost}{\$2.9e-04}
\newcommand{\bdForgetModPhaseTwoCostSE}{\$4.6e-06}
\newcommand{\bdForgetModPhaseTwoRatioSE}{0.01}
\newcommand{\bdForgetModPhaseTwoRatioCILo}{0.42}
\newcommand{\bdForgetModPhaseTwoRatioCIHi}{0.45}
\newcommand{\bdForgetModPhaseThreeRatio}{4.13\times}
\newcommand{\bdForgetModPhaseThreeCost}{\$2.7e-03}
\newcommand{\bdForgetModPhaseThreeCostSE}{\$1.4e-04}
\newcommand{\bdForgetModPhaseThreeRatioSE}{0.21}
\newcommand{\bdForgetModPhaseThreeRatioCILo}{3.74}
\newcommand{\bdForgetModPhaseThreeRatioCIHi}{4.54}
\newcommand{\bdParetoBanditModPhaseOneRatio}{0.99\times}
\newcommand{\bdParetoBanditModPhaseOneCost}{\$6.5e-04}
\newcommand{\bdParetoBanditModPhaseOneCostSE}{\$4.2e-06}
\newcommand{\bdParetoBanditModPhaseOneRatioSE}{0.01}
\newcommand{\bdParetoBanditModPhaseOneRatioCILo}{0.98}
\newcommand{\bdParetoBanditModPhaseOneRatioCIHi}{1.00}
\newcommand{\bdParetoBanditModPhaseTwoRatio}{0.65\times}
\newcommand{\bdParetoBanditModPhaseTwoCost}{\$4.3e-04}
\newcommand{\bdParetoBanditModPhaseTwoCostSE}{\$6.6e-06}
\newcommand{\bdParetoBanditModPhaseTwoRatioSE}{0.01}
\newcommand{\bdParetoBanditModPhaseTwoRatioCILo}{0.63}
\newcommand{\bdParetoBanditModPhaseTwoRatioCIHi}{0.67}
\newcommand{\bdParetoBanditModPhaseThreeRatio}{1.00\times}
\newcommand{\bdParetoBanditModPhaseThreeCost}{\$6.6e-04}
\newcommand{\bdParetoBanditModPhaseThreeCostSE}{\$2.2e-06}
\newcommand{\bdParetoBanditModPhaseThreeRatioSE}{0.00}
\newcommand{\bdParetoBanditModPhaseThreeRatioCILo}{1.00}
\newcommand{\bdParetoBanditModPhaseThreeRatioCIHi}{1.01}
\newcommand{\bdParetoBanditModLambdaPhaseOne}{0.34}
\newcommand{\bdParetoBanditModGeminiPhaseOne}{2}
\newcommand{\bdParetoBanditModRewardPhaseOne}{0.8823}
\newcommand{\bdParetoBanditModRewardPhaseOneSE}{0.0025}
\newcommand{\bdParetoBanditModRewardPhaseOneCILo}{0.8774}
\newcommand{\bdParetoBanditModRewardPhaseOneCIHi}{0.8869}
\newcommand{\bdParetoBanditModLambdaPhaseTwo}{0.01}
\newcommand{\bdParetoBanditModGeminiPhaseTwo}{28}
\newcommand{\bdParetoBanditModRewardPhaseTwo}{0.9259}
\newcommand{\bdParetoBanditModRewardPhaseTwoSE}{0.0007}
\newcommand{\bdParetoBanditModRewardPhaseTwoCILo}{0.9247}
\newcommand{\bdParetoBanditModRewardPhaseTwoCIHi}{0.9272}
\newcommand{\bdParetoBanditModLambdaPhaseThree}{0.30}
\newcommand{\bdParetoBanditModGeminiPhaseThree}{2}
\newcommand{\bdParetoBanditModRewardPhaseThree}{0.8914}
\newcommand{\bdParetoBanditModRewardPhaseThreeSE}{0.0027}
\newcommand{\bdParetoBanditModRewardPhaseThreeCILo}{0.8860}
\newcommand{\bdParetoBanditModRewardPhaseThreeCIHi}{0.8964}
\newcommand{\bdParetoBanditModRewardLift}{0.044}
\newcommand{\bdParetoBanditModRewardLiftSE}{0.0027}
\newcommand{\bdParetoBanditModRewardLiftCILo}{0.039}
\newcommand{\bdParetoBanditModRewardLiftCIHi}{0.049}
\newcommand{\bdParetoBanditModPhaseOneUtil}{0.99\times}
\newcommand{\bdNaiveLoosePhaseOneRatio}{3.07\times}
\newcommand{\bdNaiveLoosePhaseOneCost}{\$5.7e-03}
\newcommand{\bdNaiveLoosePhaseOneCostSE}{\$7.2e-05}
\newcommand{\bdNaiveLoosePhaseOneRatioSE}{0.04}
\newcommand{\bdNaiveLoosePhaseOneRatioCILo}{3.00}
\newcommand{\bdNaiveLoosePhaseOneRatioCIHi}{3.15}
\newcommand{\bdNaiveLoosePhaseTwoRatio}{0.14\times}
\newcommand{\bdNaiveLoosePhaseTwoCost}{\$2.7e-04}
\newcommand{\bdNaiveLoosePhaseTwoCostSE}{\$3.5e-08}
\newcommand{\bdNaiveLoosePhaseTwoRatioSE}{0.00}
\newcommand{\bdNaiveLoosePhaseTwoRatioCILo}{0.14}
\newcommand{\bdNaiveLoosePhaseTwoRatioCIHi}{0.14}
\newcommand{\bdNaiveLoosePhaseThreeRatio}{3.26\times}
\newcommand{\bdNaiveLoosePhaseThreeCost}{\$6.1e-03}
\newcommand{\bdNaiveLoosePhaseThreeCostSE}{\$8.4e-05}
\newcommand{\bdNaiveLoosePhaseThreeRatioSE}{0.05}
\newcommand{\bdNaiveLoosePhaseThreeRatioCILo}{3.18}
\newcommand{\bdNaiveLoosePhaseThreeRatioCIHi}{3.35}
\newcommand{\bdRecalLoosePhaseOneRatio}{3.07\times}
\newcommand{\bdRecalLoosePhaseOneCost}{\$5.7e-03}
\newcommand{\bdRecalLoosePhaseOneCostSE}{\$7.2e-05}
\newcommand{\bdRecalLoosePhaseOneRatioSE}{0.04}
\newcommand{\bdRecalLoosePhaseOneRatioCILo}{3.00}
\newcommand{\bdRecalLoosePhaseOneRatioCIHi}{3.15}
\newcommand{\bdRecalLoosePhaseTwoRatio}{0.17\times}
\newcommand{\bdRecalLoosePhaseTwoCost}{\$3.2e-04}
\newcommand{\bdRecalLoosePhaseTwoCostSE}{\$1.2e-06}
\newcommand{\bdRecalLoosePhaseTwoRatioSE}{0.00}
\newcommand{\bdRecalLoosePhaseTwoRatioCILo}{0.17}
\newcommand{\bdRecalLoosePhaseTwoRatioCIHi}{0.17}
\newcommand{\bdRecalLoosePhaseThreeRatio}{0.94\times}
\newcommand{\bdRecalLoosePhaseThreeCost}{\$1.8e-03}
\newcommand{\bdRecalLoosePhaseThreeCostSE}{\$4.5e-05}
\newcommand{\bdRecalLoosePhaseThreeRatioSE}{0.02}
\newcommand{\bdRecalLoosePhaseThreeRatioCILo}{0.90}
\newcommand{\bdRecalLoosePhaseThreeRatioCIHi}{0.99}
\newcommand{\bdForgetLoosePhaseOneRatio}{3.08\times}
\newcommand{\bdForgetLoosePhaseOneCost}{\$5.8e-03}
\newcommand{\bdForgetLoosePhaseOneCostSE}{\$2.9e-04}
\newcommand{\bdForgetLoosePhaseOneRatioSE}{0.15}
\newcommand{\bdForgetLoosePhaseOneRatioCILo}{2.80}
\newcommand{\bdForgetLoosePhaseOneRatioCIHi}{3.39}
\newcommand{\bdForgetLoosePhaseTwoRatio}{0.17\times}
\newcommand{\bdForgetLoosePhaseTwoCost}{\$3.1e-04}
\newcommand{\bdForgetLoosePhaseTwoCostSE}{\$6.2e-06}
\newcommand{\bdForgetLoosePhaseTwoRatioSE}{0.00}
\newcommand{\bdForgetLoosePhaseTwoRatioCILo}{0.16}
\newcommand{\bdForgetLoosePhaseTwoRatioCIHi}{0.17}
\newcommand{\bdForgetLoosePhaseThreeRatio}{4.93\times}
\newcommand{\bdForgetLoosePhaseThreeCost}{\$9.2e-03}
\newcommand{\bdForgetLoosePhaseThreeCostSE}{\$3.1e-04}
\newcommand{\bdForgetLoosePhaseThreeRatioSE}{0.17}
\newcommand{\bdForgetLoosePhaseThreeRatioCILo}{4.61}
\newcommand{\bdForgetLoosePhaseThreeRatioCIHi}{5.24}
\newcommand{\bdParetoBanditLoosePhaseOneRatio}{0.96\times}
\newcommand{\bdParetoBanditLoosePhaseOneCost}{\$1.8e-03}
\newcommand{\bdParetoBanditLoosePhaseOneCostSE}{\$6.6e-06}
\newcommand{\bdParetoBanditLoosePhaseOneRatioSE}{0.00}
\newcommand{\bdParetoBanditLoosePhaseOneRatioCILo}{0.95}
\newcommand{\bdParetoBanditLoosePhaseOneRatioCIHi}{0.96}
\newcommand{\bdParetoBanditLoosePhaseTwoRatio}{0.23\times}
\newcommand{\bdParetoBanditLoosePhaseTwoCost}{\$4.2e-04}
\newcommand{\bdParetoBanditLoosePhaseTwoCostSE}{\$4.9e-06}
\newcommand{\bdParetoBanditLoosePhaseTwoRatioSE}{0.00}
\newcommand{\bdParetoBanditLoosePhaseTwoRatioCILo}{0.22}
\newcommand{\bdParetoBanditLoosePhaseTwoRatioCIHi}{0.23}
\newcommand{\bdParetoBanditLoosePhaseThreeRatio}{0.95\times}
\newcommand{\bdParetoBanditLoosePhaseThreeCost}{\$1.8e-03}
\newcommand{\bdParetoBanditLoosePhaseThreeCostSE}{\$2.3e-05}
\newcommand{\bdParetoBanditLoosePhaseThreeRatioSE}{0.01}
\newcommand{\bdParetoBanditLoosePhaseThreeRatioCILo}{0.92}
\newcommand{\bdParetoBanditLoosePhaseThreeRatioCIHi}{0.97}
\newcommand{\bdParetoBanditLooseLambdaPhaseOne}{0.09}
\newcommand{\bdParetoBanditLooseGeminiPhaseOne}{9}
\newcommand{\bdParetoBanditLooseRewardPhaseOne}{0.9094}
\newcommand{\bdParetoBanditLooseRewardPhaseOneSE}{0.0015}
\newcommand{\bdParetoBanditLooseRewardPhaseOneCILo}{0.9064}
\newcommand{\bdParetoBanditLooseRewardPhaseOneCIHi}{0.9122}
\newcommand{\bdParetoBanditLooseLambdaPhaseTwo}{0.00}
\newcommand{\bdParetoBanditLooseGeminiPhaseTwo}{30}
\newcommand{\bdParetoBanditLooseRewardPhaseTwo}{0.9276}
\newcommand{\bdParetoBanditLooseRewardPhaseTwoSE}{0.0006}
\newcommand{\bdParetoBanditLooseRewardPhaseTwoCILo}{0.9263}
\newcommand{\bdParetoBanditLooseRewardPhaseTwoCIHi}{0.9288}
\newcommand{\bdParetoBanditLooseLambdaPhaseThree}{0.10}
\newcommand{\bdParetoBanditLooseGeminiPhaseThree}{8}
\newcommand{\bdParetoBanditLooseRewardPhaseThree}{0.9119}
\newcommand{\bdParetoBanditLooseRewardPhaseThreeSE}{0.0017}
\newcommand{\bdParetoBanditLooseRewardPhaseThreeCILo}{0.9085}
\newcommand{\bdParetoBanditLooseRewardPhaseThreeCIHi}{0.9150}
\newcommand{\bdParetoBanditLooseRewardLift}{0.018}
\newcommand{\bdParetoBanditLooseRewardLiftSE}{0.0014}
\newcommand{\bdParetoBanditLooseRewardLiftCILo}{0.015}
\newcommand{\bdParetoBanditLooseRewardLiftCIHi}{0.021}
\newcommand{\bdParetoBanditLoosePhaseOneUtil}{0.96\times}
\newcommand{\bdUncPhaseOneCostEng}{\$1.2e-02}
\newcommand{\bdUncPhaseOneReward}{0.9248}
\newcommand{\bdUncPhaseOneRewardCILo}{0.9236}
\newcommand{\bdUncPhaseOneRewardCIHi}{0.9260}
\newcommand{\bdParetoBanditTightCostRatioVsUnc}{38.3}
\newcommand{\bdParetoBanditTightCostSavingPct}{97}
\newcommand{\bdParetoBanditTightRewardGapPct}{6.8}
\newcommand{\bdParetoBanditModCostRatioVsUnc}{17.7}
\newcommand{\bdParetoBanditModCostSavingPct}{94}
\newcommand{\bdParetoBanditModRewardGapPct}{4.6}
\newcommand{\bdParetoBanditLooseCostRatioVsUnc}{6.5}
\newcommand{\bdParetoBanditLooseCostSavingPct}{85}
\newcommand{\bdParetoBanditLooseRewardGapPct}{1.7}


\newcommand{\cfFailureArm}{Mistral-Large}
\newcommand{\cfFailureReward}{0.75}
\newcommand{\cfNseeds}{20}
\newcommand{\cfPhaseN}{608}
\newcommand{\cfGamma}{0.997}
\newcommand{\cfNaiveTightPhaseOneRatio}{1.06\times}
\newcommand{\cfNaiveTightPhaseOneCost}{\$3.2e-04}
\newcommand{\cfNaiveTightPhaseOneReward}{0.8709}
\newcommand{\cfNaiveTightPhaseOneRatioCILo}{1.05}
\newcommand{\cfNaiveTightPhaseOneRatioCIHi}{1.06}
\newcommand{\cfNaiveTightPhaseOneRewardCILo}{0.8700}
\newcommand{\cfNaiveTightPhaseOneRewardCIHi}{0.8717}
\newcommand{\cfNaiveTightPhaseTwoRatio}{0.89\times}
\newcommand{\cfNaiveTightPhaseTwoCost}{\$2.7e-04}
\newcommand{\cfNaiveTightPhaseTwoReward}{0.8024}
\newcommand{\cfNaiveTightPhaseTwoRatioCILo}{0.88}
\newcommand{\cfNaiveTightPhaseTwoRatioCIHi}{0.90}
\newcommand{\cfNaiveTightPhaseTwoRewardCILo}{0.8019}
\newcommand{\cfNaiveTightPhaseTwoRewardCIHi}{0.8030}
\newcommand{\cfNaiveTightPhaseThreeRatio}{0.91\times}
\newcommand{\cfNaiveTightPhaseThreeCost}{\$2.7e-04}
\newcommand{\cfNaiveTightPhaseThreeReward}{0.8551}
\newcommand{\cfNaiveTightPhaseThreeRatioCILo}{0.90}
\newcommand{\cfNaiveTightPhaseThreeRatioCIHi}{0.92}
\newcommand{\cfNaiveTightPhaseThreeRewardCILo}{0.8541}
\newcommand{\cfNaiveTightPhaseThreeRewardCIHi}{0.8560}
\newcommand{\cfForgetTightPhaseOneRatio}{2.56\times}
\newcommand{\cfForgetTightPhaseOneCost}{\$7.7e-04}
\newcommand{\cfForgetTightPhaseOneReward}{0.8837}
\newcommand{\cfForgetTightPhaseOneRatioCILo}{2.35}
\newcommand{\cfForgetTightPhaseOneRatioCIHi}{2.77}
\newcommand{\cfForgetTightPhaseOneRewardCILo}{0.8807}
\newcommand{\cfForgetTightPhaseOneRewardCIHi}{0.8862}
\newcommand{\cfForgetTightMistralPhaseOne}{65}
\newcommand{\cfForgetTightPhaseTwoRatio}{2.80\times}
\newcommand{\cfForgetTightPhaseTwoCost}{\$8.4e-04}
\newcommand{\cfForgetTightPhaseTwoReward}{0.8071}
\newcommand{\cfForgetTightPhaseTwoRatioCILo}{2.19}
\newcommand{\cfForgetTightPhaseTwoRatioCIHi}{3.47}
\newcommand{\cfForgetTightPhaseTwoRewardCILo}{0.8050}
\newcommand{\cfForgetTightPhaseTwoRewardCIHi}{0.8091}
\newcommand{\cfForgetTightMistralPhaseTwo}{40}
\newcommand{\cfForgetTightPhaseThreeRatio}{1.84\times}
\newcommand{\cfForgetTightPhaseThreeCost}{\$5.5e-04}
\newcommand{\cfForgetTightPhaseThreeReward}{0.8257}
\newcommand{\cfForgetTightPhaseThreeRatioCILo}{1.36}
\newcommand{\cfForgetTightPhaseThreeRatioCIHi}{2.42}
\newcommand{\cfForgetTightPhaseThreeRewardCILo}{0.8203}
\newcommand{\cfForgetTightPhaseThreeRewardCIHi}{0.8314}
\newcommand{\cfForgetTightMistralPhaseThree}{27}
\newcommand{\cfParetoBanditTightPhaseOneRatio}{1.00\times}
\newcommand{\cfParetoBanditTightPhaseOneCost}{\$3.0e-04}
\newcommand{\cfParetoBanditTightPhaseOneReward}{0.8632}
\newcommand{\cfParetoBanditTightPhaseOneRatioCILo}{0.99}
\newcommand{\cfParetoBanditTightPhaseOneRatioCIHi}{1.01}
\newcommand{\cfParetoBanditTightPhaseOneRewardCILo}{0.8615}
\newcommand{\cfParetoBanditTightPhaseOneRewardCIHi}{0.8648}
\newcommand{\cfParetoBanditTightMistralPhaseOne}{54}
\newcommand{\cfParetoBanditTightLambdaPhaseOne}{0.45}
\newcommand{\cfParetoBanditTightPhaseTwoRatio}{1.00\times}
\newcommand{\cfParetoBanditTightPhaseTwoCost}{\$3.0e-04}
\newcommand{\cfParetoBanditTightPhaseTwoReward}{0.8004}
\newcommand{\cfParetoBanditTightPhaseTwoRatioCILo}{0.98}
\newcommand{\cfParetoBanditTightPhaseTwoRatioCIHi}{1.01}
\newcommand{\cfParetoBanditTightPhaseTwoRewardCILo}{0.7986}
\newcommand{\cfParetoBanditTightPhaseTwoRewardCIHi}{0.8023}
\newcommand{\cfParetoBanditTightMistralPhaseTwo}{33}
\newcommand{\cfParetoBanditTightLambdaPhaseTwo}{0.58}
\newcommand{\cfParetoBanditTightPhaseThreeRatio}{0.99\times}
\newcommand{\cfParetoBanditTightPhaseThreeCost}{\$3.0e-04}
\newcommand{\cfParetoBanditTightPhaseThreeReward}{0.8185}
\newcommand{\cfParetoBanditTightPhaseThreeRatioCILo}{0.97}
\newcommand{\cfParetoBanditTightPhaseThreeRatioCIHi}{1.02}
\newcommand{\cfParetoBanditTightPhaseThreeRewardCILo}{0.8119}
\newcommand{\cfParetoBanditTightPhaseThreeRewardCIHi}{0.8253}
\newcommand{\cfParetoBanditTightMistralPhaseThree}{26}
\newcommand{\cfParetoBanditTightLambdaPhaseThree}{0.76}
\newcommand{\cfParetoBanditTightRewardDrop}{0.063}
\newcommand{\cfParetoBanditTightRecoveryGap}{0.045}
\newcommand{\cfParetoBanditTightRecoveryCILo}{0.940}
\newcommand{\cfParetoBanditTightRecoveryCIHi}{0.957}
\newcommand{\cfParetoBanditTightRecoveryMean}{0.948}
\newcommand{\cfParetoBanditTightRewardDropCILo}{0.061}
\newcommand{\cfParetoBanditTightRewardDropCIHi}{0.065}
\newcommand{\cfNaiveModPhaseOneRatio}{0.57\times}
\newcommand{\cfNaiveModPhaseOneCost}{\$3.8e-04}
\newcommand{\cfNaiveModPhaseOneReward}{0.8929}
\newcommand{\cfNaiveModPhaseOneRatioCILo}{0.57}
\newcommand{\cfNaiveModPhaseOneRatioCIHi}{0.58}
\newcommand{\cfNaiveModPhaseOneRewardCILo}{0.8922}
\newcommand{\cfNaiveModPhaseOneRewardCIHi}{0.8936}
\newcommand{\cfNaiveModPhaseTwoRatio}{0.56\times}
\newcommand{\cfNaiveModPhaseTwoCost}{\$3.7e-04}
\newcommand{\cfNaiveModPhaseTwoReward}{0.8001}
\newcommand{\cfNaiveModPhaseTwoRatioCILo}{0.54}
\newcommand{\cfNaiveModPhaseTwoRatioCIHi}{0.58}
\newcommand{\cfNaiveModPhaseTwoRewardCILo}{0.7996}
\newcommand{\cfNaiveModPhaseTwoRewardCIHi}{0.8006}
\newcommand{\cfNaiveModPhaseThreeRatio}{1.00\times}
\newcommand{\cfNaiveModPhaseThreeCost}{\$6.6e-04}
\newcommand{\cfNaiveModPhaseThreeReward}{0.8807}
\newcommand{\cfNaiveModPhaseThreeRatioCILo}{0.90}
\newcommand{\cfNaiveModPhaseThreeRatioCIHi}{1.09}
\newcommand{\cfNaiveModPhaseThreeRewardCILo}{0.8800}
\newcommand{\cfNaiveModPhaseThreeRewardCIHi}{0.8815}
\newcommand{\cfForgetModPhaseOneRatio}{1.48\times}
\newcommand{\cfForgetModPhaseOneCost}{\$9.8e-04}
\newcommand{\cfForgetModPhaseOneReward}{0.8979}
\newcommand{\cfForgetModPhaseOneRatioCILo}{1.32}
\newcommand{\cfForgetModPhaseOneRatioCIHi}{1.64}
\newcommand{\cfForgetModPhaseOneRewardCILo}{0.8968}
\newcommand{\cfForgetModPhaseOneRewardCIHi}{0.8990}
\newcommand{\cfForgetModMistralPhaseOne}{72}
\newcommand{\cfForgetModPhaseTwoRatio}{2.07\times}
\newcommand{\cfForgetModPhaseTwoCost}{\$1.4e-03}
\newcommand{\cfForgetModPhaseTwoReward}{0.8085}
\newcommand{\cfForgetModPhaseTwoRatioCILo}{1.83}
\newcommand{\cfForgetModPhaseTwoRatioCIHi}{2.33}
\newcommand{\cfForgetModPhaseTwoRewardCILo}{0.8064}
\newcommand{\cfForgetModPhaseTwoRewardCIHi}{0.8105}
\newcommand{\cfForgetModMistralPhaseTwo}{54}
\newcommand{\cfForgetModPhaseThreeRatio}{2.21\times}
\newcommand{\cfForgetModPhaseThreeCost}{\$1.5e-03}
\newcommand{\cfForgetModPhaseThreeReward}{0.8711}
\newcommand{\cfForgetModPhaseThreeRatioCILo}{1.77}
\newcommand{\cfForgetModPhaseThreeRatioCIHi}{2.69}
\newcommand{\cfForgetModPhaseThreeRewardCILo}{0.8645}
\newcommand{\cfForgetModPhaseThreeRewardCIHi}{0.8780}
\newcommand{\cfForgetModMistralPhaseThree}{50}
\newcommand{\cfParetoBanditModPhaseOneRatio}{0.98\times}
\newcommand{\cfParetoBanditModPhaseOneCost}{\$6.5e-04}
\newcommand{\cfParetoBanditModPhaseOneReward}{0.8850}
\newcommand{\cfParetoBanditModPhaseOneRatioCILo}{0.96}
\newcommand{\cfParetoBanditModPhaseOneRatioCIHi}{1.00}
\newcommand{\cfParetoBanditModPhaseOneRewardCILo}{0.8806}
\newcommand{\cfParetoBanditModPhaseOneRewardCIHi}{0.8891}
\newcommand{\cfParetoBanditModMistralPhaseOne}{71}
\newcommand{\cfParetoBanditModLambdaPhaseOne}{0.33}
\newcommand{\cfParetoBanditModPhaseTwoRatio}{0.99\times}
\newcommand{\cfParetoBanditModPhaseTwoCost}{\$6.6e-04}
\newcommand{\cfParetoBanditModPhaseTwoReward}{0.7947}
\newcommand{\cfParetoBanditModPhaseTwoRatioCILo}{0.98}
\newcommand{\cfParetoBanditModPhaseTwoRatioCIHi}{1.00}
\newcommand{\cfParetoBanditModPhaseTwoRewardCILo}{0.7928}
\newcommand{\cfParetoBanditModPhaseTwoRewardCIHi}{0.7964}
\newcommand{\cfParetoBanditModMistralPhaseTwo}{50}
\newcommand{\cfParetoBanditModLambdaPhaseTwo}{0.37}
\newcommand{\cfParetoBanditModPhaseThreeRatio}{0.97\times}
\newcommand{\cfParetoBanditModPhaseThreeCost}{\$6.4e-04}
\newcommand{\cfParetoBanditModPhaseThreeReward}{0.8630}
\newcommand{\cfParetoBanditModPhaseThreeRatioCILo}{0.95}
\newcommand{\cfParetoBanditModPhaseThreeRatioCIHi}{0.98}
\newcommand{\cfParetoBanditModPhaseThreeRewardCILo}{0.8572}
\newcommand{\cfParetoBanditModPhaseThreeRewardCIHi}{0.8684}
\newcommand{\cfParetoBanditModMistralPhaseThree}{54}
\newcommand{\cfParetoBanditModLambdaPhaseThree}{0.25}
\newcommand{\cfParetoBanditModRewardDrop}{0.090}
\newcommand{\cfParetoBanditModRecoveryGap}{0.022}
\newcommand{\cfParetoBanditModRecoveryCILo}{0.967}
\newcommand{\cfParetoBanditModRecoveryCIHi}{0.982}
\newcommand{\cfParetoBanditModRecoveryMean}{0.975}
\newcommand{\cfParetoBanditModRewardDropCILo}{0.085}
\newcommand{\cfParetoBanditModRewardDropCIHi}{0.096}
\newcommand{\cfNaiveLoosePhaseOneRatio}{3.03\times}
\newcommand{\cfNaiveLoosePhaseOneCost}{\$5.7e-03}
\newcommand{\cfNaiveLoosePhaseOneReward}{0.9280}
\newcommand{\cfNaiveLoosePhaseOneRatioCILo}{2.94}
\newcommand{\cfNaiveLoosePhaseOneRatioCIHi}{3.12}
\newcommand{\cfNaiveLoosePhaseOneRewardCILo}{0.9274}
\newcommand{\cfNaiveLoosePhaseOneRewardCIHi}{0.9286}
\newcommand{\cfNaiveLoosePhaseTwoRatio}{4.23\times}
\newcommand{\cfNaiveLoosePhaseTwoCost}{\$7.9e-03}
\newcommand{\cfNaiveLoosePhaseTwoReward}{0.8453}
\newcommand{\cfNaiveLoosePhaseTwoRatioCILo}{4.16}
\newcommand{\cfNaiveLoosePhaseTwoRatioCIHi}{4.29}
\newcommand{\cfNaiveLoosePhaseTwoRewardCILo}{0.8440}
\newcommand{\cfNaiveLoosePhaseTwoRewardCIHi}{0.8466}
\newcommand{\cfNaiveLoosePhaseThreeRatio}{5.15\times}
\newcommand{\cfNaiveLoosePhaseThreeCost}{\$9.6e-03}
\newcommand{\cfNaiveLoosePhaseThreeReward}{0.9277}
\newcommand{\cfNaiveLoosePhaseThreeRatioCILo}{5.10}
\newcommand{\cfNaiveLoosePhaseThreeRatioCIHi}{5.21}
\newcommand{\cfNaiveLoosePhaseThreeRewardCILo}{0.9272}
\newcommand{\cfNaiveLoosePhaseThreeRewardCIHi}{0.9281}
\newcommand{\cfForgetLoosePhaseOneRatio}{3.04\times}
\newcommand{\cfForgetLoosePhaseOneCost}{\$5.7e-03}
\newcommand{\cfForgetLoosePhaseOneReward}{0.9190}
\newcommand{\cfForgetLoosePhaseOneRatioCILo}{2.74}
\newcommand{\cfForgetLoosePhaseOneRatioCIHi}{3.33}
\newcommand{\cfForgetLoosePhaseOneRewardCILo}{0.9176}
\newcommand{\cfForgetLoosePhaseOneRewardCIHi}{0.9206}
\newcommand{\cfForgetLooseMistralPhaseOne}{63}
\newcommand{\cfForgetLoosePhaseTwoRatio}{6.19\times}
\newcommand{\cfForgetLoosePhaseTwoCost}{\$1.2e-02}
\newcommand{\cfForgetLoosePhaseTwoReward}{0.8992}
\newcommand{\cfForgetLoosePhaseTwoRatioCILo}{5.96}
\newcommand{\cfForgetLoosePhaseTwoRatioCIHi}{6.40}
\newcommand{\cfForgetLoosePhaseTwoRewardCILo}{0.8938}
\newcommand{\cfForgetLoosePhaseTwoRewardCIHi}{0.9042}
\newcommand{\cfForgetLooseMistralPhaseTwo}{19}
\newcommand{\cfForgetLoosePhaseThreeRatio}{6.92\times}
\newcommand{\cfForgetLoosePhaseThreeCost}{\$1.3e-02}
\newcommand{\cfForgetLoosePhaseThreeReward}{0.9232}
\newcommand{\cfForgetLoosePhaseThreeRatioCILo}{6.80}
\newcommand{\cfForgetLoosePhaseThreeRatioCIHi}{7.05}
\newcommand{\cfForgetLoosePhaseThreeRewardCILo}{0.9223}
\newcommand{\cfForgetLoosePhaseThreeRewardCIHi}{0.9242}
\newcommand{\cfForgetLooseMistralPhaseThree}{13}
\newcommand{\cfParetoBanditLoosePhaseOneRatio}{0.95\times}
\newcommand{\cfParetoBanditLoosePhaseOneCost}{\$1.8e-03}
\newcommand{\cfParetoBanditLoosePhaseOneReward}{0.9103}
\newcommand{\cfParetoBanditLoosePhaseOneRatioCILo}{0.95}
\newcommand{\cfParetoBanditLoosePhaseOneRatioCIHi}{0.96}
\newcommand{\cfParetoBanditLoosePhaseOneRewardCILo}{0.9083}
\newcommand{\cfParetoBanditLoosePhaseOneRewardCIHi}{0.9122}
\newcommand{\cfParetoBanditLooseMistralPhaseOne}{84}
\newcommand{\cfParetoBanditLooseLambdaPhaseOne}{0.08}
\newcommand{\cfParetoBanditLoosePhaseTwoRatio}{0.97\times}
\newcommand{\cfParetoBanditLoosePhaseTwoCost}{\$1.8e-03}
\newcommand{\cfParetoBanditLoosePhaseTwoReward}{0.7936}
\newcommand{\cfParetoBanditLoosePhaseTwoRatioCILo}{0.96}
\newcommand{\cfParetoBanditLoosePhaseTwoRatioCIHi}{0.98}
\newcommand{\cfParetoBanditLoosePhaseTwoRewardCILo}{0.7910}
\newcommand{\cfParetoBanditLoosePhaseTwoRewardCIHi}{0.7961}
\newcommand{\cfParetoBanditLooseMistralPhaseTwo}{69}
\newcommand{\cfParetoBanditLooseLambdaPhaseTwo}{0.09}
\newcommand{\cfParetoBanditLoosePhaseThreeRatio}{0.97\times}
\newcommand{\cfParetoBanditLoosePhaseThreeCost}{\$1.8e-03}
\newcommand{\cfParetoBanditLoosePhaseThreeReward}{0.8977}
\newcommand{\cfParetoBanditLoosePhaseThreeRatioCILo}{0.96}
\newcommand{\cfParetoBanditLoosePhaseThreeRatioCIHi}{0.98}
\newcommand{\cfParetoBanditLoosePhaseThreeRewardCILo}{0.8940}
\newcommand{\cfParetoBanditLoosePhaseThreeRewardCIHi}{0.9012}
\newcommand{\cfParetoBanditLooseMistralPhaseThree}{69}
\newcommand{\cfParetoBanditLooseLambdaPhaseThree}{0.10}
\newcommand{\cfParetoBanditLooseRewardDrop}{0.117}
\newcommand{\cfParetoBanditLooseRecoveryGap}{0.013}
\newcommand{\cfParetoBanditLooseRecoveryCILo}{0.982}
\newcommand{\cfParetoBanditLooseRecoveryCIHi}{0.991}
\newcommand{\cfParetoBanditLooseRecoveryMean}{0.986}
\newcommand{\cfParetoBanditLooseRewardDropCILo}{0.114}
\newcommand{\cfParetoBanditLooseRewardDropCIHi}{0.120}
\newcommand{\cfParetoBanditTightTroughOffset}{192}
\newcommand{\cfParetoBanditTightTroughReward}{0.790}
\newcommand{\cfParetoBanditTightTroughDropPct}{8.5}
\newcommand{\cfParetoBanditTightAdaptSteps}{192}
\newcommand{\cfParetoBanditTightPhaseTwoSteadyReward}{0.800}
\newcommand{\cfParetoBanditTightPhaseTwoRewardGapPct}{7.3}
\newcommand{\cfParetoBanditTightPhaseTwoCostChangePct}{-14.8}
\newcommand{\cfParetoBanditModTroughOffset}{52}
\newcommand{\cfParetoBanditModTroughReward}{0.780}
\newcommand{\cfParetoBanditModTroughDropPct}{12.2}
\newcommand{\cfParetoBanditModAdaptSteps}{52}
\newcommand{\cfParetoBanditModPhaseTwoSteadyReward}{0.801}
\newcommand{\cfParetoBanditModPhaseTwoRewardGapPct}{9.8}
\newcommand{\cfParetoBanditModPhaseTwoCostChangePct}{-13.2}
\newcommand{\cfParetoBanditLooseTroughOffset}{72}
\newcommand{\cfParetoBanditLooseTroughReward}{0.777}
\newcommand{\cfParetoBanditLooseTroughDropPct}{14.7}
\newcommand{\cfParetoBanditLooseAdaptSteps}{72}
\newcommand{\cfParetoBanditLoosePhaseTwoSteadyReward}{0.802}
\newcommand{\cfParetoBanditLoosePhaseTwoRewardGapPct}{11.9}
\newcommand{\cfParetoBanditLoosePhaseTwoCostChangePct}{-1.7}
\newcommand{\cfUncPhaseOneReward}{0.9245}
\newcommand{\cfUncPhaseOneCost}{\$1.2e-02}
\newcommand{\cfUncPhaseOneRewardCILo}{0.9235}
\newcommand{\cfUncPhaseOneRewardCIHi}{0.9255}
\newcommand{\cfUncPhaseTwoReward}{0.9263}
\newcommand{\cfUncPhaseTwoCost}{\$1.4e-02}
\newcommand{\cfUncPhaseTwoRewardCILo}{0.9253}
\newcommand{\cfUncPhaseTwoRewardCIHi}{0.9271}
\newcommand{\cfUncPhaseThreeReward}{0.9251}
\newcommand{\cfUncPhaseThreeCost}{\$1.5e-02}
\newcommand{\cfUncPhaseThreeRewardCILo}{0.9244}
\newcommand{\cfUncPhaseThreeRewardCIHi}{0.9257}
\newcommand{\cfUncPhaseOneCostBaseline}{\$1.2e-02}
\newcommand{\cfUncPhaseTwoSteadyCost}{\$1.5e-02}
\newcommand{\cfUncCostSpikePct}{24.2}
\newcommand{\cfUncPhaseTwoSteadyRewardShort}{0.934}


\newcommand{\hpGridAlpha}{6}
\newcommand{\hpGridGamma}{7}
\newcommand{\hpTAdapt}{500}
\newcommand{\hpSelectionMethod}{pareto\_knee\_point}
\newcommand{\hpParetoBanditAlpha}{0.01}
\newcommand{\hpParetoBanditNeff}{1164}
\newcommand{\hpParetoBanditGamma}{0.997}
\newcommand{\hpParetoBanditAUC}{0.928}
\newcommand{\hpParetoBanditPTwoReward}{0.7312}
\newcommand{\hpEffMemSteps}{333}
\newcommand{\hpHalfLife}{231}
\newcommand{\hpParetoBanditTestAUC}{0.9221}
\newcommand{\hpParetoBanditTestStd}{0.0012}
\newcommand{\hpParetoBanditTestDelta}{-0.35}
\newcommand{\hpParetoBanditTestAUCCILo}{0.9216}
\newcommand{\hpParetoBanditTestAUCCIHi}{0.9226}
\newcommand{\hpTabulaAlpha}{0.05}
\newcommand{\hpTabulaGamma}{0.997}
\newcommand{\hpTabulaAUC}{0.923}
\newcommand{\hpTabulaPTwoReward}{0.7287}
\newcommand{\hpTREffMemSteps}{333}
\newcommand{\hpTRHalfLife}{231}
\newcommand{\hpTabulaTestAUC}{0.9200}
\newcommand{\hpTabulaTestStd}{0.0020}
\newcommand{\hpTabulaTestDelta}{-0.58}
\newcommand{\hpTabulaTestAUCCILo}{0.9191}
\newcommand{\hpTabulaTestAUCCIHi}{0.9208}
\newcommand{\hpFixedTestAUC}{0.9253}
\newcommand{\hpAUCOnlyAlpha}{0.01}
\newcommand{\hpAUCOnlyNeff}{500}
\newcommand{\hpAUCOnlyGamma}{1.0}
\newcommand{\hpAUCOnlyAUC}{0.928}
\newcommand{\hpTabulaAUCOnlyAlpha}{0.10}
\newcommand{\hpTabulaAUCOnlyGamma}{1.0}
\newcommand{\hpTabulaAUCOnlyAUC}{0.925}
\newcommand{\hpAUCSacrifice}{0.08}
\newcommand{\hpCrossArmLlamaPTwo}{0.8470}
\newcommand{\hpCrossArmLlamaStd}{0.0752}
\newcommand{\hpCrossArmLlamaPTwoCILo}{0.8140}
\newcommand{\hpCrossArmLlamaPTwoCIHi}{0.8800}
\newcommand{\hpCrossArmMistralPTwo}{0.7312}
\newcommand{\hpCrossArmMistralStd}{0.0184}
\newcommand{\hpCrossArmMistralPTwoCILo}{0.7231}
\newcommand{\hpCrossArmMistralPTwoCIHi}{0.7393}
\newcommand{\hpCrossArmGeminiPTwo}{0.8916}
\newcommand{\hpCrossArmGeminiStd}{0.0147}
\newcommand{\hpCrossArmGeminiPTwoCILo}{0.8852}
\newcommand{\hpCrossArmGeminiPTwoCIHi}{0.8980}
\newcommand{\hpTestCrossArmLlamaPTwo}{0.8500}
\newcommand{\hpTestCrossArmLlamaStd}{0.0775}
\newcommand{\hpTestCrossArmMistralPTwo}{0.7470}
\newcommand{\hpTestCrossArmMistralStd}{0.0236}
\newcommand{\hpTestCrossArmGeminiPTwo}{0.8935}
\newcommand{\hpTestCrossArmGeminiStd}{0.0165}
\newcommand{\hpBootKneeFreq}{50.9}
\newcommand{\hpBootNeighborFreq}{91.0}
\newcommand{\hpBootNUnique}{6}
\newcommand{\hpBootNIter}{2000}
\newcommand{\hpTabulaBootKneeFreq}{32.0}
\newcommand{\hpTabulaBootNeighborFreq}{48.4}
\newcommand{\hpSensLoAlpha}{0.01}
\newcommand{\hpSensLoGamma}{0.996}
\newcommand{\hpSensLoNeff}{431}
\newcommand{\hpSensLoAUC}{0.9255}
\newcommand{\hpSensLoPTwo}{0.7934}
\newcommand{\hpSensMidAlpha}{0.01}
\newcommand{\hpSensMidGamma}{0.997}
\newcommand{\hpSensMidNeff}{1164}
\newcommand{\hpSensMidAUC}{0.9277}
\newcommand{\hpSensMidPTwo}{0.7933}
\newcommand{\hpSensHiAlpha}{0.01}
\newcommand{\hpSensHiGamma}{0.994}
\newcommand{\hpSensHiNeff}{68298}
\newcommand{\hpSensHiAUC}{0.9267}
\newcommand{\hpSensHiPTwo}{0.7945}
\newcommand{\hpSensStable}{no}


\newcommand{\chNPromptsK}{1766}
\newcommand{\chNDroppedK}{0}
\newcommand{\chFullOrderK}{100.0}
\newcommand{\chFullOrderKCILo}{99.8}
\newcommand{\chFullOrderKCIHi}{100.0}
\newcommand{\chMinPairwiseK}{100.0}
\newcommand{\chCVRangeK}{0.63--0.92}
\newcommand{\chLogStdFracRange}{8--11}
\newcommand{\chMinCohensD}{4.39}
\newcommand{\chMaxCohensD}{5.92}
\newcommand{\chCorrRange}{0.12--0.27}
\newcommand{\chCrossCorrRange}{0.56--0.68}
\newcommand{\chNPromptsKFour}{1766}
\newcommand{\chNDroppedKFour}{0}
\newcommand{\chKFourMissing}{19}
\newcommand{\chFullOrderKFour}{79.7}
\newcommand{\chFullOrderKFourCILo}{77.7}
\newcommand{\chFullOrderKFourCIHi}{81.5}
\newcommand{\chMistralFlashPW}{79.7}
\newcommand{\chMistralFlashPWCILo}{77.7}
\newcommand{\chMistralFlashPWCIHi}{81.5}
\newcommand{\chMistralFlashTies}{0}
\newcommand{\chFlashCtilde}{0.382}
\newcommand{\chMistralCtilde}{0.333}
\newcommand{\chProCtilde}{0.583}
\newcommand{\chFlashCV}{1.56}
\newcommand{\chMistralFlashGap}{0.049}
\newcommand{\chCVRangeKFour}{0.63--1.56}
\newcommand{\chMFCohensD}{0.68}


\newcommand{\moNseeds}{20}
\newcommand{\moPhaseOneN}{1785}
\newcommand{\moPhaseTwoN}{1793}
\newcommand{\moBurninPulls}{20}
\newcommand{\moAlpha}{0.01}
\newcommand{\moNeff}{1163.9}
\newcommand{\moGamma}{0.997}
\newcommand{\moBudgetTight}{\$3.0e-04}
\newcommand{\moBudgetMod}{\$6.6e-04}
\newcommand{\moBudgetLoose}{\$1.9e-03}
\newcommand{\moGoodCheapPBTightPhaseTwoReward}{0.866}
\newcommand{\moGoodCheapPBTightPhaseTwoCost}{\$3.0e-04}
\newcommand{\moGoodCheapPBTightOverallReward}{0.872}
\newcommand{\moGoodCheapPBTightOverallCost}{\$3.0e-04}
\newcommand{\moGoodCheapPBTightPhaseTwoRewardCILo}{0.864}
\newcommand{\moGoodCheapPBTightPhaseTwoRewardCIHi}{0.869}
\newcommand{\moGoodCheapPBTightOverallRewardCILo}{0.870}
\newcommand{\moGoodCheapPBTightOverallRewardCIHi}{0.873}
\newcommand{\moGoodCheapPBTightPhaseTwoCostCILo}{0.000299}
\newcommand{\moGoodCheapPBTightPhaseTwoCostCIHi}{0.000301}
\newcommand{\moGoodCheapPBTightOverallCostCILo}{0.000300}
\newcommand{\moGoodCheapPBTightOverallCostCIHi}{0.000301}
\newcommand{\moGoodCheapPBTightFlashPct}{8.6}
\newcommand{\moGoodCheapPBTightFlashPctCILo}{7.9}
\newcommand{\moGoodCheapPBTightFlashPctCIHi}{9.4}
\newcommand{\moGoodCheapPBTightNSustained}{20}
\newcommand{\moGoodCheapPBTightSustainedStep}{142}
\newcommand{\moGoodCheapPBTightFlashFinalPct}{4.4}
\newcommand{\moGoodCheapPBTightFlashFinalPctCILo}{3.6}
\newcommand{\moGoodCheapPBTightFlashFinalPctCIHi}{5.2}
\newcommand{\moGoodCheapPBModPhaseTwoReward}{0.899}
\newcommand{\moGoodCheapPBModPhaseTwoCost}{\$6.5e-04}
\newcommand{\moGoodCheapPBModOverallReward}{0.894}
\newcommand{\moGoodCheapPBModOverallCost}{\$6.5e-04}
\newcommand{\moGoodCheapPBModPhaseTwoRewardCILo}{0.897}
\newcommand{\moGoodCheapPBModPhaseTwoRewardCIHi}{0.902}
\newcommand{\moGoodCheapPBModOverallRewardCILo}{0.892}
\newcommand{\moGoodCheapPBModOverallRewardCIHi}{0.896}
\newcommand{\moGoodCheapPBModPhaseTwoCostCILo}{0.000649}
\newcommand{\moGoodCheapPBModPhaseTwoCostCIHi}{0.000654}
\newcommand{\moGoodCheapPBModOverallCostCILo}{0.000654}
\newcommand{\moGoodCheapPBModOverallCostCIHi}{0.000656}
\newcommand{\moGoodCheapPBModFlashPct}{12.9}
\newcommand{\moGoodCheapPBModFlashPctCILo}{11.1}
\newcommand{\moGoodCheapPBModFlashPctCIHi}{14.8}
\newcommand{\moGoodCheapPBModNSustained}{20}
\newcommand{\moGoodCheapPBModSustainedStep}{121}
\newcommand{\moGoodCheapPBModFlashFinalPct}{7.8}
\newcommand{\moGoodCheapPBModFlashFinalPctCILo}{6.4}
\newcommand{\moGoodCheapPBModFlashFinalPctCIHi}{9.2}
\newcommand{\moGoodCheapPBLoosePhaseTwoReward}{0.918}
\newcommand{\moGoodCheapPBLoosePhaseTwoCost}{\$1.7e-03}
\newcommand{\moGoodCheapPBLooseOverallReward}{0.918}
\newcommand{\moGoodCheapPBLooseOverallCost}{\$1.8e-03}
\newcommand{\moGoodCheapPBLoosePhaseTwoRewardCILo}{0.916}
\newcommand{\moGoodCheapPBLoosePhaseTwoRewardCIHi}{0.919}
\newcommand{\moGoodCheapPBLooseOverallRewardCILo}{0.917}
\newcommand{\moGoodCheapPBLooseOverallRewardCIHi}{0.919}
\newcommand{\moGoodCheapPBLoosePhaseTwoCostCILo}{0.001669}
\newcommand{\moGoodCheapPBLoosePhaseTwoCostCIHi}{0.001710}
\newcommand{\moGoodCheapPBLooseOverallCostCILo}{0.001746}
\newcommand{\moGoodCheapPBLooseOverallCostCIHi}{0.001770}
\newcommand{\moGoodCheapPBLooseFlashPct}{15.9}
\newcommand{\moGoodCheapPBLooseFlashPctCILo}{13.2}
\newcommand{\moGoodCheapPBLooseFlashPctCIHi}{18.5}
\newcommand{\moGoodCheapPBLooseNSustained}{20}
\newcommand{\moGoodCheapPBLooseSustainedStep}{121}
\newcommand{\moGoodCheapPBLooseFlashFinalPct}{10.2}
\newcommand{\moGoodCheapPBLooseFlashFinalPctCILo}{7.4}
\newcommand{\moGoodCheapPBLooseFlashFinalPctCIHi}{13.1}
\newcommand{\moGoodCheapPBUncPhaseTwoReward}{0.929}
\newcommand{\moGoodCheapPBUncPhaseTwoCost}{\$1.1e-02}
\newcommand{\moGoodCheapPBUncOverallReward}{0.932}
\newcommand{\moGoodCheapPBUncOverallCost}{\$1.1e-02}
\newcommand{\moGoodCheapPBUncPhaseTwoRewardCILo}{0.928}
\newcommand{\moGoodCheapPBUncPhaseTwoRewardCIHi}{0.930}
\newcommand{\moGoodCheapPBUncOverallRewardCILo}{0.932}
\newcommand{\moGoodCheapPBUncOverallRewardCIHi}{0.932}
\newcommand{\moGoodCheapPBUncPhaseTwoCostCILo}{0.010180}
\newcommand{\moGoodCheapPBUncPhaseTwoCostCIHi}{0.010917}
\newcommand{\moGoodCheapPBUncOverallCostCILo}{0.010574}
\newcommand{\moGoodCheapPBUncOverallCostCIHi}{0.010942}
\newcommand{\moGoodCheapPBUncFlashPct}{13.3}
\newcommand{\moGoodCheapPBUncFlashPctCILo}{12.1}
\newcommand{\moGoodCheapPBUncFlashPctCIHi}{14.5}
\newcommand{\moGoodCheapPBUncNSustained}{20}
\newcommand{\moGoodCheapPBUncSustainedStep}{121}
\newcommand{\moGoodCheapPBUncFlashFinalPct}{7.3}
\newcommand{\moGoodCheapPBUncFlashFinalPctCILo}{6.2}
\newcommand{\moGoodCheapPBUncFlashFinalPctCIHi}{8.3}
\newcommand{\moBadCheapPBTightPhaseTwoReward}{0.871}
\newcommand{\moBadCheapPBTightPhaseTwoCost}{\$3.0e-04}
\newcommand{\moBadCheapPBTightOverallReward}{0.874}
\newcommand{\moBadCheapPBTightOverallCost}{\$3.0e-04}
\newcommand{\moBadCheapPBTightPhaseTwoRewardCILo}{0.868}
\newcommand{\moBadCheapPBTightPhaseTwoRewardCIHi}{0.873}
\newcommand{\moBadCheapPBTightOverallRewardCILo}{0.872}
\newcommand{\moBadCheapPBTightOverallRewardCIHi}{0.876}
\newcommand{\moBadCheapPBTightPhaseTwoCostCILo}{0.000299}
\newcommand{\moBadCheapPBTightPhaseTwoCostCIHi}{0.000301}
\newcommand{\moBadCheapPBTightOverallCostCILo}{0.000300}
\newcommand{\moBadCheapPBTightOverallCostCIHi}{0.000302}
\newcommand{\moBadCheapPBTightFlashPct}{1.1}
\newcommand{\moBadCheapPBTightFlashPctCILo}{1.1}
\newcommand{\moBadCheapPBTightFlashPctCIHi}{1.1}
\newcommand{\moBadCheapPBTightNSustained}{0}
\newcommand{\moBadCheapPBTightFlashFinalPct}{0.0}
\newcommand{\moBadCheapPBTightFlashFinalPctCILo}{0.0}
\newcommand{\moBadCheapPBTightFlashFinalPctCIHi}{0.0}
\newcommand{\moBadCheapPBModPhaseTwoReward}{0.895}
\newcommand{\moBadCheapPBModPhaseTwoCost}{\$6.5e-04}
\newcommand{\moBadCheapPBModOverallReward}{0.892}
\newcommand{\moBadCheapPBModOverallCost}{\$6.5e-04}
\newcommand{\moBadCheapPBModPhaseTwoRewardCILo}{0.892}
\newcommand{\moBadCheapPBModPhaseTwoRewardCIHi}{0.899}
\newcommand{\moBadCheapPBModOverallRewardCILo}{0.890}
\newcommand{\moBadCheapPBModOverallRewardCIHi}{0.895}
\newcommand{\moBadCheapPBModPhaseTwoCostCILo}{0.000646}
\newcommand{\moBadCheapPBModPhaseTwoCostCIHi}{0.000653}
\newcommand{\moBadCheapPBModOverallCostCILo}{0.000652}
\newcommand{\moBadCheapPBModOverallCostCIHi}{0.000655}
\newcommand{\moBadCheapPBModFlashPct}{1.1}
\newcommand{\moBadCheapPBModFlashPctCILo}{1.1}
\newcommand{\moBadCheapPBModFlashPctCIHi}{1.2}
\newcommand{\moBadCheapPBModNSustained}{0}
\newcommand{\moBadCheapPBModFlashFinalPct}{0.0}
\newcommand{\moBadCheapPBModFlashFinalPctCILo}{0.0}
\newcommand{\moBadCheapPBModFlashFinalPctCIHi}{0.0}
\newcommand{\moBadCheapPBLoosePhaseTwoReward}{0.913}
\newcommand{\moBadCheapPBLoosePhaseTwoCost}{\$1.7e-03}
\newcommand{\moBadCheapPBLooseOverallReward}{0.916}
\newcommand{\moBadCheapPBLooseOverallCost}{\$1.8e-03}
\newcommand{\moBadCheapPBLoosePhaseTwoRewardCILo}{0.912}
\newcommand{\moBadCheapPBLoosePhaseTwoRewardCIHi}{0.915}
\newcommand{\moBadCheapPBLooseOverallRewardCILo}{0.915}
\newcommand{\moBadCheapPBLooseOverallRewardCIHi}{0.917}
\newcommand{\moBadCheapPBLoosePhaseTwoCostCILo}{0.001656}
\newcommand{\moBadCheapPBLoosePhaseTwoCostCIHi}{0.001698}
\newcommand{\moBadCheapPBLooseOverallCostCILo}{0.001741}
\newcommand{\moBadCheapPBLooseOverallCostCIHi}{0.001762}
\newcommand{\moBadCheapPBLooseFlashPct}{1.1}
\newcommand{\moBadCheapPBLooseFlashPctCILo}{1.1}
\newcommand{\moBadCheapPBLooseFlashPctCIHi}{1.1}
\newcommand{\moBadCheapPBLooseNSustained}{0}
\newcommand{\moBadCheapPBLooseFlashFinalPct}{0.0}
\newcommand{\moBadCheapPBLooseFlashFinalPctCILo}{0.0}
\newcommand{\moBadCheapPBLooseFlashFinalPctCIHi}{0.0}
\newcommand{\moBadCheapPBUncPhaseTwoReward}{0.925}
\newcommand{\moBadCheapPBUncPhaseTwoCost}{\$1.2e-02}
\newcommand{\moBadCheapPBUncOverallReward}{0.930}
\newcommand{\moBadCheapPBUncOverallCost}{\$1.1e-02}
\newcommand{\moBadCheapPBUncPhaseTwoRewardCILo}{0.924}
\newcommand{\moBadCheapPBUncPhaseTwoRewardCIHi}{0.926}
\newcommand{\moBadCheapPBUncOverallRewardCILo}{0.930}
\newcommand{\moBadCheapPBUncOverallRewardCIHi}{0.930}
\newcommand{\moBadCheapPBUncPhaseTwoCostCILo}{0.011714}
\newcommand{\moBadCheapPBUncPhaseTwoCostCIHi}{0.012203}
\newcommand{\moBadCheapPBUncOverallCostCILo}{0.011315}
\newcommand{\moBadCheapPBUncOverallCostCIHi}{0.011615}
\newcommand{\moBadCheapPBUncFlashPct}{1.1}
\newcommand{\moBadCheapPBUncFlashPctCILo}{1.1}
\newcommand{\moBadCheapPBUncFlashPctCIHi}{1.1}
\newcommand{\moBadCheapPBUncNSustained}{0}
\newcommand{\moBadCheapPBUncFlashFinalPct}{0.0}
\newcommand{\moBadCheapPBUncFlashFinalPctCILo}{0.0}
\newcommand{\moBadCheapPBUncFlashFinalPctCIHi}{0.0}
\newcommand{\moGoodExpPBTightPhaseTwoReward}{0.832}
\newcommand{\moGoodExpPBTightPhaseTwoCost}{\$3.9e-04}
\newcommand{\moGoodExpPBTightOverallReward}{0.854}
\newcommand{\moGoodExpPBTightOverallCost}{\$3.5e-04}
\newcommand{\moGoodExpPBTightPhaseTwoRewardCILo}{0.828}
\newcommand{\moGoodExpPBTightPhaseTwoRewardCIHi}{0.835}
\newcommand{\moGoodExpPBTightOverallRewardCILo}{0.852}
\newcommand{\moGoodExpPBTightOverallRewardCIHi}{0.857}
\newcommand{\moGoodExpPBTightPhaseTwoCostCILo}{0.000378}
\newcommand{\moGoodExpPBTightPhaseTwoCostCIHi}{0.000409}
\newcommand{\moGoodExpPBTightOverallCostCILo}{0.000340}
\newcommand{\moGoodExpPBTightOverallCostCIHi}{0.000356}
\newcommand{\moGoodExpPBTightFlashPct}{3.0}
\newcommand{\moGoodExpPBTightFlashPctCILo}{2.8}
\newcommand{\moGoodExpPBTightFlashPctCIHi}{3.1}
\newcommand{\moGoodExpPBTightNSustained}{0}
\newcommand{\moGoodExpPBTightFlashFinalPct}{1.2}
\newcommand{\moGoodExpPBTightFlashFinalPctCILo}{0.8}
\newcommand{\moGoodExpPBTightFlashFinalPctCIHi}{1.5}
\newcommand{\moGoodExpPBModPhaseTwoReward}{0.868}
\newcommand{\moGoodExpPBModPhaseTwoCost}{\$7.1e-04}
\newcommand{\moGoodExpPBModOverallReward}{0.879}
\newcommand{\moGoodExpPBModOverallCost}{\$6.8e-04}
\newcommand{\moGoodExpPBModPhaseTwoRewardCILo}{0.863}
\newcommand{\moGoodExpPBModPhaseTwoRewardCIHi}{0.873}
\newcommand{\moGoodExpPBModOverallRewardCILo}{0.876}
\newcommand{\moGoodExpPBModOverallRewardCIHi}{0.882}
\newcommand{\moGoodExpPBModPhaseTwoCostCILo}{0.000694}
\newcommand{\moGoodExpPBModPhaseTwoCostCIHi}{0.000723}
\newcommand{\moGoodExpPBModOverallCostCILo}{0.000676}
\newcommand{\moGoodExpPBModOverallCostCIHi}{0.000691}
\newcommand{\moGoodExpPBModFlashPct}{4.8}
\newcommand{\moGoodExpPBModFlashPctCILo}{4.5}
\newcommand{\moGoodExpPBModFlashPctCIHi}{5.1}
\newcommand{\moGoodExpPBModNSustained}{0}
\newcommand{\moGoodExpPBModFlashFinalPct}{2.5}
\newcommand{\moGoodExpPBModFlashFinalPctCILo}{1.9}
\newcommand{\moGoodExpPBModFlashFinalPctCIHi}{3.2}
\newcommand{\moGoodExpPBLoosePhaseTwoReward}{0.904}
\newcommand{\moGoodExpPBLoosePhaseTwoCost}{\$1.8e-03}
\newcommand{\moGoodExpPBLooseOverallReward}{0.911}
\newcommand{\moGoodExpPBLooseOverallCost}{\$1.8e-03}
\newcommand{\moGoodExpPBLoosePhaseTwoRewardCILo}{0.900}
\newcommand{\moGoodExpPBLoosePhaseTwoRewardCIHi}{0.907}
\newcommand{\moGoodExpPBLooseOverallRewardCILo}{0.909}
\newcommand{\moGoodExpPBLooseOverallRewardCIHi}{0.913}
\newcommand{\moGoodExpPBLoosePhaseTwoCostCILo}{0.001764}
\newcommand{\moGoodExpPBLoosePhaseTwoCostCIHi}{0.001789}
\newcommand{\moGoodExpPBLooseOverallCostCILo}{0.001794}
\newcommand{\moGoodExpPBLooseOverallCostCIHi}{0.001808}
\newcommand{\moGoodExpPBLooseFlashPct}{9.4}
\newcommand{\moGoodExpPBLooseFlashPctCILo}{8.4}
\newcommand{\moGoodExpPBLooseFlashPctCIHi}{10.4}
\newcommand{\moGoodExpPBLooseNSustained}{20}
\newcommand{\moGoodExpPBLooseSustainedStep}{150}
\newcommand{\moGoodExpPBLooseFlashFinalPct}{6.1}
\newcommand{\moGoodExpPBLooseFlashFinalPctCILo}{4.8}
\newcommand{\moGoodExpPBLooseFlashFinalPctCIHi}{7.5}
\newcommand{\moGoodExpPBUncPhaseTwoReward}{0.929}
\newcommand{\moGoodExpPBUncPhaseTwoCost}{\$1.2e-02}
\newcommand{\moGoodExpPBUncOverallReward}{0.932}
\newcommand{\moGoodExpPBUncOverallCost}{\$1.1e-02}
\newcommand{\moGoodExpPBUncPhaseTwoRewardCILo}{0.928}
\newcommand{\moGoodExpPBUncPhaseTwoRewardCIHi}{0.930}
\newcommand{\moGoodExpPBUncOverallRewardCILo}{0.932}
\newcommand{\moGoodExpPBUncOverallRewardCIHi}{0.932}
\newcommand{\moGoodExpPBUncPhaseTwoCostCILo}{0.011181}
\newcommand{\moGoodExpPBUncPhaseTwoCostCIHi}{0.011830}
\newcommand{\moGoodExpPBUncOverallCostCILo}{0.011070}
\newcommand{\moGoodExpPBUncOverallCostCIHi}{0.011405}
\newcommand{\moGoodExpPBUncFlashPct}{13.3}
\newcommand{\moGoodExpPBUncFlashPctCILo}{12.1}
\newcommand{\moGoodExpPBUncFlashPctCIHi}{14.5}
\newcommand{\moGoodExpPBUncNSustained}{20}
\newcommand{\moGoodExpPBUncSustainedStep}{121}
\newcommand{\moGoodExpPBUncFlashFinalPct}{7.3}
\newcommand{\moGoodExpPBUncFlashFinalPctCILo}{6.2}
\newcommand{\moGoodExpPBUncFlashFinalPctCIHi}{8.3}
\newcommand{\moGoodExpLooseFlashFinalPct}{6.1}
\newcommand{\moGoodExpLooseNSustained}{20}
\newcommand{\moGoodExpUncFlashFinalPct}{7.3}
\newcommand{\moBadCheapUncOverallReward}{0.930}


\newcommand{\waNseeds}{20}
\newcommand{\waNPrompts}{1824}
\newcommand{\waEarlyStep}{200}
\newcommand{\waK}{3}
\newcommand{\waCatMultiplier}{2}
\newcommand{\waNTests}{4}
\newcommand{\waNDiffPairs}{4}
\newcommand{\waBootstrapResamples}{10{,}000}
\newcommand{\waCILevel}{95}
\newcommand{\waAlpha}{0.01}
\newcommand{\waNeff}{1164}
\newcommand{\waGamma}{0.997}
\newcommand{\waTRAlpha}{0.05}
\newcommand{\waTRGamma}{0.997}
\newcommand{\waTightBudget}{3{\times}10^{-4}}
\newcommand{\waModBudget}{6.6{\times}10^{-4}}
\newcommand{\waLooseBudget}{1.9{\times}10^{-3}}
\newcommand{\waUncWarmupRegret}{55.0}
\newcommand{\waUncWarmupRegretStd}{1.6}
\newcommand{\waUncWarmupRegretSE}{0.4}
\newcommand{\waUncWarmupReward}{0.933}
\newcommand{\waUncWarmupRAtTwoHundred}{6.2}
\newcommand{\waUncWarmupRegretCILo}{54.3}
\newcommand{\waUncWarmupRegretCIHi}{55.7}
\newcommand{\waUncWarmupRAtTwoHundredCILo}{5.8}
\newcommand{\waUncWarmupRAtTwoHundredCIHi}{6.6}
\newcommand{\waUncTRRegret}{87.0}
\newcommand{\waUncTRRegretStd}{24.8}
\newcommand{\waUncTRRegretSE}{5.6}
\newcommand{\waUncTRReward}{0.916}
\newcommand{\waUncTRRAtTwoHundred}{16.2}
\newcommand{\waUncTRRegretCILo}{77.1}
\newcommand{\waUncTRRegretCIHi}{99.0}
\newcommand{\waUncTRRAtTwoHundredCILo}{11.9}
\newcommand{\waUncTRRAtTwoHundredCIHi}{20.8}
\newcommand{\waUncRandomRegret}{146.5}
\newcommand{\waUncRandomRegretStd}{5.0}
\newcommand{\waUncRandomRegretSE}{1.1}
\newcommand{\waUncRandomReward}{0.883}
\newcommand{\waUncRandomRAtTwoHundred}{16.0}
\newcommand{\waUncRandomRegretCILo}{144.4}
\newcommand{\waUncRandomRegretCIHi}{148.8}
\newcommand{\waUncRandomRAtTwoHundredCILo}{15.1}
\newcommand{\waUncRandomRAtTwoHundredCIHi}{17.0}
\newcommand{\waUncRegretReductionPct}{37}
\newcommand{\waTightWarmupRegret}{161.2}
\newcommand{\waTightWarmupRegretStd}{4.6}
\newcommand{\waTightWarmupRegretSE}{1.0}
\newcommand{\waTightWarmupReward}{0.875}
\newcommand{\waTightWarmupRAtTwoHundred}{19.5}
\newcommand{\waTightWarmupRegretCILo}{159.1}
\newcommand{\waTightWarmupRegretCIHi}{163.1}
\newcommand{\waTightWarmupRAtTwoHundredCILo}{18.0}
\newcommand{\waTightWarmupRAtTwoHundredCIHi}{21.0}
\newcommand{\waTightTRRegret}{213.3}
\newcommand{\waTightTRRegretStd}{41.8}
\newcommand{\waTightTRRegretSE}{9.4}
\newcommand{\waTightTRReward}{0.846}
\newcommand{\waTightTRRAtTwoHundred}{29.2}
\newcommand{\waTightTRRegretCILo}{195.8}
\newcommand{\waTightTRRegretCIHi}{232.1}
\newcommand{\waTightTRRAtTwoHundredCILo}{26.4}
\newcommand{\waTightTRRAtTwoHundredCIHi}{31.9}
\newcommand{\waModWarmupRegret}{133.9}
\newcommand{\waModWarmupRegretStd}{11.0}
\newcommand{\waModWarmupRegretSE}{2.5}
\newcommand{\waModWarmupReward}{0.890}
\newcommand{\waModWarmupRAtTwoHundred}{14.7}
\newcommand{\waModWarmupRegretCILo}{129.2}
\newcommand{\waModWarmupRegretCIHi}{138.8}
\newcommand{\waModWarmupRAtTwoHundredCILo}{12.9}
\newcommand{\waModWarmupRAtTwoHundredCIHi}{16.6}
\newcommand{\waModTRRegret}{147.5}
\newcommand{\waModTRRegretStd}{40.5}
\newcommand{\waModTRRegretSE}{9.1}
\newcommand{\waModTRReward}{0.882}
\newcommand{\waModTRRAtTwoHundred}{23.5}
\newcommand{\waModTRRegretCILo}{130.0}
\newcommand{\waModTRRegretCIHi}{165.3}
\newcommand{\waModTRRAtTwoHundredCILo}{19.4}
\newcommand{\waModTRRAtTwoHundredCIHi}{27.5}
\newcommand{\waLooseWarmupRegret}{83.6}
\newcommand{\waLooseWarmupRegretStd}{5.1}
\newcommand{\waLooseWarmupRegretSE}{1.1}
\newcommand{\waLooseWarmupReward}{0.918}
\newcommand{\waLooseWarmupRAtTwoHundred}{8.6}
\newcommand{\waLooseWarmupRegretCILo}{81.4}
\newcommand{\waLooseWarmupRegretCIHi}{85.9}
\newcommand{\waLooseWarmupRAtTwoHundredCILo}{8.2}
\newcommand{\waLooseWarmupRAtTwoHundredCIHi}{9.1}
\newcommand{\waLooseTRRegret}{114.0}
\newcommand{\waLooseTRRegretStd}{30.8}
\newcommand{\waLooseTRRegretSE}{6.9}
\newcommand{\waLooseTRReward}{0.901}
\newcommand{\waLooseTRRAtTwoHundred}{22.2}
\newcommand{\waLooseTRRegretCILo}{101.1}
\newcommand{\waLooseTRRegretCIHi}{128.3}
\newcommand{\waLooseTRRAtTwoHundredCILo}{18.3}
\newcommand{\waLooseTRRAtTwoHundredCIHi}{26.0}
\newcommand{\waTightRegretReductionPct}{24}
\newcommand{\waModRegretReductionPct}{9}
\newcommand{\waLooseRegretReductionPct}{27}
\newcommand{\waUncVsTRWarmupWins}{20}
\newcommand{\waUncVsTRBaselineWins}{0}
\newcommand{\waUncVsTRTies}{0}
\newcommand{\waUncVsTRSignP}{{<}10^{-5}}
\newcommand{\waUncVsTRFisherP}{0.974}
\newcommand{\waUncVsTRWarmupCat}{0}
\newcommand{\waUncVsTRBaselineCat}{2}
\newcommand{\waUncVsTRBaselineCatRate}{10}
\newcommand{\waTightVsTRWarmupWins}{17}
\newcommand{\waTightVsTRBaselineWins}{3}
\newcommand{\waTightVsTRTies}{0}
\newcommand{\waTightVsTRSignP}{0.004}
\newcommand{\waTightVsTRFisherP}{1.000}
\newcommand{\waTightVsTRWarmupCat}{0}
\newcommand{\waTightVsTRBaselineCat}{0}
\newcommand{\waTightVsTRBaselineCatRate}{0}
\newcommand{\waModVsTRWarmupWins}{11}
\newcommand{\waModVsTRBaselineWins}{9}
\newcommand{\waModVsTRTies}{0}
\newcommand{\waModVsTRSignP}{0.412}
\newcommand{\waModVsTRFisherP}{1.000}
\newcommand{\waModVsTRWarmupCat}{0}
\newcommand{\waModVsTRBaselineCat}{0}
\newcommand{\waModVsTRBaselineCatRate}{0}
\newcommand{\waLooseVsTRWarmupWins}{17}
\newcommand{\waLooseVsTRBaselineWins}{3}
\newcommand{\waLooseVsTRTies}{0}
\newcommand{\waLooseVsTRSignP}{0.004}
\newcommand{\waLooseVsTRFisherP}{1.000}
\newcommand{\waLooseVsTRWarmupCat}{0}
\newcommand{\waLooseVsTRBaselineCat}{0}
\newcommand{\waLooseVsTRBaselineCatRate}{0}
\newcommand{\waUncVsTRDiffMean}{32.0}
\newcommand{\waUncVsTRDiffCILo}{19.8}
\newcommand{\waUncVsTRDiffCIHi}{48.4}
\newcommand{\waUncVsTRRAtTwoHundredDiffMean}{10.0}
\newcommand{\waUncVsTRRAtTwoHundredDiffCILo}{4.7}
\newcommand{\waUncVsTRRAtTwoHundredDiffCIHi}{15.8}
\newcommand{\waTightVsTRDiffMean}{52.2}
\newcommand{\waTightVsTRDiffCILo}{29.6}
\newcommand{\waTightVsTRDiffCIHi}{76.9}
\newcommand{\waTightVsTRRAtTwoHundredDiffMean}{9.7}
\newcommand{\waTightVsTRRAtTwoHundredDiffCILo}{5.5}
\newcommand{\waTightVsTRRAtTwoHundredDiffCIHi}{13.7}
\newcommand{\waModVsTRDiffMean}{13.7}
\newcommand{\waModVsTRDiffCILo}{-10.9}
\newcommand{\waModVsTRDiffCIHi}{38.5}
\newcommand{\waModVsTRRAtTwoHundredDiffMean}{8.8}
\newcommand{\waModVsTRRAtTwoHundredDiffCILo}{3.6}
\newcommand{\waModVsTRRAtTwoHundredDiffCIHi}{13.8}
\newcommand{\waLooseVsTRDiffMean}{30.4}
\newcommand{\waLooseVsTRDiffCILo}{13.9}
\newcommand{\waLooseVsTRDiffCIHi}{49.2}
\newcommand{\waLooseVsTRRAtTwoHundredDiffMean}{13.6}
\newcommand{\waLooseVsTRRAtTwoHundredDiffCILo}{8.2}
\newcommand{\waLooseVsTRRAtTwoHundredDiffCIHi}{18.8}
\newcommand{\waWarmupEffMem}{333}
\newcommand{\waTREffMem}{333}
\newcommand{\waTightStdRatio}{9}


\newcommand{\prmNseeds}{20}
\newcommand{\prmNPrompts}{1824}
\newcommand{\prmCatThreshold}{156.5}
\newcommand{\prmCatRefMedian}{78.2}
\newcommand{\prmBootstrapResamples}{10{,}000}
\newcommand{\prmCILevel}{95}
\newcommand{\prmAlpha}{0.01}
\newcommand{\prmGamma}{0.997}
\newcommand{\prmTRGamma}{0.997}
\newcommand{\prmSeedOffset}{9{,}000}
\newcommand{\prmNQualities}{5}
\newcommand{\prmNNeff}{3}
\newcommand{\prmNConditions}{16}
\newcommand{\prmNPairwise}{15}
\newcommand{\prmCatMultiplier}{2}
\newcommand{\prmFeatureDim}{26}
\newcommand{\prmWellCalNPrompts}{8{,}373}
\newcommand{\prmRandNPrompts}{1{,}680}
\newcommand{\prmMMLUNPrompts}{1{,}855}
\newcommand{\prmGSMNPrompts}{1{,}680}
\newcommand{\prmInvNPrompts}{8{,}374}
\newcommand{\prmGSMMinBiasPred}{0.86}
\newcommand{\prmEffMemWarmup}{333}
\newcommand{\prmEffMemTR}{333}
\newcommand{\prmEffMemDiffPct}{0}
\newcommand{\prmTRRegretMean}{87.0}
\newcommand{\prmTRRegretMedian}{78.2}
\newcommand{\prmTRRegretStd}{24.8}
\newcommand{\prmTRRegretSE}{5.6}
\newcommand{\prmTRRegretMeanCILo}{77.1}
\newcommand{\prmTRRegretMeanCIHi}{99.0}
\newcommand{\prmTRRegretMedianCILo}{70.7}
\newcommand{\prmTRRegretMedianCIHi}{90.1}
\newcommand{\prmTRReward}{0.916}
\newcommand{\prmTRRewardCILo}{0.909}
\newcommand{\prmTRRewardCIHi}{0.921}
\newcommand{\prmTRRAtTwoHundred}{16.2}
\newcommand{\prmTRRAtTwoHundredCILo}{11.9}
\newcommand{\prmTRRAtTwoHundredCIHi}{20.8}
\newcommand{\prmTRMatchedRegretMean}{117.7}
\newcommand{\prmTRMatchedRegretMedian}{73.4}
\newcommand{\prmTRMatchedRegretStd}{86.5}
\newcommand{\prmTRMatchedRegretSE}{19.3}
\newcommand{\prmTRMatchedRegretMeanCILo}{83.3}
\newcommand{\prmTRMatchedRegretMeanCIHi}{158.7}
\newcommand{\prmTRMatchedRegretMedianCILo}{66.2}
\newcommand{\prmTRMatchedRegretMedianCIHi}{125.2}
\newcommand{\prmTRMatchedReward}{0.899}
\newcommand{\prmTRMatchedRewardCILo}{0.876}
\newcommand{\prmTRMatchedRewardCIHi}{0.918}
\newcommand{\prmTRMatchedRAtTwoHundred}{17.9}
\newcommand{\prmTRMatchedRAtTwoHundredCILo}{12.9}
\newcommand{\prmTRMatchedRAtTwoHundredCIHi}{23.2}
\newcommand{\prmWellCalTenRegretMean}{65.4}
\newcommand{\prmWellCalTenRegretMedian}{65.2}
\newcommand{\prmWellCalTenRegretStd}{2.6}
\newcommand{\prmWellCalTenRegretSE}{0.6}
\newcommand{\prmWellCalTenRegretMeanCILo}{64.3}
\newcommand{\prmWellCalTenRegretMeanCIHi}{66.5}
\newcommand{\prmWellCalTenRegretMedianCILo}{63.6}
\newcommand{\prmWellCalTenRegretMedianCIHi}{67.3}
\newcommand{\prmWellCalTenReward}{0.927}
\newcommand{\prmWellCalTenRewardCILo}{0.927}
\newcommand{\prmWellCalTenRewardCIHi}{0.928}
\newcommand{\prmWellCalTenRAtTwoHundred}{7.8}
\newcommand{\prmWellCalTenRAtTwoHundredCILo}{7.1}
\newcommand{\prmWellCalTenRAtTwoHundredCIHi}{8.4}
\newcommand{\prmWellCalHundredRegretMean}{61.5}
\newcommand{\prmWellCalHundredRegretMedian}{61.9}
\newcommand{\prmWellCalHundredRegretStd}{2.5}
\newcommand{\prmWellCalHundredRegretSE}{0.6}
\newcommand{\prmWellCalHundredRegretMeanCILo}{60.4}
\newcommand{\prmWellCalHundredRegretMeanCIHi}{62.6}
\newcommand{\prmWellCalHundredRegretMedianCILo}{60.4}
\newcommand{\prmWellCalHundredRegretMedianCIHi}{63.5}
\newcommand{\prmWellCalHundredReward}{0.930}
\newcommand{\prmWellCalHundredRewardCILo}{0.929}
\newcommand{\prmWellCalHundredRewardCIHi}{0.930}
\newcommand{\prmWellCalHundredRAtTwoHundred}{6.6}
\newcommand{\prmWellCalHundredRAtTwoHundredCILo}{6.2}
\newcommand{\prmWellCalHundredRAtTwoHundredCIHi}{7.1}
\newcommand{\prmWellCalThousandRegretMean}{55.6}
\newcommand{\prmWellCalThousandRegretMedian}{55.5}
\newcommand{\prmWellCalThousandRegretStd}{1.6}
\newcommand{\prmWellCalThousandRegretSE}{0.4}
\newcommand{\prmWellCalThousandRegretMeanCILo}{54.9}
\newcommand{\prmWellCalThousandRegretMeanCIHi}{56.3}
\newcommand{\prmWellCalThousandRegretMedianCILo}{54.5}
\newcommand{\prmWellCalThousandRegretMedianCIHi}{56.2}
\newcommand{\prmWellCalThousandReward}{0.933}
\newcommand{\prmWellCalThousandRewardCILo}{0.932}
\newcommand{\prmWellCalThousandRewardCIHi}{0.933}
\newcommand{\prmWellCalThousandRAtTwoHundred}{6.2}
\newcommand{\prmWellCalThousandRAtTwoHundredCILo}{5.8}
\newcommand{\prmWellCalThousandRAtTwoHundredCIHi}{6.6}
\newcommand{\prmRandTenRegretMean}{63.6}
\newcommand{\prmRandTenRegretMedian}{64.2}
\newcommand{\prmRandTenRegretStd}{3.0}
\newcommand{\prmRandTenRegretSE}{0.7}
\newcommand{\prmRandTenRegretMeanCILo}{62.3}
\newcommand{\prmRandTenRegretMeanCIHi}{64.9}
\newcommand{\prmRandTenRegretMedianCILo}{61.9}
\newcommand{\prmRandTenRegretMedianCIHi}{65.4}
\newcommand{\prmRandTenReward}{0.928}
\newcommand{\prmRandTenRewardCILo}{0.928}
\newcommand{\prmRandTenRewardCIHi}{0.929}
\newcommand{\prmRandTenRAtTwoHundred}{7.4}
\newcommand{\prmRandTenRAtTwoHundredCILo}{6.6}
\newcommand{\prmRandTenRAtTwoHundredCIHi}{8.2}
\newcommand{\prmRandHundredRegretMean}{60.9}
\newcommand{\prmRandHundredRegretMedian}{61.2}
\newcommand{\prmRandHundredRegretStd}{2.3}
\newcommand{\prmRandHundredRegretSE}{0.5}
\newcommand{\prmRandHundredRegretMeanCILo}{59.9}
\newcommand{\prmRandHundredRegretMeanCIHi}{61.9}
\newcommand{\prmRandHundredRegretMedianCILo}{59.4}
\newcommand{\prmRandHundredRegretMedianCIHi}{62.1}
\newcommand{\prmRandHundredReward}{0.930}
\newcommand{\prmRandHundredRewardCILo}{0.929}
\newcommand{\prmRandHundredRewardCIHi}{0.930}
\newcommand{\prmRandHundredRAtTwoHundred}{6.4}
\newcommand{\prmRandHundredRAtTwoHundredCILo}{5.9}
\newcommand{\prmRandHundredRAtTwoHundredCIHi}{6.9}
\newcommand{\prmRandThousandRegretMean}{55.2}
\newcommand{\prmRandThousandRegretMedian}{55.1}
\newcommand{\prmRandThousandRegretStd}{2.0}
\newcommand{\prmRandThousandRegretSE}{0.5}
\newcommand{\prmRandThousandRegretMeanCILo}{54.3}
\newcommand{\prmRandThousandRegretMeanCIHi}{56.1}
\newcommand{\prmRandThousandRegretMedianCILo}{54.4}
\newcommand{\prmRandThousandRegretMedianCIHi}{55.9}
\newcommand{\prmRandThousandReward}{0.933}
\newcommand{\prmRandThousandRewardCILo}{0.933}
\newcommand{\prmRandThousandRewardCIHi}{0.934}
\newcommand{\prmRandThousandRAtTwoHundred}{5.9}
\newcommand{\prmRandThousandRAtTwoHundredCILo}{5.5}
\newcommand{\prmRandThousandRAtTwoHundredCIHi}{6.3}
\newcommand{\prmMMLUTenRegretMean}{65.2}
\newcommand{\prmMMLUTenRegretMedian}{65.4}
\newcommand{\prmMMLUTenRegretStd}{2.3}
\newcommand{\prmMMLUTenRegretSE}{0.5}
\newcommand{\prmMMLUTenRegretMeanCILo}{64.2}
\newcommand{\prmMMLUTenRegretMeanCIHi}{66.2}
\newcommand{\prmMMLUTenRegretMedianCILo}{63.9}
\newcommand{\prmMMLUTenRegretMedianCIHi}{66.4}
\newcommand{\prmMMLUTenReward}{0.928}
\newcommand{\prmMMLUTenRewardCILo}{0.927}
\newcommand{\prmMMLUTenRewardCIHi}{0.928}
\newcommand{\prmMMLUTenRAtTwoHundred}{8.0}
\newcommand{\prmMMLUTenRAtTwoHundredCILo}{7.5}
\newcommand{\prmMMLUTenRAtTwoHundredCIHi}{8.6}
\newcommand{\prmMMLUHundredRegretMean}{59.4}
\newcommand{\prmMMLUHundredRegretMedian}{60.1}
\newcommand{\prmMMLUHundredRegretStd}{1.9}
\newcommand{\prmMMLUHundredRegretSE}{0.4}
\newcommand{\prmMMLUHundredRegretMeanCILo}{58.6}
\newcommand{\prmMMLUHundredRegretMeanCIHi}{60.2}
\newcommand{\prmMMLUHundredRegretMedianCILo}{58.1}
\newcommand{\prmMMLUHundredRegretMedianCIHi}{60.5}
\newcommand{\prmMMLUHundredReward}{0.931}
\newcommand{\prmMMLUHundredRewardCILo}{0.930}
\newcommand{\prmMMLUHundredRewardCIHi}{0.931}
\newcommand{\prmMMLUHundredRAtTwoHundred}{7.3}
\newcommand{\prmMMLUHundredRAtTwoHundredCILo}{6.8}
\newcommand{\prmMMLUHundredRAtTwoHundredCIHi}{7.9}
\newcommand{\prmMMLUThousandRegretMean}{58.2}
\newcommand{\prmMMLUThousandRegretMedian}{58.3}
\newcommand{\prmMMLUThousandRegretStd}{1.7}
\newcommand{\prmMMLUThousandRegretSE}{0.4}
\newcommand{\prmMMLUThousandRegretMeanCILo}{57.5}
\newcommand{\prmMMLUThousandRegretMeanCIHi}{58.9}
\newcommand{\prmMMLUThousandRegretMedianCILo}{57.5}
\newcommand{\prmMMLUThousandRegretMedianCIHi}{58.7}
\newcommand{\prmMMLUThousandReward}{0.931}
\newcommand{\prmMMLUThousandRewardCILo}{0.931}
\newcommand{\prmMMLUThousandRewardCIHi}{0.932}
\newcommand{\prmMMLUThousandRAtTwoHundred}{6.6}
\newcommand{\prmMMLUThousandRAtTwoHundredCILo}{6.2}
\newcommand{\prmMMLUThousandRAtTwoHundredCIHi}{7.0}
\newcommand{\prmGSMTenRegretMean}{66.8}
\newcommand{\prmGSMTenRegretMedian}{66.7}
\newcommand{\prmGSMTenRegretStd}{3.2}
\newcommand{\prmGSMTenRegretSE}{0.7}
\newcommand{\prmGSMTenRegretMeanCILo}{65.4}
\newcommand{\prmGSMTenRegretMeanCIHi}{68.2}
\newcommand{\prmGSMTenRegretMedianCILo}{65.7}
\newcommand{\prmGSMTenRegretMedianCIHi}{68.0}
\newcommand{\prmGSMTenReward}{0.927}
\newcommand{\prmGSMTenRewardCILo}{0.926}
\newcommand{\prmGSMTenRewardCIHi}{0.927}
\newcommand{\prmGSMTenRAtTwoHundred}{9.6}
\newcommand{\prmGSMTenRAtTwoHundredCILo}{8.8}
\newcommand{\prmGSMTenRAtTwoHundredCIHi}{10.5}
\newcommand{\prmGSMHundredRegretMean}{64.5}
\newcommand{\prmGSMHundredRegretMedian}{63.9}
\newcommand{\prmGSMHundredRegretStd}{2.7}
\newcommand{\prmGSMHundredRegretSE}{0.6}
\newcommand{\prmGSMHundredRegretMeanCILo}{63.4}
\newcommand{\prmGSMHundredRegretMeanCIHi}{65.8}
\newcommand{\prmGSMHundredRegretMedianCILo}{63.0}
\newcommand{\prmGSMHundredRegretMedianCIHi}{65.0}
\newcommand{\prmGSMHundredReward}{0.928}
\newcommand{\prmGSMHundredRewardCILo}{0.927}
\newcommand{\prmGSMHundredRewardCIHi}{0.929}
\newcommand{\prmGSMHundredRAtTwoHundred}{8.5}
\newcommand{\prmGSMHundredRAtTwoHundredCILo}{7.9}
\newcommand{\prmGSMHundredRAtTwoHundredCIHi}{9.1}
\newcommand{\prmGSMThousandRegretMean}{61.4}
\newcommand{\prmGSMThousandRegretMedian}{60.9}
\newcommand{\prmGSMThousandRegretStd}{2.8}
\newcommand{\prmGSMThousandRegretSE}{0.6}
\newcommand{\prmGSMThousandRegretMeanCILo}{60.2}
\newcommand{\prmGSMThousandRegretMeanCIHi}{62.7}
\newcommand{\prmGSMThousandRegretMedianCILo}{59.4}
\newcommand{\prmGSMThousandRegretMedianCIHi}{62.9}
\newcommand{\prmGSMThousandReward}{0.930}
\newcommand{\prmGSMThousandRewardCILo}{0.929}
\newcommand{\prmGSMThousandRewardCIHi}{0.930}
\newcommand{\prmGSMThousandRAtTwoHundred}{8.2}
\newcommand{\prmGSMThousandRAtTwoHundredCILo}{7.7}
\newcommand{\prmGSMThousandRAtTwoHundredCIHi}{8.8}
\newcommand{\prmInvTenRegretMean}{79.1}
\newcommand{\prmInvTenRegretMedian}{78.2}
\newcommand{\prmInvTenRegretStd}{4.9}
\newcommand{\prmInvTenRegretSE}{1.1}
\newcommand{\prmInvTenRegretMeanCILo}{76.9}
\newcommand{\prmInvTenRegretMeanCIHi}{81.2}
\newcommand{\prmInvTenRegretMedianCILo}{76.3}
\newcommand{\prmInvTenRegretMedianCIHi}{81.8}
\newcommand{\prmInvTenReward}{0.920}
\newcommand{\prmInvTenRewardCILo}{0.919}
\newcommand{\prmInvTenRewardCIHi}{0.921}
\newcommand{\prmInvTenRAtTwoHundred}{13.3}
\newcommand{\prmInvTenRAtTwoHundredCILo}{12.7}
\newcommand{\prmInvTenRAtTwoHundredCIHi}{13.9}
\newcommand{\prmInvHundredRegretMean}{90.1}
\newcommand{\prmInvHundredRegretMedian}{89.9}
\newcommand{\prmInvHundredRegretStd}{1.6}
\newcommand{\prmInvHundredRegretSE}{0.4}
\newcommand{\prmInvHundredRegretMeanCILo}{89.4}
\newcommand{\prmInvHundredRegretMeanCIHi}{90.8}
\newcommand{\prmInvHundredRegretMedianCILo}{89.1}
\newcommand{\prmInvHundredRegretMedianCIHi}{91.3}
\newcommand{\prmInvHundredReward}{0.914}
\newcommand{\prmInvHundredRewardCILo}{0.914}
\newcommand{\prmInvHundredRewardCIHi}{0.914}
\newcommand{\prmInvHundredRAtTwoHundred}{15.8}
\newcommand{\prmInvHundredRAtTwoHundredCILo}{15.2}
\newcommand{\prmInvHundredRAtTwoHundredCIHi}{16.5}
\newcommand{\prmInvThousandRegretMean}{106.8}
\newcommand{\prmInvThousandRegretMedian}{106.9}
\newcommand{\prmInvThousandRegretStd}{2.2}
\newcommand{\prmInvThousandRegretSE}{0.5}
\newcommand{\prmInvThousandRegretMeanCILo}{105.9}
\newcommand{\prmInvThousandRegretMeanCIHi}{107.7}
\newcommand{\prmInvThousandRegretMedianCILo}{105.6}
\newcommand{\prmInvThousandRegretMedianCIHi}{108.0}
\newcommand{\prmInvThousandReward}{0.905}
\newcommand{\prmInvThousandRewardCILo}{0.904}
\newcommand{\prmInvThousandRewardCIHi}{0.905}
\newcommand{\prmInvThousandRAtTwoHundred}{24.1}
\newcommand{\prmInvThousandRAtTwoHundredCILo}{23.3}
\newcommand{\prmInvThousandRAtTwoHundredCIHi}{24.9}
\newcommand{\prmWellCalTenReductionPct}{16.6}
\newcommand{\prmWellCalTenReductionPctCILo}{6.5}
\newcommand{\prmWellCalTenReductionPctCIHi}{28.5}
\newcommand{\prmWellCalHundredReductionPct}{20.9}
\newcommand{\prmWellCalHundredReductionPctCILo}{11.9}
\newcommand{\prmWellCalHundredReductionPctCIHi}{32.1}
\newcommand{\prmWellCalThousandReductionPct}{29.1}
\newcommand{\prmWellCalThousandReductionPctCILo}{21.7}
\newcommand{\prmWellCalThousandReductionPctCIHi}{38.8}
\newcommand{\prmRandTenReductionPct}{17.9}
\newcommand{\prmRandTenReductionPctCILo}{9.0}
\newcommand{\prmRandTenReductionPctCIHi}{29.8}
\newcommand{\prmRandHundredReductionPct}{21.8}
\newcommand{\prmRandHundredReductionPctCILo}{13.9}
\newcommand{\prmRandHundredReductionPctCIHi}{32.5}
\newcommand{\prmRandThousandReductionPct}{29.6}
\newcommand{\prmRandThousandReductionPctCILo}{21.6}
\newcommand{\prmRandThousandReductionPctCIHi}{39.5}
\newcommand{\prmMMLUTenReductionPct}{16.4}
\newcommand{\prmMMLUTenReductionPctCILo}{7.4}
\newcommand{\prmMMLUTenReductionPctCIHi}{28.4}
\newcommand{\prmMMLUHundredReductionPct}{23.2}
\newcommand{\prmMMLUHundredReductionPctCILo}{14.8}
\newcommand{\prmMMLUHundredReductionPctCIHi}{35.3}
\newcommand{\prmMMLUThousandReductionPct}{25.4}
\newcommand{\prmMMLUThousandReductionPctCILo}{17.5}
\newcommand{\prmMMLUThousandReductionPctCIHi}{35.9}
\newcommand{\prmGSMTenReductionPct}{14.8}
\newcommand{\prmGSMTenReductionPctCILo}{5.5}
\newcommand{\prmGSMTenReductionPctCIHi}{26.0}
\newcommand{\prmGSMHundredReductionPct}{18.3}
\newcommand{\prmGSMHundredReductionPctCILo}{9.2}
\newcommand{\prmGSMHundredReductionPctCIHi}{29.5}
\newcommand{\prmGSMThousandReductionPct}{22.2}
\newcommand{\prmGSMThousandReductionPctCILo}{12.1}
\newcommand{\prmGSMThousandReductionPctCIHi}{32.7}
\newcommand{\prmInvTenReductionPct}{0.1}
\newcommand{\prmInvTenReductionPctCILo}{-11.9}
\newcommand{\prmInvTenReductionPctCIHi}{12.2}
\newcommand{\prmInvHundredReductionPct}{-14.9}
\newcommand{\prmInvHundredReductionPctCILo}{-27.2}
\newcommand{\prmInvHundredReductionPctCIHi}{0.7}
\newcommand{\prmInvThousandReductionPct}{-36.6}
\newcommand{\prmInvThousandReductionPctCILo}{-51.4}
\newcommand{\prmInvThousandReductionPctCIHi}{-18.1}
\newcommand{\prmInvThousandIncreasePct}{37}
\newcommand{\prmTestWellCalTenCondWins}{17}
\newcommand{\prmTestWellCalTenBaseWins}{3}
\newcommand{\prmTestWellCalTenDeltaMean}{-21.6}
\newcommand{\prmTestWellCalTenDeltaMedian}{-13.1}
\newcommand{\prmTestWellCalTenCondMedian}{65.2}
\newcommand{\prmTestWellCalTenSignPHolm}{0.0103}
\newcommand{\prmTestWellCalTenFisherPHolm}{1.00}
\newcommand{\prmTestWellCalTenCondCat}{0}
\newcommand{\prmTestWellCalTenBaseCat}{1}
\newcommand{\prmTestWellCalHundredCondWins}{19}
\newcommand{\prmTestWellCalHundredBaseWins}{1}
\newcommand{\prmTestWellCalHundredDeltaMean}{-25.5}
\newcommand{\prmTestWellCalHundredDeltaMedian}{-18.3}
\newcommand{\prmTestWellCalHundredCondMedian}{61.9}
\newcommand{\prmTestWellCalHundredSignPHolm}{0.0004}
\newcommand{\prmTestWellCalHundredFisherPHolm}{1.00}
\newcommand{\prmTestWellCalHundredCondCat}{0}
\newcommand{\prmTestWellCalHundredBaseCat}{1}
\newcommand{\prmTestWellCalThousandCondWins}{20}
\newcommand{\prmTestWellCalThousandBaseWins}{0}
\newcommand{\prmTestWellCalThousandDeltaMean}{-31.4}
\newcommand{\prmTestWellCalThousandDeltaMedian}{-22.6}
\newcommand{\prmTestWellCalThousandCondMedian}{55.5}
\newcommand{\prmTestWellCalThousandSignPHolm}{{<}10^{-4}}
\newcommand{\prmTestWellCalThousandFisherPHolm}{1.00}
\newcommand{\prmTestWellCalThousandCondCat}{0}
\newcommand{\prmTestWellCalThousandBaseCat}{1}
\newcommand{\prmTestRandTenCondWins}{18}
\newcommand{\prmTestRandTenBaseWins}{2}
\newcommand{\prmTestRandTenDeltaMean}{-23.3}
\newcommand{\prmTestRandTenDeltaMedian}{-15.4}
\newcommand{\prmTestRandTenCondMedian}{64.2}
\newcommand{\prmTestRandTenSignPHolm}{0.0032}
\newcommand{\prmTestRandTenFisherPHolm}{1.00}
\newcommand{\prmTestRandTenCondCat}{0}
\newcommand{\prmTestRandTenBaseCat}{1}
\newcommand{\prmTestRandHundredCondWins}{20}
\newcommand{\prmTestRandHundredBaseWins}{0}
\newcommand{\prmTestRandHundredDeltaMean}{-26.0}
\newcommand{\prmTestRandHundredDeltaMedian}{-18.7}
\newcommand{\prmTestRandHundredCondMedian}{61.2}
\newcommand{\prmTestRandHundredSignPHolm}{{<}10^{-4}}
\newcommand{\prmTestRandHundredFisherPHolm}{1.00}
\newcommand{\prmTestRandHundredCondCat}{0}
\newcommand{\prmTestRandHundredBaseCat}{1}
\newcommand{\prmTestRandThousandCondWins}{20}
\newcommand{\prmTestRandThousandBaseWins}{0}
\newcommand{\prmTestRandThousandDeltaMean}{-31.8}
\newcommand{\prmTestRandThousandDeltaMedian}{-22.5}
\newcommand{\prmTestRandThousandCondMedian}{55.1}
\newcommand{\prmTestRandThousandSignPHolm}{{<}10^{-4}}
\newcommand{\prmTestRandThousandFisherPHolm}{1.00}
\newcommand{\prmTestRandThousandCondCat}{0}
\newcommand{\prmTestRandThousandBaseCat}{1}
\newcommand{\prmTestMMLUTenCondWins}{18}
\newcommand{\prmTestMMLUTenBaseWins}{2}
\newcommand{\prmTestMMLUTenDeltaMean}{-21.8}
\newcommand{\prmTestMMLUTenDeltaMedian}{-12.9}
\newcommand{\prmTestMMLUTenCondMedian}{65.4}
\newcommand{\prmTestMMLUTenSignPHolm}{0.0032}
\newcommand{\prmTestMMLUTenFisherPHolm}{1.00}
\newcommand{\prmTestMMLUTenCondCat}{0}
\newcommand{\prmTestMMLUTenBaseCat}{1}
\newcommand{\prmTestMMLUHundredCondWins}{20}
\newcommand{\prmTestMMLUHundredBaseWins}{0}
\newcommand{\prmTestMMLUHundredDeltaMean}{-27.6}
\newcommand{\prmTestMMLUHundredDeltaMedian}{-19.4}
\newcommand{\prmTestMMLUHundredCondMedian}{60.1}
\newcommand{\prmTestMMLUHundredSignPHolm}{{<}10^{-4}}
\newcommand{\prmTestMMLUHundredFisherPHolm}{1.00}
\newcommand{\prmTestMMLUHundredCondCat}{0}
\newcommand{\prmTestMMLUHundredBaseCat}{1}
\newcommand{\prmTestMMLUThousandCondWins}{20}
\newcommand{\prmTestMMLUThousandBaseWins}{0}
\newcommand{\prmTestMMLUThousandDeltaMean}{-28.8}
\newcommand{\prmTestMMLUThousandDeltaMedian}{-20.7}
\newcommand{\prmTestMMLUThousandCondMedian}{58.3}
\newcommand{\prmTestMMLUThousandSignPHolm}{{<}10^{-4}}
\newcommand{\prmTestMMLUThousandFisherPHolm}{1.00}
\newcommand{\prmTestMMLUThousandCondCat}{0}
\newcommand{\prmTestMMLUThousandBaseCat}{1}
\newcommand{\prmTestGSMTenCondWins}{18}
\newcommand{\prmTestGSMTenBaseWins}{2}
\newcommand{\prmTestGSMTenDeltaMean}{-20.2}
\newcommand{\prmTestGSMTenDeltaMedian}{-13.0}
\newcommand{\prmTestGSMTenCondMedian}{66.7}
\newcommand{\prmTestGSMTenSignPHolm}{0.0032}
\newcommand{\prmTestGSMTenFisherPHolm}{1.00}
\newcommand{\prmTestGSMTenCondCat}{0}
\newcommand{\prmTestGSMTenBaseCat}{1}
\newcommand{\prmTestGSMHundredCondWins}{18}
\newcommand{\prmTestGSMHundredBaseWins}{2}
\newcommand{\prmTestGSMHundredDeltaMean}{-22.5}
\newcommand{\prmTestGSMHundredDeltaMedian}{-14.8}
\newcommand{\prmTestGSMHundredCondMedian}{63.9}
\newcommand{\prmTestGSMHundredSignPHolm}{0.0032}
\newcommand{\prmTestGSMHundredFisherPHolm}{1.00}
\newcommand{\prmTestGSMHundredCondCat}{0}
\newcommand{\prmTestGSMHundredBaseCat}{1}
\newcommand{\prmTestGSMThousandCondWins}{20}
\newcommand{\prmTestGSMThousandBaseWins}{0}
\newcommand{\prmTestGSMThousandDeltaMean}{-25.6}
\newcommand{\prmTestGSMThousandDeltaMedian}{-19.2}
\newcommand{\prmTestGSMThousandCondMedian}{60.9}
\newcommand{\prmTestGSMThousandSignPHolm}{{<}10^{-4}}
\newcommand{\prmTestGSMThousandFisherPHolm}{1.00}
\newcommand{\prmTestGSMThousandCondCat}{0}
\newcommand{\prmTestGSMThousandBaseCat}{1}
\newcommand{\prmTestInvTenCondWins}{12}
\newcommand{\prmTestInvTenBaseWins}{8}
\newcommand{\prmTestInvTenDeltaMean}{-7.9}
\newcommand{\prmTestInvTenDeltaMedian}{-2.6}
\newcommand{\prmTestInvTenCondMedian}{78.2}
\newcommand{\prmTestInvTenSignPHolm}{0.5034}
\newcommand{\prmTestInvTenFisherPHolm}{1.00}
\newcommand{\prmTestInvTenCondCat}{0}
\newcommand{\prmTestInvTenBaseCat}{1}
\newcommand{\prmTestInvHundredCondWins}{6}
\newcommand{\prmTestInvHundredBaseWins}{14}
\newcommand{\prmTestInvHundredDeltaMean}{3.1}
\newcommand{\prmTestInvHundredDeltaMedian}{13.4}
\newcommand{\prmTestInvHundredCondMedian}{89.9}
\newcommand{\prmTestInvHundredSignPHolm}{0.2306}
\newcommand{\prmTestInvHundredFisherPHolm}{1.00}
\newcommand{\prmTestInvHundredCondCat}{0}
\newcommand{\prmTestInvHundredBaseCat}{1}
\newcommand{\prmTestInvThousandCondWins}{3}
\newcommand{\prmTestInvThousandBaseWins}{17}
\newcommand{\prmTestInvThousandDeltaMean}{19.8}
\newcommand{\prmTestInvThousandDeltaMedian}{28.5}
\newcommand{\prmTestInvThousandCondMedian}{106.9}
\newcommand{\prmTestInvThousandSignPHolm}{0.0103}
\newcommand{\prmTestInvThousandFisherPHolm}{1.00}
\newcommand{\prmTestInvThousandCondCat}{0}
\newcommand{\prmTestInvThousandBaseCat}{1}
\newcommand{\prmTRCatCount}{0}
\newcommand{\prmTRCatPct}{0}
\newcommand{\prmNonInvStdMin}{1.6}
\newcommand{\prmNonInvStdMax}{3.2}
\newcommand{\prmInvThousandExcessRegret}{29}
\newcommand{\prmInvThousandExcessRegretCILo}{16.3}
\newcommand{\prmInvThousandExcessRegretCIHi}{36.4}
\newcommand{\prmInvTenDamagePct}{0}
\newcommand{\prmDomainMinPHolm}{{<}10^{-4}}
\newcommand{\prmWaBudgetCatPct}{0}
\newcommand{\prmWaBudgetCatCount}{0}
\newcommand{\prmWaBudgetCatSeeds}{40}
\newcommand{\prmWaBudgetFisherP}{1.000}


\newcommand{\rlNseeds}{20}
\newcommand{\rlPhaseN}{608}
\newcommand{\rlExtPhaseThreeN}{1216}
\newcommand{\rlBudgetTarget}{\$6.6\ensuremath{{\times}10^{-4}}}
\newcommand{\rlGamma}{0.997}
\newcommand{\rlHalfLife}{231}
\newcommand{\rlMistralNormalReward}{0.92}
\newcommand{\rlStdRecoveryThresholdPct}{17}
\newcommand{\rlStdFloorPct}{90}
\newcommand{\rlMildDegPct}{4.0}
\newcommand{\rlMildRecoveryRatio}{100.0}
\newcommand{\rlStdDegFourRecovery}{100.0}
\newcommand{\rlStdDegFourRecoveryCILo}{99.0}
\newcommand{\rlStdDegFourRecoveryCIHi}{101.0}
\newcommand{\rlStdFREightFiveRecovery}{100.0}
\newcommand{\rlStdFREightFiveDegPct}{4}
\newcommand{\rlStdFREightFiveRecoveryCILo}{99.0}
\newcommand{\rlStdFREightFiveRecoveryCIHi}{101.0}
\newcommand{\rlStdDegOneZeroRecovery}{98.4}
\newcommand{\rlStdDegOneZeroRecoveryCILo}{97.7}
\newcommand{\rlStdDegOneZeroRecoveryCIHi}{99.1}
\newcommand{\rlStdFREightZeroRecovery}{98.4}
\newcommand{\rlStdFREightZeroDegPct}{10}
\newcommand{\rlStdFREightZeroRecoveryCILo}{97.7}
\newcommand{\rlStdFREightZeroRecoveryCIHi}{99.1}
\newcommand{\rlStdDegOneFiveRecovery}{97.5}
\newcommand{\rlStdDegOneFiveRecoveryCILo}{96.7}
\newcommand{\rlStdDegOneFiveRecoveryCIHi}{98.3}
\newcommand{\rlStdFRSevenFiveRecovery}{97.5}
\newcommand{\rlStdFRSevenFiveDegPct}{15}
\newcommand{\rlStdFRSevenFiveRecoveryCILo}{96.7}
\newcommand{\rlStdFRSevenFiveRecoveryCIHi}{98.3}
\newcommand{\rlStdDegTwoOneRecovery}{95.4}
\newcommand{\rlStdDegTwoOneRecoveryCILo}{94.6}
\newcommand{\rlStdDegTwoOneRecoveryCIHi}{96.1}
\newcommand{\rlStdFRSevenZeroRecovery}{95.4}
\newcommand{\rlStdFRSevenZeroDegPct}{21}
\newcommand{\rlStdFRSevenZeroRecoveryCILo}{94.6}
\newcommand{\rlStdFRSevenZeroRecoveryCIHi}{96.1}
\newcommand{\rlStdDegThreeTwoRecovery}{93.2}
\newcommand{\rlStdDegThreeTwoRecoveryCILo}{92.3}
\newcommand{\rlStdDegThreeTwoRecoveryCIHi}{94.2}
\newcommand{\rlStdFRSixZeroRecovery}{93.2}
\newcommand{\rlStdFRSixZeroDegPct}{32}
\newcommand{\rlStdFRSixZeroRecoveryCILo}{92.3}
\newcommand{\rlStdFRSixZeroRecoveryCIHi}{94.2}
\newcommand{\rlStdDegFourFourRecovery}{91.6}
\newcommand{\rlStdDegFourFourRecoveryCILo}{90.8}
\newcommand{\rlStdDegFourFourRecoveryCIHi}{92.5}
\newcommand{\rlStdFRFiveZeroRecovery}{91.6}
\newcommand{\rlStdFRFiveZeroDegPct}{44}
\newcommand{\rlStdFRFiveZeroRecoveryCILo}{90.8}
\newcommand{\rlStdFRFiveZeroRecoveryCIHi}{92.5}
\newcommand{\rlStdDegFiveFiveRecovery}{90.7}
\newcommand{\rlStdDegFiveFiveRecoveryCILo}{89.8}
\newcommand{\rlStdDegFiveFiveRecoveryCIHi}{91.6}
\newcommand{\rlStdFRFourZeroRecovery}{90.7}
\newcommand{\rlStdFRFourZeroDegPct}{55}
\newcommand{\rlStdFRFourZeroRecoveryCILo}{89.8}
\newcommand{\rlStdFRFourZeroRecoveryCIHi}{91.6}
\newcommand{\rlStdDegSixSixRecovery}{90.2}
\newcommand{\rlStdDegSixSixRecoveryCILo}{89.6}
\newcommand{\rlStdDegSixSixRecoveryCIHi}{90.9}
\newcommand{\rlStdFRThreeZeroRecovery}{90.2}
\newcommand{\rlStdFRThreeZeroDegPct}{66}
\newcommand{\rlStdFRThreeZeroRecoveryCILo}{89.6}
\newcommand{\rlStdFRThreeZeroRecoveryCIHi}{90.9}
\newcommand{\rlStdDegSevenSevenRecovery}{90.2}
\newcommand{\rlStdDegSevenSevenRecoveryCILo}{89.5}
\newcommand{\rlStdDegSevenSevenRecoveryCIHi}{91.0}
\newcommand{\rlStdFRTwoZeroRecovery}{90.2}
\newcommand{\rlStdFRTwoZeroDegPct}{77}
\newcommand{\rlStdFRTwoZeroRecoveryCILo}{89.5}
\newcommand{\rlStdFRTwoZeroRecoveryCIHi}{91.0}
\newcommand{\rlStdDegEightNineRecovery}{89.8}
\newcommand{\rlStdDegEightNineRecoveryCILo}{89.3}
\newcommand{\rlStdDegEightNineRecoveryCIHi}{90.4}
\newcommand{\rlStdFROneZeroRecovery}{89.8}
\newcommand{\rlStdFROneZeroDegPct}{89}
\newcommand{\rlStdFROneZeroRecoveryCILo}{89.3}
\newcommand{\rlStdFROneZeroRecoveryCIHi}{90.4}
\newcommand{\rlStdDegNineFourRecovery}{89.8}
\newcommand{\rlStdDegNineFourRecoveryCILo}{89.1}
\newcommand{\rlStdDegNineFourRecoveryCIHi}{90.7}
\newcommand{\rlStdFRFiveRecovery}{89.8}
\newcommand{\rlStdFRFiveDegPct}{94}
\newcommand{\rlStdFRFiveRecoveryCILo}{89.1}
\newcommand{\rlStdFRFiveRecoveryCIHi}{90.7}
\newcommand{\rlExtRecoveryThresholdPct}{30}
\newcommand{\rlExtFloorPct}{93}
\newcommand{\rlExtDegFourRecovery}{100.8}
\newcommand{\rlExtDegFourRecoveryCILo}{100.2}
\newcommand{\rlExtDegFourRecoveryCIHi}{101.5}
\newcommand{\rlExtFREightFiveRecovery}{100.8}
\newcommand{\rlExtFREightFiveRecoveryCILo}{100.2}
\newcommand{\rlExtFREightFiveRecoveryCIHi}{101.5}
\newcommand{\rlExtDegOneZeroRecovery}{99.9}
\newcommand{\rlExtDegOneZeroRecoveryCILo}{99.2}
\newcommand{\rlExtDegOneZeroRecoveryCIHi}{100.6}
\newcommand{\rlExtFREightZeroRecovery}{99.9}
\newcommand{\rlExtFREightZeroRecoveryCILo}{99.2}
\newcommand{\rlExtFREightZeroRecoveryCIHi}{100.6}
\newcommand{\rlExtDegOneFiveRecovery}{99.5}
\newcommand{\rlExtDegOneFiveRecoveryCILo}{98.8}
\newcommand{\rlExtDegOneFiveRecoveryCIHi}{100.2}
\newcommand{\rlExtFRSevenFiveRecovery}{99.5}
\newcommand{\rlExtFRSevenFiveRecoveryCILo}{98.8}
\newcommand{\rlExtFRSevenFiveRecoveryCIHi}{100.2}
\newcommand{\rlExtDegTwoOneRecovery}{98.3}
\newcommand{\rlExtDegTwoOneRecoveryCILo}{97.6}
\newcommand{\rlExtDegTwoOneRecoveryCIHi}{99.0}
\newcommand{\rlExtFRSevenZeroRecovery}{98.3}
\newcommand{\rlExtFRSevenZeroRecoveryCILo}{97.6}
\newcommand{\rlExtFRSevenZeroRecoveryCIHi}{99.0}
\newcommand{\rlExtDegThreeTwoRecovery}{96.7}
\newcommand{\rlExtDegThreeTwoRecoveryCILo}{95.8}
\newcommand{\rlExtDegThreeTwoRecoveryCIHi}{97.6}
\newcommand{\rlExtFRSixZeroRecovery}{96.7}
\newcommand{\rlExtFRSixZeroRecoveryCILo}{95.8}
\newcommand{\rlExtFRSixZeroRecoveryCIHi}{97.6}
\newcommand{\rlExtDegFourFourRecovery}{95.2}
\newcommand{\rlExtDegFourFourRecoveryCILo}{94.1}
\newcommand{\rlExtDegFourFourRecoveryCIHi}{96.3}
\newcommand{\rlExtFRFiveZeroRecovery}{95.2}
\newcommand{\rlExtFRFiveZeroRecoveryCILo}{94.1}
\newcommand{\rlExtFRFiveZeroRecoveryCIHi}{96.3}
\newcommand{\rlExtDegFiveFiveRecovery}{94.5}
\newcommand{\rlExtDegFiveFiveRecoveryCILo}{93.2}
\newcommand{\rlExtDegFiveFiveRecoveryCIHi}{95.8}
\newcommand{\rlExtFRFourZeroRecovery}{94.5}
\newcommand{\rlExtFRFourZeroRecoveryCILo}{93.2}
\newcommand{\rlExtFRFourZeroRecoveryCIHi}{95.8}
\newcommand{\rlExtDegSixSixRecovery}{93.4}
\newcommand{\rlExtDegSixSixRecoveryCILo}{92.5}
\newcommand{\rlExtDegSixSixRecoveryCIHi}{94.3}
\newcommand{\rlExtFRThreeZeroRecovery}{93.4}
\newcommand{\rlExtFRThreeZeroRecoveryCILo}{92.5}
\newcommand{\rlExtFRThreeZeroRecoveryCIHi}{94.3}
\newcommand{\rlExtDegSevenSevenRecovery}{93.9}
\newcommand{\rlExtDegSevenSevenRecoveryCILo}{93.0}
\newcommand{\rlExtDegSevenSevenRecoveryCIHi}{94.8}
\newcommand{\rlExtFRTwoZeroRecovery}{93.9}
\newcommand{\rlExtFRTwoZeroRecoveryCILo}{93.0}
\newcommand{\rlExtFRTwoZeroRecoveryCIHi}{94.8}
\newcommand{\rlExtDegEightNineRecovery}{93.6}
\newcommand{\rlExtDegEightNineRecoveryCILo}{92.5}
\newcommand{\rlExtDegEightNineRecoveryCIHi}{94.6}
\newcommand{\rlExtFROneZeroRecovery}{93.6}
\newcommand{\rlExtFROneZeroRecoveryCILo}{92.5}
\newcommand{\rlExtFROneZeroRecoveryCIHi}{94.6}
\newcommand{\rlExtDegNineFourRecovery}{92.7}
\newcommand{\rlExtDegNineFourRecoveryCILo}{91.8}
\newcommand{\rlExtDegNineFourRecoveryCIHi}{93.8}
\newcommand{\rlExtFRFiveRecovery}{92.7}
\newcommand{\rlExtFRFiveRecoveryCILo}{91.8}
\newcommand{\rlExtFRFiveRecoveryCIHi}{93.8}


\newcommand{\latRounds}{5000}
\newcommand{\latWarmup}{500}
\newcommand{\latPBdTwentySixRouteMedian}{22.5}
\newcommand{\latPBdTwentySixRoutePNineFive}{26.5}
\newcommand{\latPBdTwentySixRoutePNineNine}{39.2}
\newcommand{\latPBdTwentySixUpdateMedian}{20.4}
\newcommand{\latPBdTwentySixUpdatePNineFive}{23.0}
\newcommand{\latPBdTwentySixUpdatePNineNine}{46.0}
\newcommand{\latPBdTwentySixTotalMedian}{43.0}
\newcommand{\latPBdTwentySixTotalPNineFive}{48.8}
\newcommand{\latPBdTwentySixThroughput}{22199}
\newcommand{\latPBdFullRouteMedian}{163.0}
\newcommand{\latPBdFullRoutePNineFive}{234.5}
\newcommand{\latPBdFullRoutePNineNine}{271.9}
\newcommand{\latPBdFullUpdateMedian}{456.9}
\newcommand{\latPBdFullUpdatePNineFive}{539.0}
\newcommand{\latPBdFullUpdatePNineNine}{617.9}
\newcommand{\latPBdFullTotalMedian}{622.8}
\newcommand{\latPBdFullTotalPNineFive}{753.6}
\newcommand{\latPBdFullThroughput}{1582}
\newcommand{\latSMdTwentySixRouteMedian}{5.8}
\newcommand{\latSMdTwentySixRoutePNineFive}{6.2}
\newcommand{\latSMdTwentySixRoutePNineNine}{8.6}
\newcommand{\latSMdTwentySixUpdateMedian}{8.1}
\newcommand{\latSMdTwentySixUpdatePNineFive}{8.4}
\newcommand{\latSMdTwentySixUpdatePNineNine}{10.3}
\newcommand{\latSMdTwentySixTotalMedian}{13.9}
\newcommand{\latSMdTwentySixTotalPNineFive}{15.0}
\newcommand{\latSMdTwentySixThroughput}{70459}
\newcommand{\latSMdFullRouteMedian}{105.2}
\newcommand{\latSMdFullRoutePNineFive}{188.1}
\newcommand{\latSMdFullRoutePNineNine}{245.6}
\newcommand{\latSMdFullUpdateMedian}{320.0}
\newcommand{\latSMdFullUpdatePNineFive}{970.8}
\newcommand{\latSMdFullUpdatePNineNine}{1001.2}
\newcommand{\latSMdFullTotalMedian}{447.0}
\newcommand{\latSMdFullTotalPNineFive}{1106.2}
\newcommand{\latSMdFullThroughput}{1608}
\newcommand{\latCachedDTwentySixRouteMedian}{7.2}
\newcommand{\latCachedDTwentySixRoutePNineFive}{8.0}
\newcommand{\latCachedDTwentySixRoutePNineNine}{10.2}
\newcommand{\latCachedDTwentySixUpdateMedian}{18.8}
\newcommand{\latCachedDTwentySixUpdatePNineFive}{19.3}
\newcommand{\latCachedDTwentySixUpdatePNineNine}{26.7}
\newcommand{\latCachedDTwentySixTotalMedian}{26.2}
\newcommand{\latCachedDTwentySixTotalPNineFive}{27.4}
\newcommand{\latCachedDTwentySixThroughput}{39687}
\newcommand{\latCachedDFullRouteMedian}{138.1}
\newcommand{\latCachedDFullRoutePNineFive}{230.0}
\newcommand{\latCachedDFullRoutePNineNine}{288.8}
\newcommand{\latCachedDFullUpdateMedian}{1679.3}
\newcommand{\latCachedDFullUpdatePNineFive}{2255.2}
\newcommand{\latCachedDFullUpdatePNineNine}{2417.6}
\newcommand{\latCachedDFullTotalMedian}{1859.7}
\newcommand{\latCachedDFullTotalPNineFive}{2408.0}
\newcommand{\latCachedDFullThroughput}{507}
\newcommand{\latPerRouteDTwentySixRouteMedian}{50.8}
\newcommand{\latPerRouteDTwentySixRoutePNineFive}{57.3}
\newcommand{\latPerRouteDTwentySixRoutePNineNine}{128.5}
\newcommand{\latPerRouteDTwentySixUpdateMedian}{4.0}
\newcommand{\latPerRouteDTwentySixUpdatePNineFive}{5.0}
\newcommand{\latPerRouteDTwentySixUpdatePNineNine}{12.3}
\newcommand{\latPerRouteDTwentySixTotalMedian}{55.3}
\newcommand{\latPerRouteDTwentySixTotalPNineFive}{62.0}
\newcommand{\latPerRouteDTwentySixThroughput}{17203}
\newcommand{\latPerRouteDFullRouteMedian}{4569.0}
\newcommand{\latPerRouteDFullRoutePNineFive}{4959.3}
\newcommand{\latPerRouteDFullRoutePNineNine}{5637.8}
\newcommand{\latPerRouteDFullUpdateMedian}{132.1}
\newcommand{\latPerRouteDFullUpdatePNineFive}{785.8}
\newcommand{\latPerRouteDFullUpdatePNineNine}{802.5}
\newcommand{\latPerRouteDFullTotalMedian}{4808.0}
\newcommand{\latPerRouteDFullTotalPNineFive}{5381.5}
\newcommand{\latPerRouteDFullThroughput}{203}
\newcommand{\latSpeedupAlgoSmVsCachedDTwoSixRoute}{1.2}
\newcommand{\latSpeedupAlgoSmVsCachedDTwoSixUpdate}{2.3}
\newcommand{\latSpeedupAlgoSmVsCachedDTwoSixThroughput}{1.8}
\newcommand{\latSpeedupAlgoSmVsCachedDThreeEightFiveRoute}{1.3}
\newcommand{\latSpeedupAlgoSmVsCachedDThreeEightFiveUpdate}{5.2}
\newcommand{\latSpeedupAlgoSmVsCachedDThreeEightFiveThroughput}{3.2}
\newcommand{\latSpeedupProdOverheadDTwoSixRoute}{3.9}
\newcommand{\latSpeedupProdOverheadDTwoSixUpdate}{2.5}
\newcommand{\latSpeedupProdOverheadDThreeEightFiveRoute}{1.5}
\newcommand{\latSpeedupProdOverheadDThreeEightFiveUpdate}{1.4}
\newcommand{\latSpeedupPcaVsRawProdThroughput}{14.0}
\newcommand{\latEteEmbedMedianMs}{8.8}
\newcommand{\latEteEmbedPNineFiveMs}{9.3}
\newcommand{\latEteEmbedPNineNineMs}{9.4}
\newcommand{\latEtePcaMedianMs}{0.11}
\newcommand{\latEtePcaPNineFiveMs}{0.15}
\newcommand{\latEteRouteMedianMs}{0.056}
\newcommand{\latEteRoutePNineFiveMs}{0.072}
\newcommand{\latEteRoutePNineNineMs}{0.078}
\newcommand{\latEteTotalMedianMs}{9.0}
\newcommand{\latEteTotalPNineFiveMs}{9.4}
\newcommand{\latEteTotalPNineNineMs}{9.7}
\newcommand{\latEteEmbedPctOfTotal}{98.0}
\newcommand{\latEteRoutePctOfTotal}{0.6}
\newcommand{\latEteRounds}{200}
\newcommand{\latEtePromptPool}{20}
\newcommand{\latInfTrials}{20}
\newcommand{\latInfLlamaShortTtft}{820}
\newcommand{\latInfLlamaShortTtftCILo}{453}
\newcommand{\latInfLlamaShortTtftCIHi}{1425}
\newcommand{\latInfLlamaShortTotal}{7,001}
\newcommand{\latInfLlamaShortTotalCILo}{4526}
\newcommand{\latInfLlamaShortTotalCIHi}{10032}
\newcommand{\latInfLlamaShortRoutingPct}{0.13}
\newcommand{\latInfLlamaMedTtft}{607}
\newcommand{\latInfLlamaMedTtftCILo}{441}
\newcommand{\latInfLlamaMedTtftCIHi}{826}
\newcommand{\latInfLlamaMedTotal}{9,958}
\newcommand{\latInfLlamaMedTotalCILo}{6599}
\newcommand{\latInfLlamaMedTotalCIHi}{13558}
\newcommand{\latInfLlamaMedRoutingPct}{0.09}
\newcommand{\latInfLlamaLongTtft}{684}
\newcommand{\latInfLlamaLongTtftCILo}{469}
\newcommand{\latInfLlamaLongTtftCIHi}{962}
\newcommand{\latInfLlamaLongTotal}{7,235}
\newcommand{\latInfLlamaLongTotalCILo}{5477}
\newcommand{\latInfLlamaLongTotalCIHi}{9035}
\newcommand{\latInfLlamaLongRoutingPct}{0.12}
\newcommand{\latInfMistralShortTtft}{1044}
\newcommand{\latInfMistralShortTtftCILo}{611}
\newcommand{\latInfMistralShortTtftCIHi}{1858}
\newcommand{\latInfMistralShortTotal}{5,811}
\newcommand{\latInfMistralShortTotalCILo}{4428}
\newcommand{\latInfMistralShortTotalCIHi}{7361}
\newcommand{\latInfMistralShortRoutingPct}{0.16}
\newcommand{\latInfMistralMedTtft}{665}
\newcommand{\latInfMistralMedTtftCILo}{592}
\newcommand{\latInfMistralMedTtftCIHi}{766}
\newcommand{\latInfMistralMedTotal}{6,442}
\newcommand{\latInfMistralMedTotalCILo}{5357}
\newcommand{\latInfMistralMedTotalCIHi}{7414}
\newcommand{\latInfMistralMedRoutingPct}{0.14}
\newcommand{\latInfMistralLongTtft}{636}
\newcommand{\latInfMistralLongTtftCILo}{593}
\newcommand{\latInfMistralLongTtftCIHi}{682}
\newcommand{\latInfMistralLongTotal}{8,445}
\newcommand{\latInfMistralLongTotalCILo}{8233}
\newcommand{\latInfMistralLongTotalCIHi}{8670}
\newcommand{\latInfMistralLongRoutingPct}{0.11}
\newcommand{\latInfGemFlashShortTtft}{758}
\newcommand{\latInfGemFlashShortTtftCILo}{567}
\newcommand{\latInfGemFlashShortTtftCIHi}{1012}
\newcommand{\latInfGemFlashShortTotal}{2,574}
\newcommand{\latInfGemFlashShortTotalCILo}{2014}
\newcommand{\latInfGemFlashShortTotalCIHi}{3154}
\newcommand{\latInfGemFlashShortRoutingPct}{0.35}
\newcommand{\latInfGemFlashMedTtft}{858}
\newcommand{\latInfGemFlashMedTtftCILo}{584}
\newcommand{\latInfGemFlashMedTtftCIHi}{1200}
\newcommand{\latInfGemFlashMedTotal}{3,256}
\newcommand{\latInfGemFlashMedTotalCILo}{2701}
\newcommand{\latInfGemFlashMedTotalCIHi}{3861}
\newcommand{\latInfGemFlashMedRoutingPct}{0.28}
\newcommand{\latInfGemFlashLongTtft}{890}
\newcommand{\latInfGemFlashLongTtftCILo}{566}
\newcommand{\latInfGemFlashLongTtftCIHi}{1275}
\newcommand{\latInfGemFlashLongTotal}{3,756}
\newcommand{\latInfGemFlashLongTotalCILo}{3298}
\newcommand{\latInfGemFlashLongTotalCIHi}{4303}
\newcommand{\latInfGemFlashLongRoutingPct}{0.24}
\newcommand{\latInfGemProShortTtft}{6508}
\newcommand{\latInfGemProShortTtftCILo}{6167}
\newcommand{\latInfGemProShortTtftCIHi}{6903}
\newcommand{\latInfGemProShortTotal}{6,617}
\newcommand{\latInfGemProShortTotalCILo}{6279}
\newcommand{\latInfGemProShortTotalCIHi}{7005}
\newcommand{\latInfGemProShortRoutingPct}{0.14}
\newcommand{\latInfGemProMedTtft}{7901}
\newcommand{\latInfGemProMedTtftCILo}{6686}
\newcommand{\latInfGemProMedTtftCIHi}{10004}
\newcommand{\latInfGemProMedTotal}{8,068}
\newcommand{\latInfGemProMedTotalCILo}{6830}
\newcommand{\latInfGemProMedTotalCIHi}{10194}
\newcommand{\latInfGemProMedRoutingPct}{0.11}
\newcommand{\latInfGemProLongTtft}{8188}
\newcommand{\latInfGemProLongTtftCILo}{7158}
\newcommand{\latInfGemProLongTtftCIHi}{9429}
\newcommand{\latInfGemProLongTotal}{8,638}
\newcommand{\latInfGemProLongTotalCILo}{7294}
\newcommand{\latInfGemProLongTotalCIHi}{10533}
\newcommand{\latInfGemProLongRoutingPct}{0.10}


\newcommand{\jrNseeds}{20}
\newcommand{\jrNTest}{1328}
\newcommand{\jrTRAlpha}{0.05}
\newcommand{\jrTRGamma}{0.997}
\newcommand{\jrROneUncTRRegret}{51.0}
\newcommand{\jrROneUncTRRegretSE}{1.8}
\newcommand{\jrROneUncTRRegretCILo}{47.7}
\newcommand{\jrROneUncTRRegretCIHi}{54.4}
\newcommand{\jrROneUncRandRegret}{106.2}
\newcommand{\jrROneUncRandRegretSE}{0.8}
\newcommand{\jrROneUncRandRegretCILo}{104.6}
\newcommand{\jrROneUncRandRegretCIHi}{107.8}
\newcommand{\jrROneTightTRRegret}{127.7}
\newcommand{\jrROneTightTRRegretSE}{5.0}
\newcommand{\jrROneTightTRRegretCILo}{118.3}
\newcommand{\jrROneTightTRRegretCIHi}{137.4}
\newcommand{\jrROneTightRandRegret}{106.2}
\newcommand{\jrROneTightRandRegretSE}{0.8}
\newcommand{\jrROneTightRandRegretCILo}{104.6}
\newcommand{\jrROneTightRandRegretCIHi}{107.8}
\newcommand{\jrROneModTRRegret}{90.6}
\newcommand{\jrROneModTRRegretSE}{2.9}
\newcommand{\jrROneModTRRegretCILo}{85.0}
\newcommand{\jrROneModTRRegretCIHi}{96.2}
\newcommand{\jrROneModRandRegret}{106.2}
\newcommand{\jrROneModRandRegretSE}{0.8}
\newcommand{\jrROneModRandRegretCILo}{104.6}
\newcommand{\jrROneModRandRegretCIHi}{107.8}
\newcommand{\jrROneLooseTRRegret}{65.5}
\newcommand{\jrROneLooseTRRegretSE}{1.1}
\newcommand{\jrROneLooseTRRegretCILo}{63.6}
\newcommand{\jrROneLooseTRRegretCIHi}{67.7}
\newcommand{\jrROneLooseRandRegret}{106.2}
\newcommand{\jrROneLooseRandRegretSE}{0.8}
\newcommand{\jrROneLooseRandRegretCILo}{104.6}
\newcommand{\jrROneLooseRandRegretCIHi}{107.8}
\newcommand{\jrGPTUncTRRegret}{32.8}
\newcommand{\jrGPTUncTRRegretSE}{1.7}
\newcommand{\jrGPTUncTRRegretCILo}{29.7}
\newcommand{\jrGPTUncTRRegretCIHi}{36.1}
\newcommand{\jrGPTUncRandRegret}{71.9}
\newcommand{\jrGPTUncRandRegretSE}{0.5}
\newcommand{\jrGPTUncRandRegretCILo}{70.9}
\newcommand{\jrGPTUncRandRegretCIHi}{72.9}
\newcommand{\jrGPTTightTRRegret}{113.1}
\newcommand{\jrGPTTightTRRegretSE}{5.4}
\newcommand{\jrGPTTightTRRegretCILo}{102.9}
\newcommand{\jrGPTTightTRRegretCIHi}{123.3}
\newcommand{\jrGPTTightRandRegret}{71.9}
\newcommand{\jrGPTTightRandRegretSE}{0.5}
\newcommand{\jrGPTTightRandRegretCILo}{70.9}
\newcommand{\jrGPTTightRandRegretCIHi}{72.9}
\newcommand{\jrGPTModTRRegret}{76.7}
\newcommand{\jrGPTModTRRegretSE}{4.8}
\newcommand{\jrGPTModTRRegretCILo}{68.0}
\newcommand{\jrGPTModTRRegretCIHi}{86.3}
\newcommand{\jrGPTModRandRegret}{71.9}
\newcommand{\jrGPTModRandRegretSE}{0.5}
\newcommand{\jrGPTModRandRegretCILo}{70.9}
\newcommand{\jrGPTModRandRegretCIHi}{72.9}
\newcommand{\jrGPTLooseTRRegret}{45.1}
\newcommand{\jrGPTLooseTRRegretSE}{1.5}
\newcommand{\jrGPTLooseTRRegretCILo}{42.5}
\newcommand{\jrGPTLooseTRRegretCIHi}{48.1}
\newcommand{\jrGPTLooseRandRegret}{71.9}
\newcommand{\jrGPTLooseRandRegretSE}{0.5}
\newcommand{\jrGPTLooseRandRegretCILo}{70.9}
\newcommand{\jrGPTLooseRandRegretCIHi}{72.9}
\newcommand{\jrClaudeUncTRRegret}{37.6}
\newcommand{\jrClaudeUncTRRegretSE}{1.8}
\newcommand{\jrClaudeUncTRRegretCILo}{34.4}
\newcommand{\jrClaudeUncTRRegretCIHi}{41.1}
\newcommand{\jrClaudeUncRandRegret}{97.6}
\newcommand{\jrClaudeUncRandRegretSE}{0.7}
\newcommand{\jrClaudeUncRandRegretCILo}{96.3}
\newcommand{\jrClaudeUncRandRegretCIHi}{98.9}
\newcommand{\jrClaudeTightTRRegret}{117.0}
\newcommand{\jrClaudeTightTRRegretSE}{4.0}
\newcommand{\jrClaudeTightTRRegretCILo}{109.8}
\newcommand{\jrClaudeTightTRRegretCIHi}{125.0}
\newcommand{\jrClaudeTightRandRegret}{97.6}
\newcommand{\jrClaudeTightRandRegretSE}{0.7}
\newcommand{\jrClaudeTightRandRegretCILo}{96.3}
\newcommand{\jrClaudeTightRandRegretCIHi}{98.9}
\newcommand{\jrClaudeModTRRegret}{77.0}
\newcommand{\jrClaudeModTRRegretSE}{2.7}
\newcommand{\jrClaudeModTRRegretCILo}{71.9}
\newcommand{\jrClaudeModTRRegretCIHi}{82.0}
\newcommand{\jrClaudeModRandRegret}{97.6}
\newcommand{\jrClaudeModRandRegretSE}{0.7}
\newcommand{\jrClaudeModRandRegretCILo}{96.3}
\newcommand{\jrClaudeModRandRegretCIHi}{98.9}
\newcommand{\jrClaudeLooseTRRegret}{52.6}
\newcommand{\jrClaudeLooseTRRegretSE}{0.6}
\newcommand{\jrClaudeLooseTRRegretCILo}{51.5}
\newcommand{\jrClaudeLooseTRRegretCIHi}{53.6}
\newcommand{\jrClaudeLooseRandRegret}{97.6}
\newcommand{\jrClaudeLooseRandRegretSE}{0.7}
\newcommand{\jrClaudeLooseRandRegretCILo}{96.3}
\newcommand{\jrClaudeLooseRandRegretCIHi}{98.9}
\newcommand{\jrROneUncReductionPct}{52}
\newcommand{\jrROneUncReductionPctCILo}{49}
\newcommand{\jrROneUncReductionPctCIHi}{55}
\newcommand{\jrROneTightReductionPct}{-20}
\newcommand{\jrROneTightReductionPctCILo}{-30}
\newcommand{\jrROneTightReductionPctCIHi}{-11}
\newcommand{\jrROneModReductionPct}{15}
\newcommand{\jrROneModReductionPctCILo}{9}
\newcommand{\jrROneModReductionPctCIHi}{20}
\newcommand{\jrROneLooseReductionPct}{38}
\newcommand{\jrROneLooseReductionPctCILo}{36}
\newcommand{\jrROneLooseReductionPctCIHi}{41}
\newcommand{\jrGPTUncReductionPct}{54}
\newcommand{\jrGPTUncReductionPctCILo}{49}
\newcommand{\jrGPTUncReductionPctCIHi}{59}
\newcommand{\jrGPTTightReductionPct}{-57}
\newcommand{\jrGPTTightReductionPctCILo}{-71}
\newcommand{\jrGPTTightReductionPctCIHi}{-43}
\newcommand{\jrGPTModReductionPct}{-7}
\newcommand{\jrGPTModReductionPctCILo}{-20}
\newcommand{\jrGPTModReductionPctCIHi}{5}
\newcommand{\jrGPTLooseReductionPct}{37}
\newcommand{\jrGPTLooseReductionPctCILo}{33}
\newcommand{\jrGPTLooseReductionPctCIHi}{41}
\newcommand{\jrClaudeUncReductionPct}{61}
\newcommand{\jrClaudeUncReductionPctCILo}{58}
\newcommand{\jrClaudeUncReductionPctCIHi}{65}
\newcommand{\jrClaudeTightReductionPct}{-20}
\newcommand{\jrClaudeTightReductionPctCILo}{-27}
\newcommand{\jrClaudeTightReductionPctCIHi}{-13}
\newcommand{\jrClaudeModReductionPct}{21}
\newcommand{\jrClaudeModReductionPctCILo}{16}
\newcommand{\jrClaudeModReductionPctCIHi}{27}
\newcommand{\jrClaudeLooseReductionPct}{46}
\newcommand{\jrClaudeLooseReductionPctCILo}{45}
\newcommand{\jrClaudeLooseReductionPctCIHi}{48}
\newcommand{\jrGPTRandRegretCompressionPct}{32}
\newcommand{\jrClaudeRandRegretCompressionPct}{8}
\newcommand{\jrUncGPTROneRatio}{0.64}
\newcommand{\jrUncGPTROneRatioCILo}{0.60}
\newcommand{\jrUncGPTROneRatioCIHi}{0.68}
\newcommand{\jrTightGPTROneRatio}{0.89}
\newcommand{\jrTightGPTROneRatioCILo}{0.82}
\newcommand{\jrTightGPTROneRatioCIHi}{0.97}
\newcommand{\jrModGPTROneRatio}{0.85}
\newcommand{\jrModGPTROneRatioCILo}{0.76}
\newcommand{\jrModGPTROneRatioCIHi}{0.94}
\newcommand{\jrLooseGPTROneRatio}{0.69}
\newcommand{\jrLooseGPTROneRatioCILo}{0.65}
\newcommand{\jrLooseGPTROneRatioCIHi}{0.74}
\newcommand{\jrUncClaudeROneRatio}{0.74}
\newcommand{\jrUncClaudeROneRatioCILo}{0.70}
\newcommand{\jrUncClaudeROneRatioCIHi}{0.79}
\newcommand{\jrTightClaudeROneRatio}{0.92}
\newcommand{\jrTightClaudeROneRatioCILo}{0.87}
\newcommand{\jrTightClaudeROneRatioCIHi}{1.00}
\newcommand{\jrModClaudeROneRatio}{0.85}
\newcommand{\jrModClaudeROneRatioCILo}{0.81}
\newcommand{\jrModClaudeROneRatioCIHi}{0.91}
\newcommand{\jrLooseClaudeROneRatio}{0.80}
\newcommand{\jrLooseClaudeROneRatioCILo}{0.78}
\newcommand{\jrLooseClaudeROneRatioCIHi}{0.83}
\newcommand{\jrRandRegretCompressionPct}{32}
\newcommand{\jrNSubset}{2000}
\newcommand{\jrGPTSpearman}{0.658}
\newcommand{\jrGPTKendall}{0.547}
\newcommand{\jrGPTMAD}{0.074}
\newcommand{\jrGPTBias}{0.039}
\newcommand{\jrGPTBALo}{-0.23}
\newcommand{\jrGPTBAHi}{0.37}
\newcommand{\jrClaudeSpearman}{0.633}
\newcommand{\jrClaudeKendall}{0.528}
\newcommand{\jrClaudeMAD}{0.075}
\newcommand{\jrClaudeBias}{-0.012}
\newcommand{\jrClaudeBALo}{-0.36}
\newcommand{\jrClaudeBAHi}{0.24}
\newcommand{\jrGPTLlamaMean}{0.857}
\newcommand{\jrGPTMistralMean}{0.952}
\newcommand{\jrGPTGeminiMean}{0.964}
\newcommand{\jrGPTOverallMean}{0.924}
\newcommand{\jrClaudeLlamaMean}{0.779}
\newcommand{\jrClaudeMistralMean}{0.915}
\newcommand{\jrClaudeGeminiMean}{0.928}
\newcommand{\jrClaudeOverallMean}{0.874}
\newcommand{\jrROneMean}{0.886}
\newcommand{\jrROneLlamaMean}{0.798}
\newcommand{\jrROneMistralMean}{0.923}
\newcommand{\jrROneGeminiMean}{0.935}
\newcommand{\jrROneMeanGap}{0.192}
\newcommand{\jrROneMedianGap}{0.110}
\newcommand{\jrGPTMeanGap}{0.140}
\newcommand{\jrGPTMedianGap}{0.055}
\newcommand{\jrClaudeMeanGap}{0.186}
\newcommand{\jrClaudeMedianGap}{0.035}
\newcommand{\jrOrderingAllSig}{true}
\newcommand{\jrOrderingWorstCILo}{0.0046}
\newcommand{\jrROneGeminiMistralDiff}{0.012}
\newcommand{\jrROneGeminiMistralCILo}{0.005}
\newcommand{\jrROneGeminiMistralCIHi}{0.019}
\newcommand{\jrROneGeminiLlamaDiff}{0.137}
\newcommand{\jrROneGeminiLlamaCILo}{0.124}
\newcommand{\jrROneGeminiLlamaCIHi}{0.150}
\newcommand{\jrROneMistralLlamaDiff}{0.125}
\newcommand{\jrROneMistralLlamaCILo}{0.112}
\newcommand{\jrROneMistralLlamaCIHi}{0.138}
\newcommand{\jrGPTGeminiMistralDiff}{0.011}
\newcommand{\jrGPTGeminiMistralCILo}{0.006}
\newcommand{\jrGPTGeminiMistralCIHi}{0.017}
\newcommand{\jrGPTGeminiLlamaDiff}{0.106}
\newcommand{\jrGPTGeminiLlamaCILo}{0.096}
\newcommand{\jrGPTGeminiLlamaCIHi}{0.117}
\newcommand{\jrGPTMistralLlamaDiff}{0.095}
\newcommand{\jrGPTMistralLlamaCILo}{0.084}
\newcommand{\jrGPTMistralLlamaCIHi}{0.106}
\newcommand{\jrClaudeGeminiMistralDiff}{0.013}
\newcommand{\jrClaudeGeminiMistralCILo}{0.005}
\newcommand{\jrClaudeGeminiMistralCIHi}{0.021}
\newcommand{\jrClaudeGeminiLlamaDiff}{0.149}
\newcommand{\jrClaudeGeminiLlamaCILo}{0.135}
\newcommand{\jrClaudeGeminiLlamaCIHi}{0.163}
\newcommand{\jrClaudeMistralLlamaDiff}{0.136}
\newcommand{\jrClaudeMistralLlamaCILo}{0.122}
\newcommand{\jrClaudeMistralLlamaCIHi}{0.150}
\newcommand{\jrCrossROneROneMean}{0.966}
\newcommand{\jrCrossROneGPTMean}{0.955}
\newcommand{\jrCrossROneGPTCapture}{97.5}
\newcommand{\jrCrossROneClaudeMean}{0.923}
\newcommand{\jrCrossROneClaudeCapture}{97.4}
\newcommand{\jrCrossGPTROneMean}{0.926}
\newcommand{\jrCrossGPTROneCapture}{95.8}
\newcommand{\jrCrossGPTGPTMean}{0.979}
\newcommand{\jrCrossGPTClaudeMean}{0.915}
\newcommand{\jrCrossGPTClaudeCapture}{96.6}
\newcommand{\jrCrossClaudeROneMean}{0.926}
\newcommand{\jrCrossClaudeROneCapture}{95.8}
\newcommand{\jrCrossClaudeGPTMean}{0.953}
\newcommand{\jrCrossClaudeGPTCapture}{97.3}
\newcommand{\jrCrossClaudeClaudeMean}{0.948}
\newcommand{\jrROneOracleLift}{0.031}
\newcommand{\jrGPTOracleLift}{0.016}
\newcommand{\jrClaudeOracleLift}{0.020}
\newcommand{\jrROneMinCapture}{97.4}
\newcommand{\jrROneMaxCapture}{97.5}
\newcommand{\jrOtherOnROneMinCapture}{95.8}
\newcommand{\jrOtherOnROneMaxCapture}{95.8}
\newcommand{\jrMinOffDiagCapture}{96}
\newcommand{\jrCaptureMarginMinPP}{0.2}
\newcommand{\jrCaptureMarginMaxPP}{0.8}
\newcommand{\jrCellCIHalfWidthPP}{0.5}
\newcommand{\jrOracleLiftMaxCompressionPct}{49}
\newcommand{\jrROneMargin}{0.0485}
\newcommand{\jrPanelMargin}{0.0338}
\newcommand{\jrMarginCompressionPct}{30}
\newcommand{\jrFracCompressed}{63}
\newcommand{\jrROneSNR}{0.617}
\newcommand{\jrPanelSNR}{0.586}


\newcommand{\ppPriceRangeX}{530}
\newcommand{\ppGamma}{0.997}
\newcommand{\ppAlpha}{0.01}
\newcommand{\ppNeff}{1164}
\newcommand{\ppEffMemSteps}{333}
\newcommand{\ppHalfLife}{231}
\newcommand{\ppGammaDecayOneK}{0.05}
\newcommand{\ppMaxBudgetOvershootPct}{0.4}
\newcommand{\ppMaxQualityLift}{0.071}
\newcommand{\ppMaxOvershootX}{6.9}
\newcommand{\ppAdoptionSteps}{142}
\newcommand{\ppEteTotalMedianMs}{9.8}
\newcommand{\ppEteRoutePctOfTotal}{1}
\newcommand{\ppEteMaxPctOfInference}{0.4}
\newcommand{\ppPBRouteMedianUs}{22.5}
\newcommand{\ppPBTotalMedianUs}{43}
\newcommand{\ppPBThroughput}{22{,}000}
\newcommand{\ppPcaThroughputGainX}{44}
\newcommand{\ppCfDegPct}{18}

\twocolumn[
\mlsystitle{ParetoBandit: Budget-Paced Adaptive Routing\\for Non-Stationary LLM Serving}

\mlsyssetsymbol{equal}{*}

\begin{mlsysauthorlist}
\mlsysauthor{Annette Taberner-Miller}{atm}
\end{mlsysauthorlist}

\mlsysaffiliation{atm}{Independent Researcher}

\mlsyscorrespondingauthor{Annette Taberner-Miller}{taberner@alum.mit.edu}

\mlsyskeywords{LLM routing, contextual bandits, budget pacing, non-stationary, model serving}

\vskip 0.3in

\begin{abstract}
Multi-model LLM serving operates in a non-stationary, noisy
environment: providers revise pricing, model quality can shift or
regress without notice, and new models arrive regularly. More than a
dozen recent methods have proposed learned routers to navigate the
resulting quality--cost tradeoff across portfolios spanning a
${\sim}\ppPriceRangeX\times$ cost range. Despite this activity, two
gaps in the current solution space limit routing effectiveness under
these conditions: no existing router enforces a dollar-denominated cost
ceiling in closed loop over an open-ended request stream, and none
provides principled online adaptation to post-deployment shifts in
pricing or model quality.
We present \textbf{ParetoBandit}, an open-source adaptive router built
on cost-aware contextual bandits that addresses both gaps. Its core
contributions are: (1)~an online primal--dual budget pacer that
enforces a per-request cost ceiling without a known horizon, and
(2)~geometric forgetting on sufficient statistics that gives the bandit
bounded memory for tracking quality and cost shifts. A hot-swap model
registry further supports runtime model changes with
budget-controlled exploration.
On 1,824 benchmark prompts with a three-model portfolio, the router
maintains budget compliance within $\ppMaxBudgetOvershootPct\%$, adapts
to price and quality shifts with up to $+\ppMaxQualityLift$ quality
lift, and integrates a cold-started model within
${\sim}\ppAdoptionSteps$ steps.

\vskip 6pt
\noindent\small Code: \url{https://github.com/ParetoBandit/ParetoBandit} \quad\textbar\quad \texttt{pip install paretobandit}
\end{abstract}

]

\raggedbottom

\printAffiliationsAndNotice{}

\section{Introduction}
\label{sec:intro}

Production LLM serving relies on multi-model portfolios spanning a ${\sim}\ppPriceRangeX\times$ cost range, yet no single model dominates on every input. This complementarity has motivated a growing line of work on learned routers that dispatch each request to the most suitable model~\cite{chen2023frugalgpt, ong2024routellm, chen2024routerdc, feng2025graphrouter, mei2025omnirouter, panda2025pilot, bhatti2026proteus}. The open question is no longer \emph{whether} to route, but how to do so under the operational realities of deployment: dollar budgets, non-stationary pricing, and evolving model portfolios.

Many existing routers learn a fixed policy offline and freeze it at serving time, whether through cascading~\cite{chen2023frugalgpt}, supervised classification on preference or embedding data~\cite{ong2024routellm, chen2024routerdc, feng2025graphrouter}, or constrained optimization with budget targets~\cite{mei2025omnirouter, bhatti2026proteus}. All navigate the quality--cost tradeoff, but none demonstrate adaptation to post-deployment shifts---despite the environment being non-stationary: providers silently update APIs in ways that shift quality~\cite{chen2023chatgptdrift, ma2024prompt}, pricing can shift substantially (for instance, in 2024 OpenAI cut GPT-4o input prices by roughly 50\%~\cite{openai2024gpt4opricing}), and new models launch regularly. PROTEUS~\cite{bhatti2026proteus} explicitly highlights cost-adaptive routing as an open problem.\looseness=-1

Online learning offers a path forward. Bandit-based routers update their policy from the outcome of each dispatched request~\cite{panda2025pilot, wang2025barp, li2025llmbandit}, and continual-learning frameworks further accommodate evolving model pools~\cite{wang2025mixllm}. Yet even among these adaptive methods, two deployment gaps remain, which we detail below.

\noindent\textbf{Challenge 1: Closed-loop enforcement of cost ceilings.}
Several routers incorporate budget or SLA targets~\cite{panda2025pilot, mei2025omnirouter, bhatti2026proteus}, each tailored to a specific deployment model: PILOT~\cite{panda2025pilot} operates over a known finite horizon, OmniRouter~\cite{mei2025omnirouter} optimizes over a fixed prompt set, and PROTEUS~\cite{bhatti2026proteus} fixes its dual variables at deployment time. Closed-loop enforcement of a per-request cost ceiling over an open-ended request stream---where the horizon is unknown and conditions may shift---remains an open problem.

\noindent\textbf{Challenge 2: Principled handling of non-stationarity.}
Non-stationary bandit algorithms are well studied~\cite{garivier2011switching, russac2019weighted}. However, we are not aware of any published LLM router that (i)~explicitly employs geometric discounting or sliding-window mechanisms to handle post-deployment shifts, or (ii)~systematically evaluates routing behaviour under model quality degradation or changes in model pricing during deployment. In our quality-degradation experiment (Section~\ref{sec:eval_catastrophic}), a bandit baseline without closed-loop budget control detects the degraded model but overshoots the cost target by up to $\ppMaxOvershootX\times$.

We address these challenges with \textbf{ParetoBandit}, an open-source adaptive routing system for budget-constrained, non-stationary LLM serving. ParetoBandit builds on cost-aware contextual bandits and introduces three system-level contributions:

\begin{enumerate}[nosep,leftmargin=*]
  \item \textbf{Autonomous budget control} (Section~\ref{sec:budget_pacer}). An online primal--dual budget pacer accepts an operator-specified per-request cost ceiling and enforces it in closed loop over an open-ended request stream with no known horizon, automatically reallocating freed budget when prices drop.

  \item \textbf{Non-stationarity resilience} (Section~\ref{sec:adaptive_learning}). Geometric forgetting gives the bandit a bounded effective memory that tracks shifts in both quality and cost. Exploration and forgetting rates are chosen via a Pareto knee-point procedure (Appendix~\ref{sec:hparam_sweep}) anchored by a single practitioner-specified adaptation horizon.

  \item \textbf{Runtime portfolio management} (Section~\ref{sec:production_infra}). A hot-swap model registry supports adding and removing models at runtime. A brief forced-exploration phase bootstraps each newcomer's posterior, after which UCB selection discovers its quality--cost niche while the budget pacer maintains compliance. The router discriminates rather than blindly adopting (Section~\ref{sec:eval_onboarding}).
\end{enumerate}

We evaluate ParetoBandit on a 1,824-prompt holdout set drawn from nine public benchmarks, using a three-tier model portfolio ($K{=}3$, with a $K{=}4$ onboarding extension) and an LLM judge with a continuous rubric (Section~\ref{sec:eval_setup}). Section~\ref{sec:system_design} details the system design, and Section~\ref{sec:evaluation} evaluates it with respect to both deployment challenges.

\section{Background and Problem}
\label{sec:background}

We formalize LLM routing as an online decision-making problem in which the router learns which model to dispatch for each prompt, respects an operator-specified cost ceiling, and adapts when model quality or pricing shifts after deployment. This section defines the three components of the problem; Section~\ref{sec:system_design} presents our solution.

\subsection{Contextual Bandit Routing}
\label{sec:bandit_routing}

At each discrete step $t = 1, 2, \ldots$, a user prompt arrives and the router observes a context vector $x_t \in \mathbb{R}^d$ derived from the prompt (Section~\ref{sec:context_repr}). The router selects a single model $a_t$ from a portfolio $\mathcal{A} = \{1, \ldots, K\}$ and dispatches the prompt. After inference, the system observes a scalar reward $r_t$ reflecting response quality and a realized cost $c_t$ in dollars. The reward may be derived from an automated LLM judge, explicit user feedback, or a task-completion metric; in our setup it is a continuous score normalized to $[0, 1]$ by the judge rubric (Section~\ref{sec:eval_setup}). The objective is to learn a policy $\pi$ that maximizes cumulative quality subject to a ceiling on average cost per request (detailed in Section~\ref{sec:budget_constraint}):
\begin{align}
    &\max_{\pi}\; \mathbb{E}\!\left[\sum_{t=1}^{T} r_{t,\,\pi(x_t)}\right]
    \label{eq:objective} \\
    &\text{s.t.}\quad
    \limsup_{T \to \infty}\;\tfrac{1}{T}\textstyle\sum_{t=1}^{T} c_{\pi(x_t)} \leq B
    \notag
\end{align}
where $B$ is the operator's maximum acceptable average cost per request in dollars (e.g., \$0.005/req). The router may spend well below $B$ when cheaper models deliver comparable quality; it approaches the ceiling only when doing so yields a meaningful quality gain. We relax this via a per-request Lagrange multiplier $\lambda_t$ whose scoring function and dynamics are detailed in Section~\ref{sec:budget_pacer}.

The router operates under \emph{bandit feedback}~\cite{lattimore2020bandit}: it observes the reward and cost only for the model to which a query was routed, never the counterfactual outcomes from the $K{-}1$ alternatives. This partial observability is inherent to LLM serving, where evaluating every model on every prompt would multiply cost by $K$.

\subsection{Context Representation}
\label{sec:context_repr}

The context vector $x_t$ is constructed by encoding the prompt with all-MiniLM-L6-v2, a lightweight model from the Sentence Transformers framework~\cite{reimers2019sentence}, projecting to 25 PCA components whitened to unit variance, and appending a bias term, yielding $d = 26$. The PCA projection is fitted on ${\sim}46{,}000$ disjoint LMSYS Arena prompts~\cite{zheng2023judging}. Section~\ref{sec:efficiency} details the latency impact of this dimensionality choice.

\subsection{Budget Constraint}
\label{sec:budget_constraint}

Each model $a \in \mathcal{A}$ incurs a deterministic cost $c_a(x_t)$ per request, typically computed from per-token API pricing. In a realistic portfolio, costs span orders of magnitude: the three-tier pool used in our experiments ranges from ${\sim}$\$0.00003/req (Llama-3.1-8B) to ${\sim}$\$0.015/req (Gemini-2.5-Pro), a $530{\times}$ spread.

The rate constraint in Eq.~\ref{eq:objective} departs from the classical Bandits with Knapsacks (BwK) framework~\cite{badanidiyuru2013bwk}, which imposes a cumulative budget over a known horizon $T$. Production LLM serving is open-ended: request volume is unknown, and the natural specification is a cost \emph{rate} (dollars per request), not a finite pool. Our formulation is horizon-free and self-regulates at any traffic volume.

We enforce the rate constraint through a dual variable $\lambda_t$ that rises when recent spending exceeds $B$ and falls when it is below. Section~\ref{sec:budget_pacer} defines the smoothed dual-ascent update and the two-layer enforcement mechanism (soft penalty plus hard cost ceiling) that operationalize $\lambda_t$.

\subsection{Non-Stationarity}
\label{sec:nonstationarity}

We address two forms of environment drift that affect production LLM deployments. \emph{Reward drift} occurs when providers silently update models behind their APIs~\cite{chen2023chatgptdrift, ma2024prompt}, changing the conditional expected reward. \emph{Cost drift} occurs when API pricing changes significantly~\cite{openai2024gpt4opricing}. The two interact: a price drop changes which models are cost-effective, requiring the router to jointly re-learn reward estimates and re-optimize the quality--cost frontier. Section~\ref{sec:adaptive_learning} presents our approach using geometric discounting~\cite{garivier2011switching, russac2019weighted}.

\section{System Design}
\label{sec:system_design}

ParetoBandit jointly learns reward surfaces from bandit feedback, enforces a long-run budget constraint ceiling, and adapts to non-stationarity, a combination that, to our knowledge, ParetoBandit is the first LLM routing system to integrate. It composes four foundational ingredients: LinUCB~\cite{li2010contextual} for contextual reward learning, BwK-inspired Lagrangian relaxation~\cite{badanidiyuru2013bwk, agrawal2016linear} for budget enforcement, geometric discounting~\cite{garivier2011switching, russac2019weighted} for non-stationarity, and offline-to-online warm-start priors~\cite{oetomo2023warmstart} for cold-start elimination. Their composition rests on LinUCB's sufficient-statistic representation ($A_a$, $b_a$ per arm): forgetting reduces to a scalar multiply, warmup to a matrix addition, and updates to $O(d^2)$ Sherman--Morrison operations, while the confidence bonus $x_t^\top A_a^{-1} x_t$, the exact posterior variance under a Gaussian linear model, provides context-dependent calibrated exploration from the first request. We chose UCB over Thompson Sampling~\cite{agrawal2013thompson} because its deterministic score interacts more predictably with the Lagrangian penalty.\looseness=-1

Although each ingredient is well-established individually, combining them creates cross-component interactions. For example, geometric forgetting shrinks $A_a$ toward singularity over time, inflating LinUCB's confidence bonus and making expensive but uncertain arms more attractive. The BudgetPacer's dual variable $\lambda_t$ compensates online (Section~\ref{sec:budget_pacer}), but the upstream interaction between the forgetting rate $\gamma$ and exploration coefficient $\alpha$ determines exploration aggressiveness; these should be calibrated jointly, not in isolation. Appendix~\ref{sec:hparam_sweep} describes the Pareto knee-point procedure that resolves this.\looseness=-1

\subsection{Architecture Overview}
\label{sec:architecture}

ParetoBandit operates as a closed-loop routing system with two data paths.

\textbf{Synchronous inference path.} When a user prompt arrives, the feature extractor (Section~\ref{sec:context_repr}) produces a PCA-reduced context vector $x_t$. The \texttt{BanditRouter} scores every eligible model using the budget-augmented UCB rule (Section~\ref{sec:budget_pacer}) and dispatches the prompt to the highest-scoring model $a_t$.

\textbf{Asynchronous feedback path.} After inference, the observed reward $r_t$ and realized cost $c_t$ flow back. The bandit updates its per-arm sufficient statistics with geometric forgetting (Section~\ref{sec:adaptive_learning}), while the budget pacer updates $\lambda_t$ via the EMA-smoothed cost signal (Eqs.~\ref{eq:cost_ema}--\ref{eq:dual_update}). The context vector is cached at route time so that rewards arriving hours later (e.g., human Reinforcement Learning from Human Feedback (RLHF) labels) can still update the bandit without re-encoding the prompt. Algorithm~\ref{alg:paretobandit} consolidates the full per-step loop.

\subsection{Budget-Paced Arm Selection}
\label{sec:budget_pacer}

At each step $t$, the router selects model $a_t$ by maximizing a budget-augmented utility:
\begin{equation}
    a_t = \arg\max_{a \in \mathcal{A}_t} \Bigl[
        \underbrace{\hat\theta_a^\top x_t}_{\text{exploit}}
        +\; \underbrace{\alpha \sqrt{x_t^\top A_a^{-1} x_t}}_{\text{explore}}
        \;-\; \underbrace{(\lambda_c + \lambda_t) \,\tilde{c}_a}_{\text{cost penalty}}
    \Bigr]
    \label{eq:budget_ucb}
\end{equation}
The underbraced terms trade off exploitation of the current reward estimate, uncertainty-driven exploration (further inflated for stale arms via Eq.~\ref{eq:staleness_inflation}), and a cost penalty that discourages expensive models. The exploration rate $\alpha > 0$ scales the confidence bonus directly. The cost penalty has two additive components: a static weight $\lambda_c \geq 0$ (default 0.3) encoding the operator's baseline cost--quality preference from the first request ($\lambda_c = 0$ recovers quality-only routing), and a dynamic dual variable $\lambda_t \geq 0$ that starts at zero and rises only when recent spending exceeds $B$, providing closed-loop enforcement. After each request, a smoothed dual-ascent step adjusts $\lambda_t$:
\begin{align}
    \bar{c}_t &= (1 - \alpha_{\mathrm{ema}})\,\bar{c}_{t-1}
                 + \alpha_{\mathrm{ema}}\,c_t
    \label{eq:cost_ema} \\
    \lambda_{t+1} &= \bigl[\lambda_t + \eta\,(\bar{c}_t / B - 1)\bigr]_{0}^{\bar\lambda}
    \label{eq:dual_update}
\end{align}
where $\bar{c}_t$ is an EMA-smoothed cost signal ($\alpha_{\mathrm{ema}}{=}0.05$, half-life ${\approx}14$ requests), $\eta{=}0.05$ is the step size, and $[\cdot]_0^{\bar\lambda}$ projects onto $[0, \bar\lambda{=}5]$. The EMA prevents sawtooth oscillations from single expensive requests; normalizing the gradient by $B$ makes $\eta$ portfolio-independent. When $\bar{c}_t / B > 1$ (overspending), $\lambda_t$ rises, penalizing expensive models; when below, $\lambda_t$ falls, releasing the router to pursue quality.\looseness=-1

\begin{algorithm}[t]
\caption{ParetoBandit: Budget-Paced Non-Stationary Routing}
\label{alg:paretobandit}
\begin{algorithmic}[1]
\REQUIRE Arms $\mathcal{A}$; regularization $\lambda_0$; budget $B$; static cost penalty $\lambda_c$; exploration $\alpha$; forgetting $\gamma$; step size $\eta$; EMA $\alpha_{\mathrm{ema}}$; cap $\bar\lambda$
\STATE \textbf{Init:} $\forall\, a$: $A_a \!\leftarrow\! \lambda_0 I$, $b_a \!\leftarrow\! \mathbf{0}$ (or warmup both, \S\ref{sec:warmup}), $\hat\theta_a \!\leftarrow\! A_a^{-1} b_a$; $\lambda_t \!\leftarrow\! 0$; $\bar{c}_0 \!\leftarrow\! B$; $t \!\leftarrow\! 0$
\FOR{each request with context $x_t$}
\STATE \textbf{--- Arm Selection ---}
\IF{$\lambda_t > 0$}
\STATE $\mathcal{A}_t \leftarrow \{a \in \mathcal{A} : c_a \leq c_{\max}^{\mathcal{A}} / (1 {+} \lambda_t)\}$ \label{line:ceiling} \hfill $\triangleright$ Hard ceiling
\ELSE
\STATE $\mathcal{A}_t \leftarrow \mathcal{A}$
\ENDIF
\FOR{each $a \in \mathcal{A}_t$}
\STATE $\mathrm{dt}_a \leftarrow t - \max(\text{last\_upd}_a,\, \text{last\_play}_a)$ \hfill $\triangleright$ Exploration staleness
\STATE $v_a \leftarrow x_t^\top A_a^{-1} x_t / \max(\gamma^{\mathrm{dt}_a}, V_{\max}^{-1})$ \hfill $\triangleright$ Variance inflation
\STATE $s_a \leftarrow \hat\theta_a^\top x_t + \alpha\sqrt{v_a} - (\lambda_c + \lambda_t)\,\tilde{c}_a$ \label{line:ucb}
\ENDFOR
\STATE $a_t \leftarrow \arg\max_{a \in \mathcal{A}_t} s_a$ \hfill $\triangleright$ Random tiebreak
\STATE $t \leftarrow t + 1$;\; $\text{last\_play}_{a_t} \leftarrow t$
\STATE \textbf{--- Observe} reward $r_t$, cost $c_t$ \textbf{---}
\STATE \textbf{--- Reward Update (Geometric Forgetting) ---}
\STATE $\mathrm{dt}' \leftarrow t - \text{last\_upd}_{a_t}$ \hfill $\triangleright$ Decay staleness (statistics age)
\STATE $A_{a_t} \leftarrow \gamma^{\mathrm{dt}'}\, A_{a_t}$;\; $b_{a_t} \leftarrow \gamma^{\mathrm{dt}'}\, b_{a_t}$ \label{line:forget} \hfill $\triangleright$ Decay stale data
\STATE $A_{a_t}^{-1} \leftarrow A_{a_t}^{-1} / \gamma^{\mathrm{dt}'}$ \hfill $\triangleright$ $O(d^2)$ scalar op
\STATE $A_{a_t} \leftarrow A_{a_t} + x_t x_t^\top$;\; $b_{a_t} \leftarrow b_{a_t} + r_t\, x_t$
\STATE Update $A_{a_t}^{-1}$ via Sherman--Morrison \hfill $\triangleright$ $O(d^2)$
\STATE $\hat\theta_{a_t} \leftarrow A_{a_t}^{-1}\, b_{a_t}$;\; $\text{last\_upd}_{a_t} \leftarrow t$
\STATE \textbf{--- Budget Pacer Dual Update ---}
\STATE $\bar{c}_t \leftarrow (1 {-} \alpha_{\mathrm{ema}})\,\bar{c}_{t-1} + \alpha_{\mathrm{ema}}\,c_t$ \hfill $\triangleright$ Eq.~\ref{eq:cost_ema}
\STATE $\lambda_{t+1} \leftarrow [\lambda_t + \eta(\bar{c}_t / B - 1)]_{0}^{\bar\lambda}$ \label{line:dual} \hfill $\triangleright$ Eq.~\ref{eq:dual_update}
\ENDFOR
\end{algorithmic}
\end{algorithm}

\paragraph{Reward model and parameter estimation.}
LinUCB models the expected reward as $\mathbb{E}[r_t \mid x_t, a] = \theta_a^{*\top} x_t$~\cite{li2010contextual}. The estimate $\hat\theta_a = A_a^{-1} b_a$ is the ridge regression solution, where
\begin{equation}
    A_a = \lambda_0 I + \textstyle\sum_{\tau:\, a_\tau = a} x_\tau x_\tau^\top,\;
    b_a = \textstyle\sum_{\tau:\, a_\tau = a} r_\tau\, x_\tau
    \label{eq:sufficient_stats}
\end{equation}
are the per-arm \emph{sufficient statistics}, the running aggregates that fully summarize past observations for parameter estimation, eliminating the need to store raw history. At cold start ($A_a = \lambda_0 I$), the confidence bonus is maximal, driving exploration; as observations accumulate, the confidence set shrinks toward exploitation.

\paragraph{Log-normalized cost.}
The term $\tilde{c}_a \in [0,1]$ is a log-normalized unit cost:
\begin{equation}
    \tilde{c}_a = \frac{\log c_a - \log c_{\mathrm{floor}}}{\log c_{\mathrm{ceil}} - \log c_{\mathrm{floor}}}
    \label{eq:cost_norm}
\end{equation}
where $c_{\mathrm{floor}}{=}\$0.0001$ and $c_{\mathrm{ceil}}{=}\$0.10$ per 1k tokens are fixed market bounds. The logarithmic scale compresses the $\ppPriceRangeX{\times}$ cost range into a bounded penalty commensurate with the $[0,1]$ reward scale (validated in Appendix~\ref{sec:cost_heuristic_validation}).

\paragraph{Two-layer enforcement (Adaptive mode).}
The budget pacer combines two mechanisms simultaneously:
\begin{itemize}[nosep,leftmargin=*]
  \item \textbf{Hard ceiling (safety net):} When $\lambda_t > 0$, the candidate set $\mathcal{A}_t$ excludes models whose blended price exceeds a dynamic ceiling $c_{\max}^{\mathcal{A}} / (1 + \lambda_t)$, where $c_{\max}^{\mathcal{A}} = \max_{a \in \mathcal{A}} c_a$ is the portfolio's most expensive rate. This acts as a circuit breaker, preventing catastrophic overspend on any single request.
  \item \textbf{Soft penalty (optimizer):} The dual penalty $\lambda_t \tilde{c}_a$ biases the UCB score toward cheaper models proportionally to their cost, allowing fine-grained, context-dependent routing within the surviving candidate set.
\end{itemize}
The hard ceiling provides per-request worst-case cost bounds independent of $\lambda_t$; the soft penalty provides smooth, quality-maximizing allocation within those bounds.

\paragraph{Theoretical status.}
Eq.~\ref{eq:budget_ucb} is a Lagrangian relaxation of Eq.~\ref{eq:objective}, but four departures from the classical BwK analysis~\cite{badanidiyuru2013bwk}---rate constraint, EMA smoothing, hard ceiling, and log-normalized costs---void the formal $\tilde{O}(\sqrt{T})$ bounds. The smoothed dual framework of \citet{balseiro2019learning} accommodates EMA-like averaging but assumes i.i.d.\ contexts and a convex program structure; our non-stationary setting with geometric forgetting violates both conditions, so their formal guarantees do not apply either. Three properties nonetheless hold by construction: (1)~$\lambda_t \in [0, \bar\lambda]$, (2)~staleness inflation ensures re-exploration of neglected arms, and (3)~when $\lambda_t > 0$, the hard ceiling caps the cost of any single request. Budget compliance is therefore validated empirically: across all conditions in Section~\ref{sec:evaluation}, per-request cost never exceeds the ceiling by more than ${\sim}4\%$.

\subsection{Non-Stationary Adaptation via Geometric Forgetting}
\label{sec:adaptive_learning}

Geometric forgetting enables ParetoBandit to override stale estimates within its effective memory window, whereas a standard LinUCB ($\gamma{=}1.0$) accumulates all historical data equally and may require hundreds or thousands of new observations to do the same. Concretely, ParetoBandit applies a forgetting factor $\gamma \in (0, 1]$ directly to the per-arm sufficient statistics. Before incorporating the new observation $(x_t, a_t, r_t)$, the design matrix and reward accumulator for the chosen arm $a_t$ are exponentially discounted:
\begin{align}
    A_{a_t} &\leftarrow \gamma^{\mathrm{dt}} \, A_{a_t} + x_t x_t^\top
    \label{eq:forget_A} \\
    b_{a_t} &\leftarrow \gamma^{\mathrm{dt}} \, b_{a_t} + r_t \, x_t
    \label{eq:forget_b}
\end{align}
where $\mathrm{dt} = t - \text{last\_update}_{a_t}$ is the number of intervening requests since $a_t$'s last update. This batched exponentiation avoids redundant per-step multiplications for arms that were not selected. The exponential weighting~\cite{garivier2011switching} gives observations an effective half-life of $\ln 2 / (1 - \gamma)$ steps and an e-folding time of $1 / (1 - \gamma)$. For $\gamma{=}\ppGamma$, the e-folding time is ${\approx}\ppEffMemSteps$ steps: an observation from \ppEffMemSteps{} steps ago retains weight $\gamma^{\ppEffMemSteps} \approx e^{-1} \approx 0.37$, and after ${\sim}1{,}000$ steps the prior's contribution is reduced to $\gamma^{1000} \approx \ppGammaDecayOneK$.\looseness=-1

\paragraph{Staleness-based variance inflation.}
Geometric forgetting passively increases $x_t^\top A_a^{-1} x_t$ for idle arms, but this may be too slow when the environment shifts. The router explicitly inflates the UCB variance using the \emph{exploration staleness} $\mathrm{dt}_a = t - \max(\text{last\_update}_a,\, \text{last\_played}_a)$, which counts from whichever happened more recently---a statistics update or a play---so that arms dispatched but still awaiting asynchronous rewards are not prematurely re-explored:
\begin{equation}
    v_a = \frac{x_t^\top A_a^{-1} x_t}{\max\!\bigl(\gamma^{\mathrm{dt}_a},\; V_{\max}^{-1}\bigr)}
    \label{eq:staleness_inflation}
\end{equation}
Without a cap, the inflation grows without bound as $\mathrm{dt}_a \to \infty$, and the exploration bonus $\alpha\sqrt{v_a}$ eventually dominates any finite cost penalty, forcing selection of stale expensive arms regardless of budget pressure. A capacity cap $V_{\max} = 200$ bounds the worst-case inflation to $\sqrt{V_{\max}} \approx 14{\times}$ the uninflated bonus, large enough to reliably trigger re-exploration yet bounded enough that the cost signal retains meaningful influence on arm selection.

\paragraph{Interaction with the budget pacer.}
Geometric forgetting interacts with budget pacing through a feedback loop. As $\gamma$ shrinks $A_a$ over time, the confidence bonus grows, increasing the probability of selecting expensive arms that happen to be uncertain. This drives up $\lambda_t$, which penalizes those arms, partially counteracting the inflation. The EMA smoothing in the dual update (Eq.~\ref{eq:cost_ema}) further dampens the resulting oscillations. The parameter $\gamma$ thus controls a bias-variance tradeoff between \emph{stability} (trusting historical evidence) and \emph{plasticity} (responding quickly to drift). The forgetting rate is therefore selected jointly with $\alpha$ (Appendix~\ref{sec:hparam_sweep}).\looseness=-1

\subsection{Offline-to-Online Warmup Priors}
\label{sec:warmup}

Without warmup priors, the router begins with uninformative $A_a = \lambda_0 I$, $b_a = \mathbf{0}$, and needs to explore each model on diverse prompts before learning which model excels where. ParetoBandit eliminates this cold start by loading offline sufficient statistics $(A_a^{\mathrm{off}}, b_a^{\mathrm{off}})$ fitted on historical prompt-reward data. A tunable prior strength $n_\mathit{eff}$ controls how many pseudo-observations the offline data contribute:
\begin{equation}
    s = \frac{n_\mathit{eff}}{A_a^{\mathrm{off}}[d, d]}
    \label{eq:prior_scale}
\end{equation}
where $A_a^{\mathrm{off}}[d, d]$ is the total precision mass in the bias direction. The scaled prior is regularized with a mean-preserving correction:
\begin{align}
    A_a &\leftarrow s \cdot A_a^{\mathrm{off}} + \lambda_0 I
    \label{eq:prior_A} \\
    b_a &\leftarrow s \cdot b_a^{\mathrm{off}} + \lambda_0\, \hat\theta_a^{\mathrm{off}}
    \label{eq:prior_b}
\end{align}
The $\lambda_0 \hat\theta_a^{\mathrm{off}}$ term in $b_a$ prevents $\lambda_0 I$ regularization from shrinking the posterior mean toward zero, ensuring $A_a^{-1} b_a \approx \hat\theta_a^{\mathrm{off}}$ at the desired confidence level. For models absent from the offline data, a heuristic prior places $n_\mathit{eff}$ pseudo-observations at isotropic uncertainty with a bias-only reward prediction.

Geometric forgetting ensures priors decay naturally (Section~\ref{sec:adaptive_learning}), so steady-state quality is determined by online evidence regardless of initialization. Priors are beneficial in deployment comparisons but not necessary---cold-start convergence is demonstrated in Section~\ref{sec:eval_onboarding} and ablated in Appendices~\ref{sec:warmup_ablation}--\ref{appendix:prior_mismatch}.

\subsection{Computational Efficiency}
\label{sec:efficiency}

The router sits on the critical path of every LLM request. ParetoBandit maintains a cached $A_a^{-1}$ and applies Sherman--Morrison rank-1 updates in $O(d^2)$; geometric discounting adds only a scalar division $A_a^{-1} \leftarrow A_a^{-1}/\gamma$. Arm selection requires one $O(d^2)$ quadratic form per arm for the UCB bonus plus an $O(d)$ dot product for the reward estimate. With $K{=}3$ and $d{=}26$, the full route-and-update cycle completes in $\ppPBTotalMedianUs\,\mu$s (p50), sustaining ${\sim}\ppPBThroughput$\,req/s on a single CPU core. Including prompt embedding and PCA projection, end-to-end routing latency is $\ppEteTotalMedianMs$\,ms---under $\ppEteRoutePctOfTotal\%$ of a typical ${\sim}1$\,s inference call~\cite{ganglani2026llmlatency}. PCA reduction and inverse caching together yield a ${\sim}\ppPcaThroughputGainX{\times}$ throughput gain over a raw-dimension baseline (Appendix~\ref{sec:latency_benchmark}).

\subsection{Runtime Portfolio Management}
\label{sec:production_infra}

Production model portfolios are not static. ParetoBandit supports runtime model additions and removals via \texttt{add\_arm()} and \texttt{delete\_arm()} without restarting the router. A newly added model is initialized with either a heuristic prior or uninformative covariance. When existing arms have strong learned posteriors and $\alpha$ is small, the UCB exploration bonus alone may be insufficient to trigger natural exploration of a cold-start arm. Section~\ref{sec:eval_onboarding} addresses this with a short forced-exploration burn-in (20~pulls routed unconditionally to the new arm), after which the bandit has enough evidence to discriminate and UCB selection takes over. The context vector $x_t$ is cached at route time so that feedback arriving later (synchronous judge scores, asynchronous RLHF labels, or batch metrics) can update the bandit without re-encoding the prompt, with both in-memory and SQLite-backed storage backends.

\section{Evaluation}
\label{sec:evaluation}

We evaluate ParetoBandit across four experiments: stationary budget pacing, budget pacing under cost drift, resilience to silent quality degradation, and cold-start model onboarding. All experiments use 20~seeds and report 95\% bootstrap confidence intervals.

\subsection{Experimental Setup}
\label{sec:eval_setup}

\paragraph{Data and models.}
We collect 11,983 prompts from nine public NLP benchmarks (MMLU~\cite{hendrycks2021measuring}, GSM8K~\cite{cobbe2021gsm8k}, HellaSwag~\cite{zellers2019hellaswag}, BIG-Bench Hard~\cite{suzgun2022challenging}, ARC-Challenge~\cite{clark2018arc}, OpenBookQA~\cite{mihaylov2018suit}, WinoGrande~\cite{sakaguchi2020winogrande}, TruthfulQA~\cite{lin2022truthfulqa}, MBPP~\cite{austin2021program}) spanning reasoning, math, code synthesis, commonsense, and factual knowledge. For each prompt, all $K{=}3$ models generate responses that are judged offline, producing a full reward--cost matrix. Prompts are partitioned into three disjoint splits stratified by source: train ($n{=}8{,}374$) for fitting warmup priors, val ($n{=}1{,}785$) for online hyperparameter tuning, and test ($n{=}1{,}824$) for evaluation. The $K{=}3$ portfolio spans a $\bpPriceRangeX{\times}$ cost range (Table~\ref{tab:setup}).

\paragraph{Quality evaluation.}
Each response is scored by DeepSeek-R1~\cite{deepseekai2025deepseekr1} on three continuous dimensions: reasoning quality (weight 0.4), instruction following (0.3), and communication quality (0.3), yielding a composite reward $r_t \in [0, 1]$. R1 produces the largest inter-model reward gaps among the judges we evaluated; Appendix~\ref{sec:judge_robustness} validates that its routing decisions capture ${\geq}97\%$ of any alternative judge's oracle reward.

\paragraph{Baselines.}
Two conditions represent increasing routing sophistication, all sharing warmup priors ($n_\mathit{eff}{=}\bdNeff$, $\alpha{=}\ppAlpha$):
\begin{enumerate}[nosep,leftmargin=*]
  \item \textbf{Naive Bandit ($\gamma{=}1.0$):} LinUCB with online learning but infinite memory and a static cost penalty.
  \item \textbf{ParetoBandit ($\gamma{=}\ppGamma$):} Geometric forgetting with warmup priors and an active BudgetPacer.
\end{enumerate}
Experiment~2 additionally includes a \textbf{Recalibrated Bandit} (oracle knowledge of price changes) and a \textbf{Forgetting Bandit} ($\gamma{=}\ppGamma$, no pacer) as ablations. Hyperparameters are selected via the Pareto knee-point procedure described in Appendix~\ref{sec:hparam_sweep}.

\noindent\textbf{Non-stationary protocol.} Experiments 2--3 follow a three-phase stress-test: normal operation (608 prompts), abrupt perturbation (608 prompts), and recovery (608 prompts). Phase~3 reuses Phase~1 prompts for a controlled within-subject comparison.

\vspace{-6pt}
\begin{table}[!htbp]
\centering
\caption{Model portfolio and budget targets. The $\bpPriceRangeX{\times}$ cost spread stresses the router's quality--cost tradeoff. Three budgets span the full operating range.
}
\label{tab:setup}
\small
\begin{tabular}{@{}llr@{}}
\toprule
\textbf{Model} & \textbf{Tier} & \textbf{Cost (\$/req)} \\
\midrule
Llama-3.1-8B   & Budget   & $\$2.9{\times}10^{-5}$ \\
Mistral-Large  & Mid-cost & $\$5.3{\times}10^{-4}$ \\
Gemini-2.5-Pro & Frontier & $\$1.5{\times}10^{-2}$ \\
\midrule
\textbf{Budget} & \textbf{Target $B$} & \textbf{Regime} \\
\midrule
Tight    & $\$3.0{\times}10^{-4}$ & Llama-dominant \\
Moderate & $\$6.6{\times}10^{-4}$ & Llama--Mistral mix \\
Loose    & $\$1.9{\times}10^{-3}$ & Selective Gemini \\
\bottomrule
\end{tabular}
\end{table}

\subsection{Stationary Budget Pacing}
\label{sec:eval_stationary}

\begin{figure*}[!tbp]
\centering
\includegraphics[width=\textwidth]{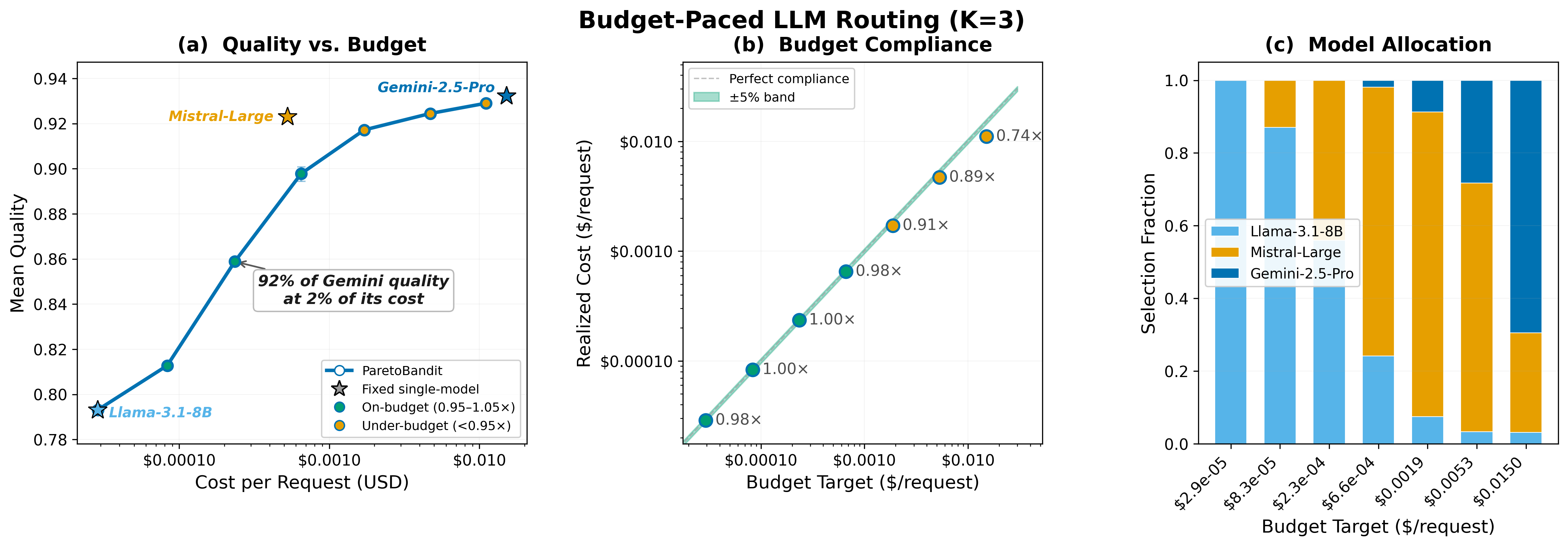}
\caption{\input{experiments/01_stationary_budget_pacing/figure_caption_pareto.tex}}
\label{fig:pareto}
\vspace{4pt}
\end{figure*}

We first ask: \emph{can the router accept a dollar budget ceiling and automatically maximize quality beneath it, filling the quality--cost gaps between fixed single-model baselines?}

Figure~\ref{fig:pareto}a shows the quality--cost Pareto frontier on the test set. Each fixed model occupies a single point on this frontier, plotted as (mean cost per request, mean quality score): Llama at ($\bpFixedLlamaCost$, $\bpFixedLlamaReward$), Mistral at ($\bpFixedMistralCost$, $\bpFixedMistralReward$), Gemini at ($\bpFixedGeminiCost$, $\bpFixedGeminiReward$). The BudgetPacer traces a continuous curve through these points by dynamically mixing models based on prompt context and budget pressure. At a budget of $\bpAnnotBudget$/req, the router achieves $\bpAnnotQualPct$\%\,[$\bpAnnotQualPctCILo$, $\bpAnnotQualPctCIHi$] of Gemini quality at $\bpAnnotCostPct$\% of its cost by blending $\bpAnnotLlamaFrac$\% Llama and $\bpAnnotMistralFrac$\% Mistral. For binding ceilings, utilisation ranges from $\bpUtilBindingLow$ to $\bpUtilBindingHigh$ (Figure~\ref{fig:pareto}b). When the ceiling is high enough that cost never binds, the cost penalty $\lambda_t$ decays to zero and the router reduces to pure quality maximization, recovering $\bpCeilingOraclePct$\%\,[$\bpCeilingOraclePctCILo$, $\bpCeilingOraclePctCIHi$] of an oracle that selects the highest-scoring model for each prompt ($\bpOracleReward$). In practice, this turns model selection from a discrete choice among $K$ fixed operating points into a continuous budget dial: the operator sets a dollar ceiling, and the router discovers the best quality mix beneath it (Figure~\ref{fig:pareto}c).

\subsection{Budget Pacing under Cost Drift}
\label{sec:eval_nonstationary}

\begin{table}[!tbp]
\centering
\caption{Budget compliance under cost drift (20~seeds).
Each cell shows realised cost as a multiple of the ceiling ($1.00{\times}$ = at ceiling); 95\% bootstrap CIs are reported inline for key claims in the text.
\textbf{Bold}: within 5\% of ceiling.
}
\label{tab:budget_compliance}
\small
\resizebox{\columnwidth}{!}{%
\begin{tabular}{@{}llccc@{}}
\toprule
\textbf{Budget} & \textbf{Condition}
  & \textbf{Phase~1} & \textbf{Phase~2} & \textbf{Phase~3} \\
\midrule
\multirow{4}{*}{\shortstack[l]{Tight\\($\$3.0{\times}10^{-4}$)}}
  & Naive Bandit        & $\mathbf{\bdNaiveTightPhaseOneRatio}$ & $\bdNaiveTightPhaseTwoRatio$ & $\bdNaiveTightPhaseThreeRatio$ \\
  & Recalibrated        & $\mathbf{\bdRecalTightPhaseOneRatio}$ & $\mathbf{\bdRecalTightPhaseTwoRatio}$ & $\bdRecalTightPhaseThreeRatio$ \\
  & Forgetting Bandit   & $\bdForgetTightPhaseOneRatio$ & $\bdForgetTightPhaseTwoRatio$ & $\bdForgetTightPhaseThreeRatio$ \\
  & \textbf{ParetoBandit} & $\mathbf{\bdParetoBanditTightPhaseOneRatio}$ & $\bdParetoBanditTightPhaseTwoRatio$ & $\mathbf{\bdParetoBanditTightPhaseThreeRatio}$ \\[3pt]
\multirow{4}{*}{\shortstack[l]{Moderate\\($\$6.6{\times}10^{-4}$)}}
  & Naive Bandit        & $\bdNaiveModPhaseOneRatio$ & $\bdNaiveModPhaseTwoRatio$ & $\bdNaiveModPhaseThreeRatio$ \\
  & Recalibrated        & $\bdRecalModPhaseOneRatio$ & $\bdRecalModPhaseTwoRatio$ & $\bdRecalModPhaseThreeRatio$ \\
  & Forgetting Bandit   & $\bdForgetModPhaseOneRatio$ & $\bdForgetModPhaseTwoRatio$ & $\bdForgetModPhaseThreeRatio$ \\
  & \textbf{ParetoBandit} & $\mathbf{\bdParetoBanditModPhaseOneRatio}$ & $\bdParetoBanditModPhaseTwoRatio$ & $\mathbf{\bdParetoBanditModPhaseThreeRatio}$ \\[3pt]
\multirow{4}{*}{\shortstack[l]{Loose\\($\$1.9{\times}10^{-3}$)}}
  & Naive Bandit        & $\bdNaiveLoosePhaseOneRatio$ & $\bdNaiveLoosePhaseTwoRatio$ & $\bdNaiveLoosePhaseThreeRatio$ \\
  & Recalibrated        & $\bdRecalLoosePhaseOneRatio$ & $\bdRecalLoosePhaseTwoRatio$ & $\bdRecalLoosePhaseThreeRatio$ \\
  & Forgetting Bandit   & $\bdForgetLoosePhaseOneRatio$ & $\bdForgetLoosePhaseTwoRatio$ & $\bdForgetLoosePhaseThreeRatio$ \\
  & \textbf{ParetoBandit} & $\bdParetoBanditLoosePhaseOneRatio$ & $\bdParetoBanditLoosePhaseTwoRatio$ & $\bdParetoBanditLoosePhaseThreeRatio$ \\[3pt]
\bottomrule
\end{tabular}%
}
\end{table}

We now evaluate whether the pacer automatically exploits a mid-stream price drop while maintaining compliance. Phase~1 uses normal pricing; Phase~2 drops Gemini-2.5-Pro's pricing to \$0.10/M tokens (normalized cost $\tilde{c} \approx 0$); Phase~3 restores original pricing.

\noindent\textbf{Budget compliance is the key differentiator.}
Table~\ref{tab:budget_compliance} summarises cost/ceiling ratios across all conditions and phases. ParetoBandit is the only condition that reliably stays within the ceiling in Phase~1 ($\bdParetoBanditLoosePhaseOneRatio$\,[$\bdParetoBanditLoosePhaseOneRatioCILo$, $\bdParetoBanditLoosePhaseOneRatioCIHi$] to $\bdParetoBanditTightPhaseOneRatio$\,[$\bdParetoBanditTightPhaseOneRatioCILo$, $\bdParetoBanditTightPhaseOneRatioCIHi$]) and recovers compliance in Phase~3 ($\bdParetoBanditLoosePhaseThreeRatio$\,[$\bdParetoBanditLoosePhaseThreeRatioCILo$, $\bdParetoBanditLoosePhaseThreeRatioCIHi$] to $\bdParetoBanditTightPhaseThreeRatio$\,[$\bdParetoBanditTightPhaseThreeRatioCILo$, $\bdParetoBanditTightPhaseThreeRatioCIHi$]). The critical ablation is the Forgetting Bandit ($\gamma{=}\ppGamma$, no pacer), which shares ParetoBandit's learning dynamics yet produces consistently poor compliance ($\bdForgetTightPhaseOneRatio$\,[$\bdForgetTightPhaseOneRatioCILo$, $\bdForgetTightPhaseOneRatioCIHi$] at tight Phase~1, $\bdForgetTightPhaseThreeRatio$\,[$\bdForgetTightPhaseThreeRatioCILo$, $\bdForgetTightPhaseThreeRatioCIHi$] at tight Phase~3), confirming that the BudgetPacer---not the forgetting factor---drives budget control.

\begin{figure}[!tbp]
\centering
\includegraphics[width=\columnwidth]{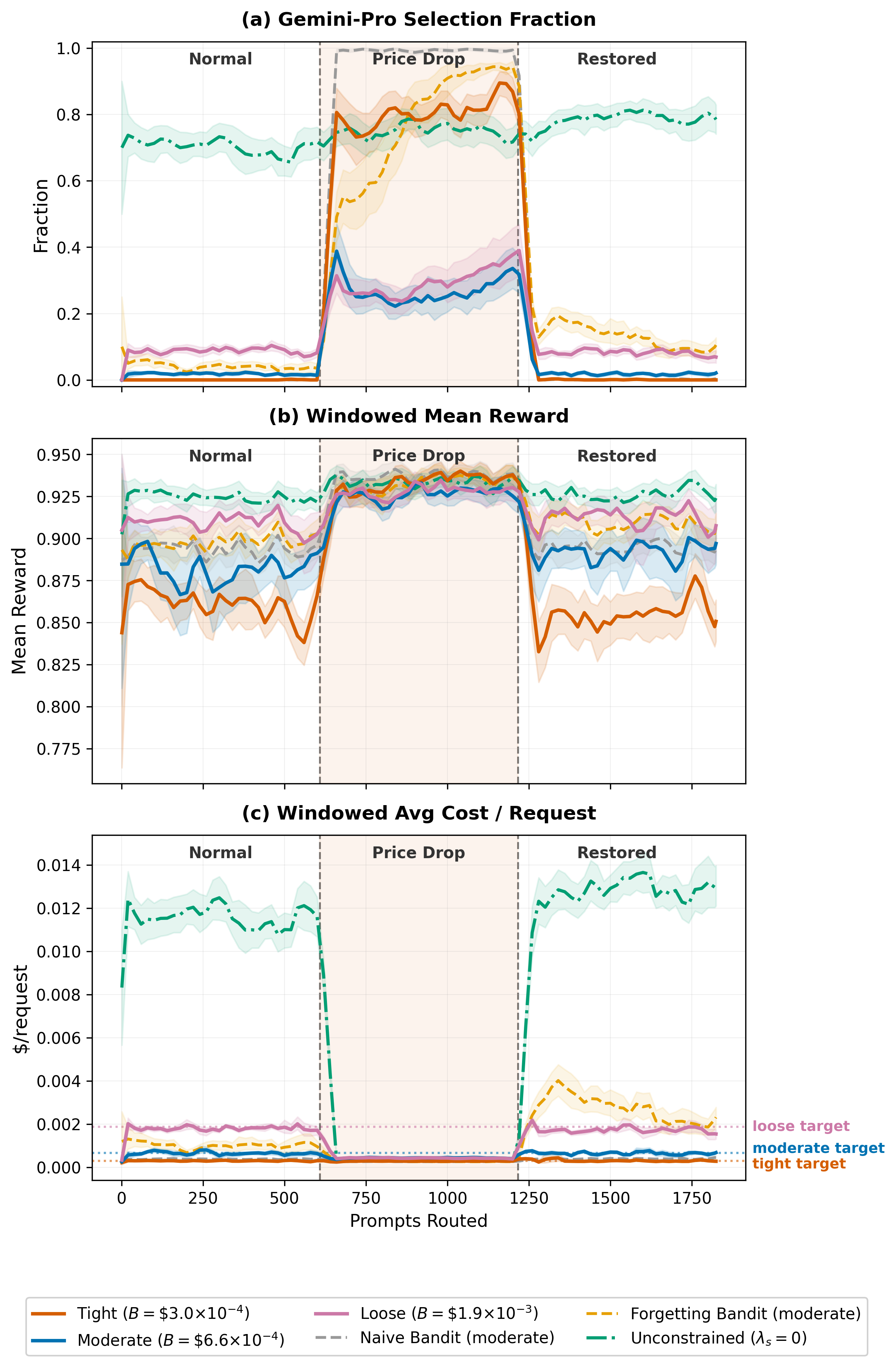}
\caption{\input{experiments/02_budget_plus_drift/figure_caption_dynamics.tex}}
\label{fig:adaptation_dynamics}
\end{figure}

\noindent\textbf{Exploiting the price drop.}
When Gemini becomes nearly free in Phase~2, the BudgetPacer detects the cost change via its EMA signal. All three budget regimes show significant reward lifts (tight $\Delta{=}+\bdParetoBanditTightRewardLift$\,[$\bdParetoBanditTightRewardLiftCILo$, $\bdParetoBanditTightRewardLiftCIHi$]; loose $\Delta{=}+\bdParetoBanditLooseRewardLift$\,[$\bdParetoBanditLooseRewardLiftCILo$, $\bdParetoBanditLooseRewardLiftCIHi$]), with the tight budget exhibiting the largest gain because its Phase~1 constraint was most binding. Figure~\ref{fig:adaptation_dynamics} shows the dual-variable dynamics: $\lambda_t$ decays as costs fall below the ceiling, increasing Gemini adoption; in Phase~3, $\lambda_t$ rises and compliance recovers. This full round-trip---exploit, then recover---demonstrates bidirectional adaptation without operator intervention.

\subsection{Silent Quality Degradation}
\label{sec:eval_catastrophic}

\begin{figure}[!tbp]
\centering
\setlength{\abovecaptionskip}{2pt}
\includegraphics[width=\columnwidth]{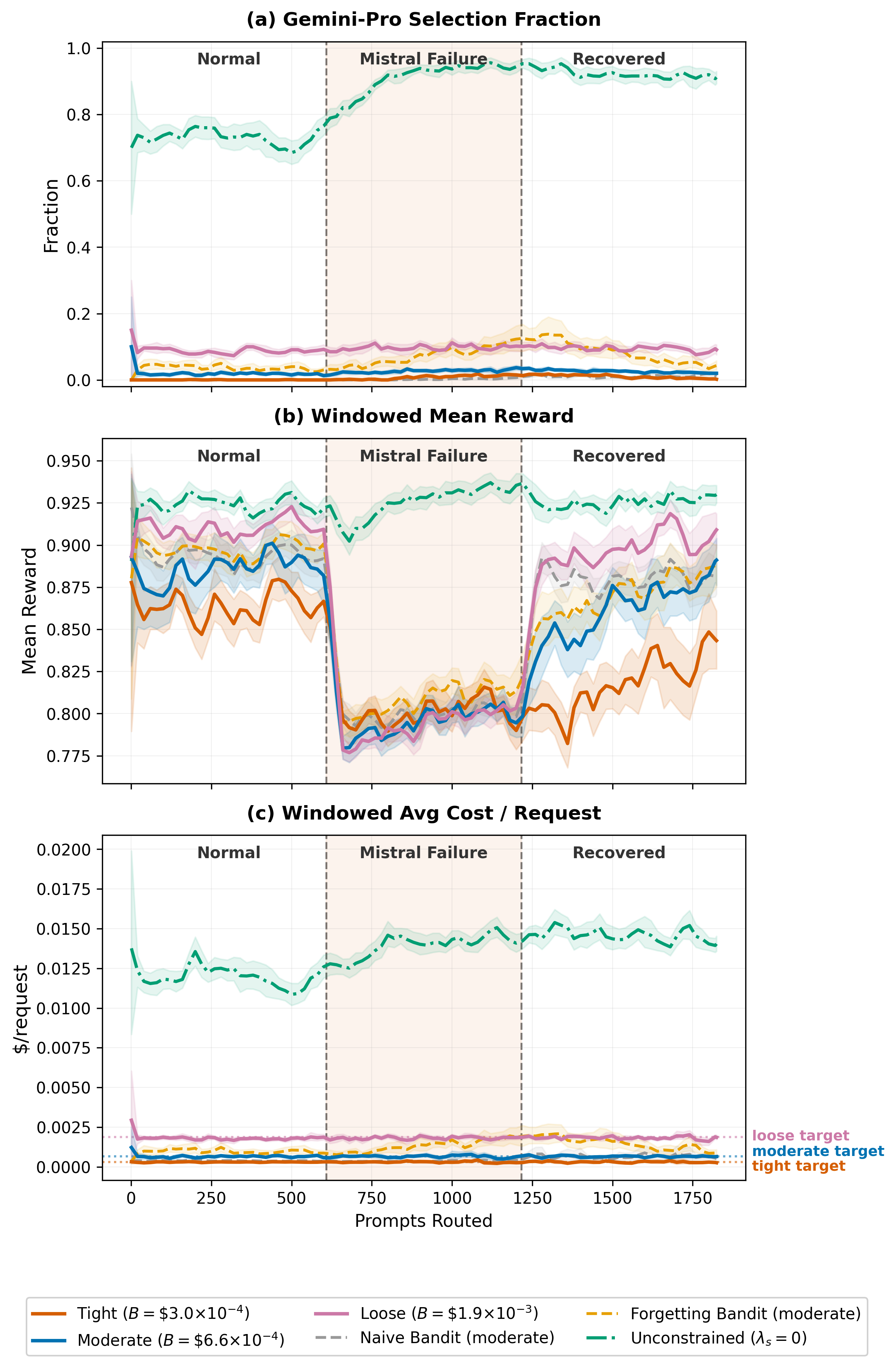}
\caption{\input{experiments/03_catastrophic_failure/figure_caption_dynamics.tex}}
\label{fig:catastrophic_failure}
\end{figure}

We stress-test against a silent quality regression: Mistral-Large's reward drops to $\cfFailureReward$ (${\sim}\ppCfDegPct$\% below normal) while its API continues to respond and charge at normal rates.  The cost signal provides no warning; only the reward reveals the problem.  Phase~3 restores normal quality, testing whether the router re-discovers the recovered model.

\noindent\textbf{Results (Figure~\ref{fig:catastrophic_failure}).}  ParetoBandit detects the quality drop purely through the reward signal, reducing Mistral allocation from $\cfParetoBanditModMistralPhaseOne$\% to $\cfParetoBanditModMistralPhaseTwo$\% in Phase~2 at moderate budget.  In Phase~3, geometric forgetting down-weights the corrupted estimates and staleness-driven exploration begins re-exploring Mistral, raising allocation to $\cfParetoBanditModMistralPhaseThree$\% and recovering Phase~3 reward to $\cfParetoBanditModPhaseThreeReward$\,[$\cfParetoBanditModPhaseThreeRewardCILo$, $\cfParetoBanditModPhaseThreeRewardCIHi$] (vs.\ Phase~1 $\cfParetoBanditModPhaseOneReward$\,[$\cfParetoBanditModPhaseOneRewardCILo$, $\cfParetoBanditModPhaseOneRewardCIHi$], recovery ratio $\cfParetoBanditModRecoveryMean$\,[$\cfParetoBanditModRecoveryCILo$, $\cfParetoBanditModRecoveryCIHi$]).  Budget compliance holds throughout ($\cfParetoBanditLoosePhaseOneRatio$\,[$\cfParetoBanditLoosePhaseOneRatioCILo$, $\cfParetoBanditLoosePhaseOneRatioCIHi$] to $\cfParetoBanditTightPhaseOneRatio$\,[$\cfParetoBanditTightPhaseOneRatioCILo$, $\cfParetoBanditTightPhaseOneRatioCIHi$] across all nine phase--budget cells).  The unconstrained baseline is unaffected (Phase~2 reward $\cfUncPhaseTwoReward$\,[$\cfUncPhaseTwoRewardCILo$, $\cfUncPhaseTwoRewardCIHi$] vs.\ Phase~1 $\cfUncPhaseOneReward$\,[$\cfUncPhaseOneRewardCILo$, $\cfUncPhaseOneRewardCIHi$]) but incurs a $\cfUncCostSpikePct$\% cost increase from over-allocating to Gemini.  This ${\sim}\ppCfDegPct$\% degradation lies within the system's recovery envelope; Appendix~\ref{appendix:recovery_limit} characterises the threshold beyond which recovery requires an extended horizon.

\subsection{Cold-Start Model Onboarding}
\label{sec:eval_onboarding}

\begin{figure*}[!tbp]
\centering
\includegraphics[width=\textwidth]{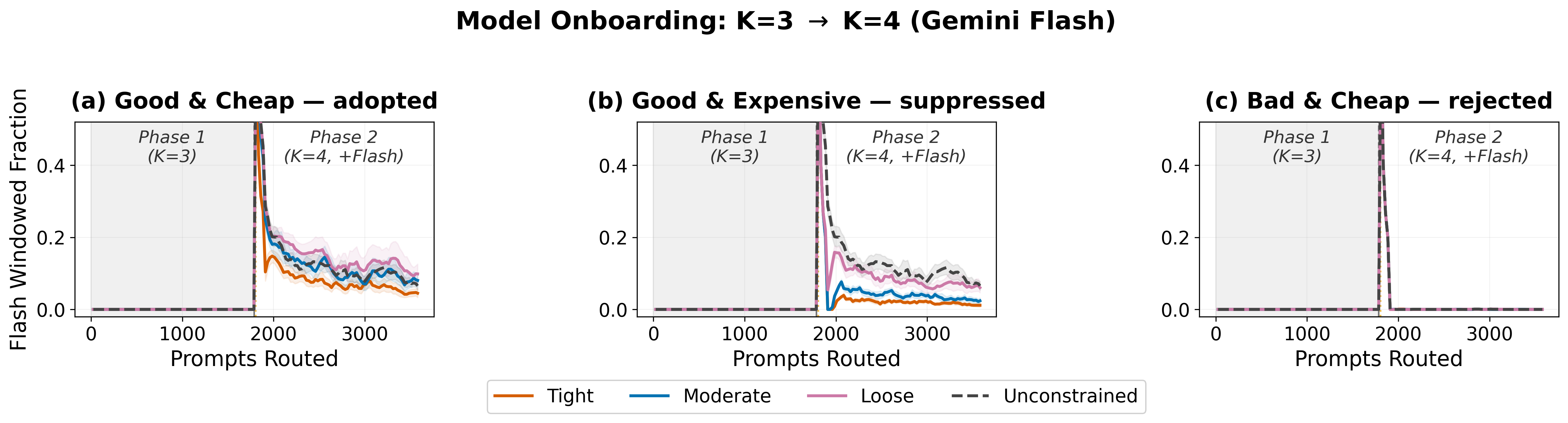}
\vspace{-10pt}
\caption{\input{experiments/04_model_onboarding/figure_caption_onboarding.tex}}
\label{fig:model_onboarding}
\vspace{10pt}
\end{figure*}

\begin{figure}[!tbp]
\centering
\includegraphics[width=0.78\columnwidth]{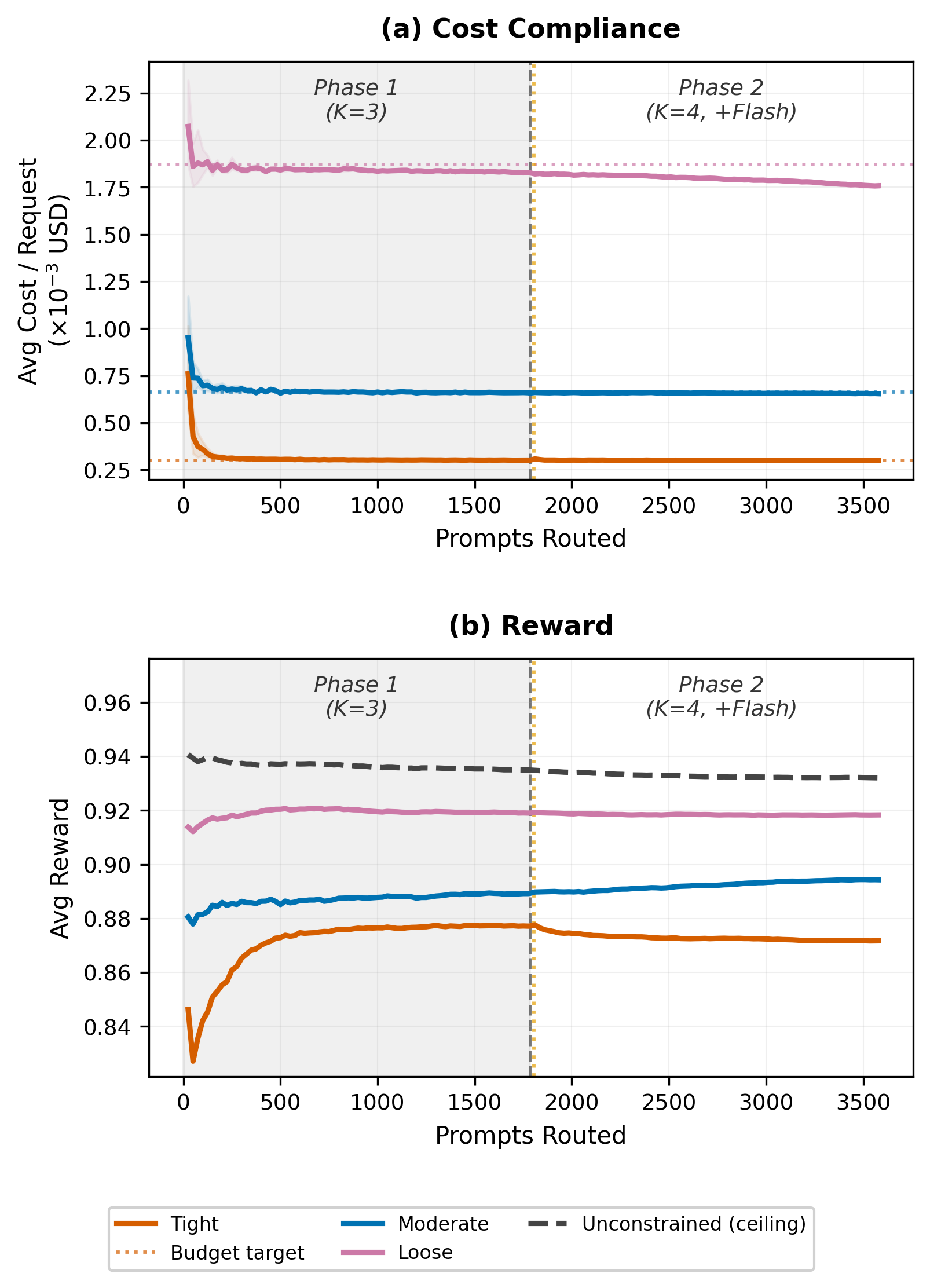}
\caption{\input{experiments/04_model_onboarding/figure_caption_cost.tex}}
\label{fig:model_onboarding_cost}
\vspace{4pt}
\end{figure}

After Phase~1 learning on the $K{=}3$ portfolio, Gemini-2.5-Flash is added as a fourth arm via \texttt{register\_model()} with no warmup priors. Figure~\ref{fig:model_onboarding} shows Flash adoption across three scenario--budget combinations. In the Good \& Cheap scenario (Figure~\ref{fig:model_onboarding}a), all 80 trials (20~seeds $\times$ 4~budget tiers) achieve sustained adoption within ${\sim}\ppAdoptionSteps$ steps. Budget determines the equilibrium share, not whether adoption occurs: loose budgets settle at ${\sim}\moGoodCheapPBLooseFlashFinalPct$\%\,[$\moGoodCheapPBLooseFlashFinalPctCILo$, $\moGoodCheapPBLooseFlashFinalPctCIHi$] Flash share while tight budgets plateau at ${\sim}\moGoodCheapPBTightFlashFinalPct$\%\,[$\moGoodCheapPBTightFlashFinalPctCILo$, $\moGoodCheapPBTightFlashFinalPctCIHi$]. The BudgetPacer maintains compliance throughout the $K{=}3 \to K{=}4$ transition without reconfiguration (Figure~\ref{fig:model_onboarding_cost}). Crucially, the bandit discriminates rather than blindly adopting: expensive models are budget-gated (Figure~\ref{fig:model_onboarding}b) and bad models are rejected after a bounded burn-in (Figure~\ref{fig:model_onboarding}c).

The trade-off relative to offline characterisation is that ${\sim}1$\% of Phase~2 requests serve as online exploration, but adoption is automatically budget-gated and requires no paired preference data.

\vspace{12pt}
\section{Related Work}
\label{sec:related}
\vspace{10pt}

\begingroup
\setlength{\parskip}{2pt}

\paragraph{Supervised and static routers.}
The simplest cost-aware strategy is cascading: FrugalGPT~\cite{chen2023frugalgpt} queries models cheapest-first and stops once a quality threshold is met. Much subsequent work trains a single-dispatch classifier offline, differing mainly in the training signal---preference pairs~\cite{ong2024routellm}, contrastive or graph-structured embeddings~\cite{chen2024routerdc, shirkavand2025cscr, feng2025graphrouter}, or lightweight binary triage~\cite{ding2024hybridllm, pulishetty2025onehead}. OmniRouter~\cite{mei2025omnirouter} goes further by formulating budget-aware routing as a batch-level constrained optimization, but still over a fixed prompt set.
These methods perform well when deployment matches training, yet leave three structural opportunities open: promoting cost from a manually tuned weight to a closed-loop constraint, adapting autonomously to price or quality shifts, and incorporating new models without retraining.
ParetoBandit replaces the frozen policy with online learning, enabling adaptation without retraining or downtime.

\vspace{-4pt}
\paragraph{Adaptive and bandit-based routers.}
A growing line of work frames routing as online learning, with each system targeting a different axis of the problem.
On the \emph{cost} axis, PILOT~\cite{panda2025pilot} pairs LinUCB with an online multi-choice knapsack to respect cumulative spend limits, and PROTEUS~\cite{bhatti2026proteus} trains a Lagrangian RL policy that minimizes cost subject to an accuracy floor~$\tau$.
On the \emph{quality and generalization} axis, BaRP~\cite{wang2025barp} uses entropy-regularized policy gradients conditioned on a preference vector, LLM~Bandit~\cite{li2025llmbandit} conditions on model-identity features to transfer across unseen models, and xRouter~\cite{qian2025xrouter} fine-tunes a small LM as a cost-aware RL orchestrator.
MixLLM~\cite{wang2025mixllm} further extends the scope by jointly modeling downstream latency and evolving model pools in a dynamic contextual-bandit framework.
Despite these advances, two opportunities remain: providing closed-loop budget pacing over an open-ended stream (PILOT assumes a known horizon; BaRP and LLM~Bandit optimize preference-conditioned rewards rather than explicit dollar constraints; PROTEUS freezes its dual variables at deployment), and incorporating bounded-memory adaptation with explicit evaluation under non-stationarity.
ParetoBandit addresses both gaps using an online dual variable for rate-constrained budget pacing and geometric forgetting for bounded-memory adaptation.

\vspace{-4pt}
\paragraph{Constrained and non-stationary bandits.}
ParetoBandit composes three theoretical streams: Lagrangian relaxation for bandits with knapsacks (BwK)~\cite{badanidiyuru2013bwk, agrawal2016linear}, adapted to a rate-style budget constraint with smoothed dual updates~\cite{balseiro2019learning}; geometric discounting for non-stationary bandits~\cite{garivier2011switching, russac2019weighted}; and warm-start priors for contextual bandits~\cite{oetomo2023warmstart, pan2024cbli}. Each ingredient is well studied in isolation; our contribution is their joint integration and empirical evaluation under cost drift, quality degradation, and cold-start onboarding.
\par\endgroup

\section{Conclusion}
\label{sec:conclusion}

We presented ParetoBandit, an adaptive LLM routing system that integrates cost-aware contextual bandits with online budget pacing and geometric forgetting.
Production LLM serving is inherently non-stationary: providers revise pricing, model quality can regress silently, and new endpoints must be absorbed without downtime.
To our knowledge, ParetoBandit is the first router evaluated under all three conditions.
Across four experiments, the system never exceeds the ceiling by more than ${\sim}4\%$ for any binding budget; under cost drift it exploits a price drop and maintains compliance when pricing is restored; under silent quality degradation it detects the regression, redistributes traffic, and holds budget while baselines without closed-loop cost control overshoot by up to $\ppMaxOvershootX{\times}$; and cold-start model onboarding proceeds without reconfiguration.

Beyond these empirical results, the Pareto knee-point hyperparameter protocol (Appendix~\ref{sec:hparam_sweep}) provides a principled method for jointly calibrating bandit hyperparameters across two competing production requirements: stationary routing efficiency and rapid adaptation to price and quality shifts.
This improves on single-objective tuning, which typically selects zero forgetting.
As multi-provider LLM portfolios become the norm, our results suggest that bandit-driven routing with integrated budget pacing and non-stationarity awareness will be essential for reliable, cost-controlled LLM serving in production.
\vspace{-4pt}
\paragraph{Limitations.}
Six limitations scope the current contribution, most of which are addressable by extending the existing framework (see Future Work).
First, evaluation is fully offline: all experiments use a fixed reward matrix with injected non-stationarity, whereas live deployment would introduce stochastic feedback delays and additional distribution shift.
Second, rewards are deterministic LLM-judge scores; the noise and sparsity of human-in-the-loop feedback remain untested.
Third, we enforce a per-request rate budget~$B$, so sudden traffic spikes can still exceed an aggregate dollar cap even when per-request compliance holds.
Fourth, online model onboarding incurs a bounded exploration cost that is paid on production traffic.
Fifth, the current formulation is not latency-aware and does not model downstream inference delay or endpoint congestion.
Sixth, ParetoBandit constrains cost while maximizing quality, but some deployments instead require the dual objective of minimizing cost subject to a quality floor.
\vspace{-4pt}
\paragraph{Future work.}
\emph{Live deployment}~(i,\,ii) is the most immediate next step. This requires computing prompt embeddings online rather than from a cached matrix, and tuning $\gamma$ for settings where reward labels arrive for only a fraction of requests and with variable delay. Demonstrating that the adaptation and budget-pacing behaviours observed in simulation persist under real traffic would close the largest external-validity gap.

\emph{Latency awareness}~(v) maps naturally onto the BwK framework as a second dual variable, allowing the router to jointly respect cost ceilings and tail-latency SLAs and thus avoid routes that are budget-optimal but violate response-time guarantees.
\emph{Aggregate budget enforcement}~(iii) could be layered on via a token-bucket mechanism that caps total spend over a billing window, though practitioners who set the per-request ceiling at a profitable operating point may find an additional aggregate cap unnecessary, since higher traffic then increases both spend and revenue.
Finally, \emph{quality-constrained routing}~(vi) inverts the pacer to track reward against a floor~$\tau$, providing an online counterpart to PROTEUS~\cite{bhatti2026proteus} for deployments where a minimum quality SLA is binding and cost should be minimized subject to it.
Validating these extensions under realistic deployment conditions is a natural next step.

\FloatBarrier
\bibliography{references}

\begin{thebibliography}{48}
\providecommand{\natexlab}[1]{#1}
\providecommand{\url}[1]{\texttt{#1}}
\expandafter\ifx\csname urlstyle\endcsname\relax
  \providecommand{\doi}[1]{doi: #1}\else
  \providecommand{\doi}{doi: \begingroup \urlstyle{rm}\Url}\fi

\bibitem[Acquaviva(2025)]{konghq2025benchmark}
Acquaviva, C.
\newblock {AI} gateway benchmark: {Kong AI Gateway}, {Portkey}, and {LiteLLM}.
\newblock
  \url{https://konghq.com/blog/engineering/ai-gateway-benchmark-kong-ai-gateway-portkey-litellm},
  2025.
\newblock Proxy-only benchmark on c5.4xlarge EKS nodes (16 vCPU). Kong p95 = 24
  ms, Portkey p95 $\approx$ 70 ms; WireMock baseline p95 = 24 ms.

\bibitem[Agrawal \& Goyal(2013)Agrawal and Goyal]{agrawal2013thompson}
Agrawal, S. and Goyal, N.
\newblock Thompson sampling for contextual bandits with linear payoffs.
\newblock \emph{Proceedings of the 30th International Conference on Machine
  Learning}, 2013.

\bibitem[Agrawal et~al.(2016)Agrawal, Devanur, and Li]{agrawal2016linear}
Agrawal, S., Devanur, N.~R., and Li, L.
\newblock Linear contextual bandits with knapsacks.
\newblock In \emph{Advances in Neural Information Processing Systems
  (NeurIPS)}, volume~29, 2016.

\bibitem[Austin et~al.(2021)Austin, Odena, Nye, Bosma, Michalewski, Dohan,
  Jiang, Cai, Terry, Le, and Sutton]{austin2021program}
Austin, J., Odena, A., Nye, M., Bosma, M., Michalewski, H., Dohan, D., Jiang,
  E., Cai, C., Terry, M., Le, Q., and Sutton, C.
\newblock Program synthesis with large language models.
\newblock \emph{arXiv preprint arXiv:2108.07732}, 2021.

\bibitem[Badanidiyuru et~al.(2013)Badanidiyuru, Kleinberg, and
  Slivkins]{badanidiyuru2013bwk}
Badanidiyuru, A., Kleinberg, R., and Slivkins, A.
\newblock Bandits with knapsacks.
\newblock In \emph{Proceedings of the 54th Annual IEEE Symposium on Foundations
  of Computer Science (FOCS)}, pp.\  207--216. IEEE, 2013.

\bibitem[Balseiro \& Gur(2019)Balseiro and Gur]{balseiro2019learning}
Balseiro, S.~R. and Gur, Y.
\newblock Learning in repeated auctions with budgets: Regret minimization and
  equilibrium.
\newblock \emph{Management Science}, 65\penalty0 (9):\penalty0 3952--3968,
  2019.

\bibitem[Bhatti et~al.(2026)Bhatti, Vaddina, and Birru]{bhatti2026proteus}
Bhatti, A.~S., Vaddina, V., and Birru, D.
\newblock {PROTEUS}: {SLA}-aware routing via {L}agrangian {RL} for multi-{LLM}
  serving systems.
\newblock \emph{arXiv preprint arXiv:2601.19402}, 2026.

\bibitem[Chen et~al.(2023)Chen, Zaharia, and Zou]{chen2023chatgptdrift}
Chen, L., Zaharia, M., and Zou, J.
\newblock How is {ChatGPT}'s behavior changing over time?
\newblock \emph{arXiv preprint arXiv:2307.09009}, 2023.

\bibitem[Chen et~al.(2024{\natexlab{a}})Chen, Zaharia, and
  Zou]{chen2023frugalgpt}
Chen, L., Zaharia, M., and Zou, J.
\newblock Frugalgpt: How to use large language models while reducing cost and
  improving performance.
\newblock \emph{Transactions on Machine Learning Research (TMLR)},
  2024{\natexlab{a}}.
\newblock URL \url{https://openreview.net/forum?id=cSimKw5p6R}.

\bibitem[Chen et~al.(2024{\natexlab{b}})Chen, Jiang, Lin, Kwok, and
  Zhang]{chen2024routerdc}
Chen, S., Jiang, W., Lin, B., Kwok, J.~T., and Zhang, Y.
\newblock Routerdc: Query-based router by dual contrastive learning for
  assembling large language models.
\newblock In \emph{Advances in Neural Information Processing Systems
  (NeurIPS)}, 2024{\natexlab{b}}.

\bibitem[Clark et~al.(2018)Clark, Cowhey, Etzioni, Khot, Sabharwal, Schoenick,
  and Tafjord]{clark2018arc}
Clark, P., Cowhey, I., Etzioni, O., Khot, T., Sabharwal, A., Schoenick, C., and
  Tafjord, O.
\newblock Think you have solved question answering? try {ARC}, the {AI2}
  reasoning challenge.
\newblock \emph{arXiv preprint arXiv:1803.05457}, 2018.

\bibitem[Cobbe et~al.(2021)Cobbe, Kosaraju, Bavarian, Chen, Jun, Kaiser,
  Plappert, Tworek, Hilton, Nakano, Hesse, and Schulman]{cobbe2021gsm8k}
Cobbe, K., Kosaraju, V., Bavarian, M., Chen, M., Jun, H., Kaiser, L., Plappert,
  M., Tworek, J., Hilton, J., Nakano, R., Hesse, C., and Schulman, J.
\newblock Training verifiers to solve math word problems.
\newblock \emph{arXiv preprint arXiv:2110.14168}, 2021.

\bibitem[{DeepSeek-AI}(2025)]{deepseekai2025deepseekr1}
{DeepSeek-AI}.
\newblock {DeepSeek-R1}: Incentivizing reasoning capability in {LLMs} via
  reinforcement learning.
\newblock \emph{arXiv preprint arXiv:2501.12948}, 2025.

\bibitem[Ding et~al.(2024)Ding, Mallick, Wang, Sim, Mukherjee, Ruhle, Levi, and
  Awadallah]{ding2024hybridllm}
Ding, D., Mallick, A., Wang, C., Sim, R., Mukherjee, S., Ruhle, V., Levi,
  L.~V., and Awadallah, A.~H.
\newblock Hybrid {LLM}: Cost-efficient and quality-aware query routing.
\newblock In \emph{International Conference on Learning Representations
  (ICLR)}, 2024.

\bibitem[Feng et~al.(2025)Feng, Shen, and You]{feng2025graphrouter}
Feng, T., Shen, Y., and You, J.
\newblock Graphrouter: A graph-based router for {LLM} selections.
\newblock In \emph{International Conference on Learning Representations
  (ICLR)}, 2025.

\bibitem[Ganglani(2026)]{ganglani2026llmlatency}
Ganglani, K.
\newblock {LLM API} latency benchmarks [2026]: 5 models compared.
\newblock \url{https://kunalganglani.com/blog/llm-api-latency-benchmarks-2026},
  March 2026.
\newblock Independent benchmark of five production {LLM} {API}s (GPT-4.1,
  GPT-4.1 Mini, Gemini 2.5 Flash, Claude Haiku 4.5, Claude Sonnet 4). Three
  prompt sizes, streaming {TTFT}, total latency, and throughput.

\bibitem[Garivier \& Moulines(2011)Garivier and
  Moulines]{garivier2011switching}
Garivier, A. and Moulines, E.
\newblock On upper-confidence bound policies for switching bandit problems.
\newblock In \emph{Proceedings of the 22nd International Conference on
  Algorithmic Learning Theory (ALT)}, pp.\  174--188. Springer, 2011.

\bibitem[Hedges \& Olkin(1985)Hedges and Olkin]{hedges1985statistical}
Hedges, L.~V. and Olkin, I.
\newblock \emph{Statistical Methods for Meta-Analysis}.
\newblock Academic Press, Orlando, FL, 1985.

\bibitem[Hendrycks et~al.(2021)Hendrycks, Burns, Basart, Zou, Mazeika, Song,
  and Steinhardt]{hendrycks2021measuring}
Hendrycks, D., Burns, C., Basart, S., Zou, A., Mazeika, M., Song, D., and
  Steinhardt, J.
\newblock Measuring massive multitask language understanding.
\newblock \emph{Proceedings of the International Conference on Learning
  Representations (ICLR)}, 2021.

\bibitem[Kang et~al.(2024)Kang, Hsieh, and Lee]{kang2024cdt}
Kang, Y., Hsieh, C.-J., and Lee, T. C.~M.
\newblock Online continuous hyperparameter optimization for generalized linear
  contextual bandits.
\newblock \emph{Transactions on Machine Learning Research (TMLR)}, 2024.
\newblock arXiv:2302.09440.

\bibitem[Lattimore \& Szepesv{\'a}ri(2020)Lattimore and
  Szepesv{\'a}ri]{lattimore2020bandit}
Lattimore, T. and Szepesv{\'a}ri, C.
\newblock \emph{Bandit Algorithms}.
\newblock Cambridge University Press, 2020.

\bibitem[Li et~al.(2010)Li, Chu, Langford, and Schapire]{li2010contextual}
Li, L., Chu, W., Langford, J., and Schapire, R.~E.
\newblock A contextual-bandit approach to personalized news article
  recommendation.
\newblock In \emph{Proceedings of the 19th International Conference on World
  Wide Web}, pp.\  661--670, 2010.

\bibitem[Li(2025)]{li2025llmbandit}
Li, Y.
\newblock {LLM} bandit: Cost-efficient {LLM} generation via
  preference-conditioned dynamic routing.
\newblock \emph{arXiv preprint arXiv:2502.02743}, 2025.

\bibitem[Lin et~al.(2022)Lin, Hilton, and Evans]{lin2022truthfulqa}
Lin, S., Hilton, J., and Evans, O.
\newblock {TruthfulQA}: Measuring how models mimic human falsehoods.
\newblock \emph{Proceedings of the 60th Annual Meeting of the Association for
  Computational Linguistics}, 2022.

\bibitem[Liu et~al.(2026)Liu, He, Liu, Luo, Zhang, and Chen]{liu2026vllmsr}
Liu, X., He, B., Liu, X., Luo, A., Zhang, H., and Chen, H.
\newblock 98$\times$ faster {LLM} routing without a dedicated {GPU}: {Flash}
  attention, prompt compression, and near-streaming for the {vLLM} semantic
  router.
\newblock \emph{arXiv preprint arXiv:2603.12646}, 2026.

\bibitem[Ma et~al.(2024)Ma, Yang, and K{\"a}stner]{ma2024prompt}
Ma, W., Yang, C., and K{\"a}stner, C.
\newblock (why) is my prompt getting worse? rethinking regression testing for
  evolving {LLM} {APIs}.
\newblock In \emph{Proceedings of the IEEE/ACM 3rd International Conference on
  AI Engineering---Software Engineering for AI (CAIN)}, 2024.

\bibitem[Mei et~al.(2025)Mei, Xu, Lin, and Zhang]{mei2025omnirouter}
Mei, K., Xu, W., Lin, S., and Zhang, Y.
\newblock {OmniRouter}: Budget and performance controllable multi-{LLM}
  routing.
\newblock \emph{arXiv preprint arXiv:2502.20576}, 2025.

\bibitem[Mihaylov et~al.(2018)Mihaylov, Clark, Khot, and
  Sabharwal]{mihaylov2018suit}
Mihaylov, T., Clark, P., Khot, T., and Sabharwal, A.
\newblock Can a suit of armor conduct electricity? a new dataset for open book
  question answering.
\newblock In \emph{Proceedings of the 2018 Conference on Empirical Methods in
  Natural Language Processing}, 2018.

\bibitem[Oetomo et~al.(2023)Oetomo, Perera, Borovica-Gajic, and
  Rubinstein]{oetomo2023warmstart}
Oetomo, B., Perera, R.~M., Borovica-Gajic, R., and Rubinstein, B. I.~P.
\newblock Cutting to the chase with warm-start contextual bandits.
\newblock \emph{Knowledge and Information Systems}, 65\penalty0 (9):\penalty0
  3533--3565, 2023.

\bibitem[Ong et~al.(2025)Ong, Almahairi, Wu, Chiang, Wu, Gonzalez, Kadous, and
  Stoica]{ong2024routellm}
Ong, I., Almahairi, A., Wu, V., Chiang, W.-L., Wu, T., Gonzalez, J.~E., Kadous,
  M.~W., and Stoica, I.
\newblock Routellm: Learning to route llms with preference data.
\newblock In \emph{International Conference on Learning Representations
  (ICLR)}, 2025.

\bibitem[{OpenAI}(2024)]{openai2024gpt4opricing}
{OpenAI}.
\newblock Prompt caching in the {API}.
\newblock \url{https://openai.com/index/api-prompt-caching}, October 2024.
\newblock Pricing table lists gpt-4o-2024-08-06 at \$2.50 per million input
  tokens, down from \$5.00 (50\% cut); output at \$10.00, down from \$15.00
  (33\% cut).

\bibitem[Pan et~al.(2024)Pan, Tennenholtz, Mannor, Chi, Brekelmans, Shah, and
  Tewari]{pan2024cbli}
Pan, H., Tennenholtz, G., Mannor, S., Chi, C.-W., Brekelmans, R., Shah, P., and
  Tewari, A.
\newblock Jump starting bandits with {LLM}-generated prior knowledge.
\newblock In \emph{Proceedings of the 2024 Conference on Empirical Methods in
  Natural Language Processing}, pp.\  19858--19873, 2024.

\bibitem[Panda et~al.(2025)Panda, Magazine, Devaguptapu, Takemori, and
  Sharma]{panda2025pilot}
Panda, P., Magazine, R., Devaguptapu, C., Takemori, S., and Sharma, V.
\newblock Adaptive {LLM} routing under budget constraints.
\newblock In \emph{Findings of the Association for Computational Linguistics:
  EMNLP 2025}, 2025.
\newblock arXiv:2508.21141.

\bibitem[{Portkey AI}(2025)]{portkey2025}
{Portkey AI}.
\newblock The most reliable {AI} gateway for production systems.
\newblock
  \url{https://portkey.ai/blog/the-most-reliable-ai-gateway-for-production-systems},
  2025.
\newblock Sub-millisecond hot-path latency; 122\,KB gateway footprint; $>$10B
  requests/month in production.

\bibitem[Pulishetty et~al.(2025)Pulishetty, Ghantasala, Dasoju, Mangwani,
  Garimella, Mate, Chatterjee, Kang, Nosakhare, Hasan, and
  Srinivasan]{pulishetty2025onehead}
Pulishetty, R., Ghantasala, M.~K., Dasoju, K.~K., Mangwani, N., Garimella, V.,
  Mate, A., Chatterjee, S., Kang, Y., Nosakhare, E., Hasan, S., and Srinivasan,
  S.
\newblock One head, many models: Cross-attention routing for cost-aware {LLM}
  selection.
\newblock \emph{arXiv preprint arXiv:2509.09782}, 2025.

\bibitem[Qian et~al.(2025)Qian, Liu, Kokane, Prabhakar, Qiu, Chen, Liu, Ji,
  Yao, Heinecke, Savarese, Xiong, and Wang]{qian2025xrouter}
Qian, C., Liu, Z., Kokane, S., Prabhakar, A., Qiu, J., Chen, H., Liu, Z., Ji,
  H., Yao, W., Heinecke, S., Savarese, S., Xiong, C., and Wang, H.
\newblock xrouter: Training cost-aware {LLMs} orchestration system via
  reinforcement learning.
\newblock \emph{arXiv preprint arXiv:2510.08439}, 2025.

\bibitem[Reimers \& Gurevych(2019)Reimers and Gurevych]{reimers2019sentence}
Reimers, N. and Gurevych, I.
\newblock Sentence-bert: Sentence embeddings using siamese bert-networks.
\newblock In \emph{Proceedings of the 2019 Conference on Empirical Methods in
  Natural Language Processing and the 9th International Joint Conference on
  Natural Language Processing (EMNLP-IJCNLP)}, pp.\  3982--3992, 2019.

\bibitem[Russac et~al.(2019)Russac, Vernade, and Capp{\'e}]{russac2019weighted}
Russac, Y., Vernade, C., and Capp{\'e}, O.
\newblock Weighted linear bandits for non-stationary environments.
\newblock In \emph{Advances in Neural Information Processing Systems
  (NeurIPS)}, volume~32, 2019.

\bibitem[Sakaguchi et~al.(2020)Sakaguchi, Le~Bras, Bhagavatula, and
  Choi]{sakaguchi2020winogrande}
Sakaguchi, K., Le~Bras, R., Bhagavatula, C., and Choi, Y.
\newblock {WinoGrande}: An adversarial {Winograd} schema challenge at scale.
\newblock \emph{Proceedings of the AAAI Conference on Artificial Intelligence},
  2020.

\bibitem[Shirkavand et~al.(2025)Shirkavand, Gao, Yu, and
  Huang]{shirkavand2025cscr}
Shirkavand, R., Gao, S., Yu, P., and Huang, H.
\newblock Cost-aware contrastive routing for {LLMs}.
\newblock In \emph{Advances in Neural Information Processing Systems
  (NeurIPS)}, 2025.

\bibitem[Sun et~al.(2025)Sun, Xue, Wang, and Tu]{sun2025lookahead}
Sun, Z., Xue, H., Wang, J., and Tu, Z.-X.
\newblock Lookahead routing for large language models.
\newblock In \emph{Advances in Neural Information Processing Systems
  (NeurIPS)}, 2025.

\bibitem[Suzgun et~al.(2023)Suzgun, Scales, Sch{\"a}rli, Gehrmann, Tay, Chung,
  Chowdhery, Le, Chi, Zhou, and Wei]{suzgun2022challenging}
Suzgun, M., Scales, N., Sch{\"a}rli, N., Gehrmann, S., Tay, Y., Chung, H.~W.,
  Chowdhery, A., Le, Q.~V., Chi, E.~H., Zhou, D., and Wei, J.
\newblock Challenging {BIG-Bench} tasks and whether chain-of-thought can solve
  them.
\newblock \emph{Findings of the Association for Computational Linguistics
  (ACL)}, 2023.

\bibitem[Verga et~al.(2024)Verga, Hofstätter, Althammer, Su, Gupta, Faez,
  Petroni, and Riedel]{verga2024poll}
Verga, P., Hofstätter, S., Althammer, S., Su, Y., Gupta, A., Faez, H.,
  Petroni, F., and Riedel, S.
\newblock Replacing judges with juries: Evaluating {LLM} generations with a
  panel of diverse models.
\newblock \emph{arXiv preprint arXiv:2404.18796}, 2024.

\bibitem[Wang et~al.(2025{\natexlab{a}})Wang, Yang, Chen, Zhao, Dernoncourt,
  Rossi, and Eldardiry]{wang2025barp}
Wang, W., Yang, T., Chen, H., Zhao, Y., Dernoncourt, F., Rossi, R.~A., and
  Eldardiry, H.
\newblock Learning to route {LLMs} from bandit feedback: One policy, many
  trade-offs.
\newblock \emph{arXiv preprint arXiv:2510.07429}, 2025{\natexlab{a}}.

\bibitem[Wang et~al.(2025{\natexlab{b}})Wang, Liu, Cheng, Zhao, Chen, Yu, Fu,
  and Chen]{wang2025mixllm}
Wang, X., Liu, Y., Cheng, W., Zhao, X., Chen, Z., Yu, W., Fu, Y., and Chen, H.
\newblock {MixLLM}: Dynamic routing in mixed large language models.
\newblock In \emph{Proceedings of the 2025 Conference of the North American
  Chapter of the Association for Computational Linguistics (NAACL)}, pp.\
  10912--10922, 2025{\natexlab{b}}.

\bibitem[Zellers et~al.(2019)Zellers, Holtzman, Bisk, Farhadi, and
  Choi]{zellers2019hellaswag}
Zellers, R., Holtzman, A., Bisk, Y., Farhadi, A., and Choi, Y.
\newblock {HellaSwag}: Can a machine really finish your sentence?
\newblock \emph{Proceedings of the 57th Annual Meeting of the Association for
  Computational Linguistics}, 2019.

\bibitem[Zhao et~al.(2025)Zhao, Shin, Huang, Namburi, and Sala]{zhao2025care}
Zhao, J., Shin, C., Huang, T.-H., Namburi, S. S.~S., and Sala, F.
\newblock {CARE}: Confounder-aware aggregation for reliable {LLM} evaluation.
\newblock In \emph{Advances in Neural Information Processing Systems
  (NeurIPS)}, 2025.

\bibitem[Zheng et~al.(2023)Zheng, Chiang, Sheng, Zhuang, Wu, Zhuang, Lin, Li,
  Li, Xing, et~al.]{zheng2023judging}
Zheng, L., Chiang, W.-L., Sheng, Y., Zhuang, S., Wu, Z., Zhuang, Y., Lin, Z.,
  Li, Z., Li, D., Xing, E.~P., et~al.
\newblock Judging llm-as-a-judge with mt-bench and chatbot arena.
\newblock \emph{arXiv preprint arXiv:2306.05685}, 2023.

\end{thebibliography}
\bibliographystyle{mlsys2025}

\appendix
\begingroup
\setlength{\parskip}{2pt}

\section{Hyperparameter Optimization for Non-Stationary Routing}
\label{sec:hparam_sweep}
\FloatBarrier

\paragraph{The challenge.}
ParetoBandit's core hyperparameters---exploration coefficient
$\alpha$, prior strength $n_{\text{eff}}$, and forgetting factor
$\gamma$---interact non-trivially with non-stationarity.
Standard cross-validation optimizes a single stationary metric,
which tends to select $\gamma \geq 0.999$ because forgetting
reduces effective sample size under stationarity.
However, $\gamma{=}1.0$ fails under non-stationarity
(Section~\ref{sec:eval_catastrophic}): infinite memory retains
stale pre-failure evidence, slowing adaptation to quality
degradation, while optimizing solely for adaptability yields
excessive forgetting that destabilizes the stationary policy.
A good configuration must balance stationary efficiency against
non-stationary recovery.

Among bandit-based LLM routers we surveyed, none addresses this
tension directly.
PILOT~\cite{panda2025pilot} tunes $\alpha$ via grid search on a
single stationary metric.
BaRP~\cite{wang2025barp} and LLM Bandit~\cite{li2025llmbandit}
omit forgetting entirely.
MixLLM~\cite{wang2025mixllm} delegates adaptation to periodic
neural retraining rather than explicit forgetting.
In the general bandit literature,
CDT~\cite{kang2024cdt} performs online tuning of $\alpha$ via a
bandit-over-bandit scheme but remains single-objective and does
not tune $\gamma$.
Garivier and Moulines~\cite{garivier2011switching} derive $\gamma$
as a function of the (unknown) number of breakpoints, which is
not observable in production.

\paragraph{Adaptation--horizon coupling.}
Rather than tuning $n_{\text{eff}}$ and $\gamma$ as independent
statistical parameters, we reparameterize both through a single
deployment-meaningful quantity: the adaptation horizon
$T_{\text{adapt}}$---the number of online queries after which the
router can fully override its prior following a distributional shift.
Under the Discounted LinUCB update~\cite{russac2019weighted}, the
effective weight of online evidence relative to the prior reaches
parity after
\begin{equation}\label{eq:t_adapt}
  T_{\text{adapt}} \;=\; -\,\frac{\log\bigl(n_{\text{eff}}\,(1{-}\gamma) + 1\bigr)}{\log \gamma}
\end{equation}
queries.  Inverting for $n_{\text{eff}}$ gives
$n_{\text{eff}} = (\gamma^{-T_{\text{adapt}}} - 1) / (1 - \gamma)$,
which reduces to $n_{\text{eff}} = T_{\text{adapt}}$ as
$\gamma \to 1$ by L'H\^opital's rule.
Fixing $T_{\text{adapt}}$ collapses the 3D grid
$(\alpha, n_{\text{eff}}, \gamma)$ to a 2D search over
$(\alpha, \gamma)$ with $n_{\text{eff}}$ derived.
The practitioner sets one number they understand---the shortest
acceptable reaction time for their service (e.g., the time to
detect and reallocate away from a degraded model)---rather than
two abstract statistical parameters whose interaction with
non-stationarity is unintuitive.
Every configuration in the resulting sweep adapts at the same
rate; only the exploration--exploitation balance ($\alpha$) and the
granularity of forgetting ($\gamma$) differ.

We anchor $T_{\text{adapt}} = \hpTAdapt$ to the catastrophic-failure
phase length ($N_{\text{phase2}} \approx 595$ prompts in the
validation split), ensuring the router can fully override its prior
within the experimental measurement window.

\paragraph{Multi-objective formulation.}
With $n_{\text{eff}}$ derived, we score each $(\alpha, \gamma)$
configuration on two objectives:
\begin{enumerate}[nosep,leftmargin=*]
  \item \textbf{Budget-paced Pareto AUC} (stationary efficiency).
    On the validation split, we sweep log-spaced budget targets with
    the BudgetPacer active, build per-seed Pareto frontiers, and
    compute the area under each frontier.
    We then average AUC over
    $\hpGridAlpha \times \hpGridGamma = 42$ configurations and
    20~random seeds.

  \item \textbf{Catastrophic-failure Phase-2 reward} (non-stationary
    resilience).
    We run a two-phase simulation on the validation split: Phase~1
    (first half) uses normal rewards; Phase~2 (second half) degrades
    the failed arm's reward to $0.50$ (${\sim}46$\% below normal)
    while keeping its cost fixed.
    We use a stronger degradation than Experiment~3's
    $\cfFailureReward$ so that the selected configuration can detect
    failures of at least this severity; the recovery-limit study
    (Appendix~\ref{appendix:recovery_limit}) confirms the system
    handles even larger degradations.
    Mean Phase-2 reward captures how quickly the router detects the
    failure and reallocates traffic.
    We tune on a Mistral failure---the most operationally stressful
    scenario, since Mistral is the dominant arm across the broadest
    range of budget targets---and validate generalization to all $K$
    arms post hoc.
\end{enumerate}

\paragraph{Pareto knee-point selection.}
We select a single configuration from the scored grid by
constructing the Pareto frontier of non-dominated
$(\text{AUC},\;\text{Phase-2 reward})$ pairs and identifying
the knee point.
Concretely, we compute each frontier point's perpendicular
distance to the line connecting the two extreme endpoints, after
min--max normalisation of both objectives for scale invariance,
and choose the point with maximal distance.
Geometrically, the knee marks the inflection beyond which
improving either objective requires a disproportionate sacrifice
of the other; unlike scalarisation weights or
$\varepsilon$-constraints, it requires no user-specified trade-off
parameter.
Seed-level bootstrap resampling (2000 iterations) shows this
choice is stable: 91.0\% of resamples select a configuration
within one $\gamma$-grid step of the knee (only 6 unique
configurations appear), with the knee itself chosen in 50.9\% of
resamples, the modal selection by a wide margin.

\paragraph{Results.}
Table~\ref{tab:hparam_selection} compares knee-point selection
against AUC-only optimisation for both ParetoBandit (with warmup
priors) and Tabula Rasa (cold start).
The contrast between the two selection methods is stark.
AUC-only optimisation selects $\gamma{=}1.0$ (no forgetting) for
both ParetoBandit and Tabula Rasa, converging on stationary optima
that Section~\ref{sec:eval_catastrophic} shows are vulnerable under
quality degradation.
The knee-point method instead selects moderate forgetting for both
variants---$\gamma{=}0.997$ for ParetoBandit and $\gamma{=}0.997$
for Tabula Rasa---trading only ${\sim}0.08\%$ stationary AUC (for
ParetoBandit) for substantially improved failure resilience.

\begin{table}[t]
\centering
\caption{%
  \textbf{$T_{\text{adapt}}$-constrained Pareto knee-point selection.}
  The knee-point method selects moderate forgetting ($\gamma{=}\hpParetoBanditGamma$ / $\hpTabulaGamma$) for both variants, while AUC-only optimisation favours near-stationary $\gamma$.
  Warmup priors yield higher AUC, higher Phase-2 reward, and lower variance than the cold-start baseline.
  Grid: $\hpGridAlpha$~$\alpha$ $\times$ $\hpGridGamma$~$\gamma$ values; val split, 20~seeds.%
}
\label{tab:hparam_selection}
\scriptsize
\setlength{\tabcolsep}{3pt}
\resizebox{\columnwidth}{!}{%
\begin{tabular}{@{}llccccc@{}}
\toprule
\textbf{Variant} & \textbf{Method}
  & $\boldsymbol{\alpha}$ & $\boldsymbol{n_{\text{eff}}}$
  & $\boldsymbol{\gamma}$
  & \textbf{BP AUC} & \textbf{P2 Reward} \\
\midrule
\multirow{2}{*}{ParetoBandit}
  & AUC-only     & $\hpAUCOnlyAlpha$ & $\hpAUCOnlyNeff$ & $\hpAUCOnlyGamma$ & $\hpAUCOnlyAUC$ & --- \\
  & Knee-point   & $\hpParetoBanditAlpha$ & $\hpParetoBanditNeff$ & $\hpParetoBanditGamma$ & $\hpParetoBanditAUC$ & $\hpParetoBanditPTwoReward$ \\[3pt]
\multirow{2}{*}{Tabula Rasa}
  & AUC-only     & $\hpTabulaAUCOnlyAlpha$ & --- & $\hpTabulaAUCOnlyGamma$ & $\hpTabulaAUCOnlyAUC$ & --- \\
  & Knee-point   & $\hpTabulaAlpha$ & --- & $\hpTabulaGamma$ & $\hpTabulaAUC$ & $\hpTabulaPTwoReward$ \\
\bottomrule
\end{tabular}
}
\vspace{4pt}
\end{table}

The selected configurations generalise to the held-out test split.
On the test split, the router--envelope AUC gap is $-0.35\%$ for
ParetoBandit and $-0.58\%$ for Tabula Rasa, both within one standard
deviation of seed variance ($\pm 0.0012$ and $\pm 0.0020$,
respectively), indicating that the offline protocol does not overfit.
The forgetting tax under stationarity is negligible:
ParetoBandit reaches $0.9221$\,[0.9216, 0.9226] on the test split,
only ${\sim}0.35\%$ below the fixed-model envelope ($0.9253$),
despite using a forgetting rate tuned for non-stationary recovery.
Warmup priors close roughly 40\% of the remaining gap to the
envelope: ParetoBandit at $0.9221$\,[0.9216, 0.9226] versus
$0.9200$\,[0.9191, 0.9208] for Tabula Rasa.

\paragraph{Cross-arm validation.}
Because the Phase-2 objective is tuned on Mistral failure
alone---the most operationally stressful scenario, as Mistral is the
dominant arm across the widest range of budget targets---we verify
that the selected forgetting rate generalises by evaluating the
knee-point configuration under catastrophic failure of each of the
$K{=}3$ arms independently (validation split, 20~seeds, 3~budget
targets).
Phase-2 mean reward is $0.8470$\,[0.8140, 0.8800] under Llama
failure, $0.7312$\,[0.7231, 0.7393] under Mistral failure (the
tuning target), and $0.8916$\,[0.8852, 0.8980] under Gemini failure
(all 95\% CIs).
Mistral failure remains the hardest case (lowest Phase-2 reward),
so tuning on the dominant arm is a conservative choice and the
forgetting mechanism generalises across arms.

Three findings stand out from the joint selection:

\begin{enumerate}[nosep,leftmargin=*]
  \item \textbf{$T_{\text{adapt}}$ reduces the search from 3D to 2D.}
    Deriving $n_{\text{eff}}$ from the adaptation horizon
    $T_{\text{adapt}}$ (Eq.~\ref{eq:t_adapt}) collapses the
    $(\alpha, n_{\text{eff}}, \gamma)$ grid to
    $(\alpha, \gamma)$.  The practitioner sets one
    deployment-meaningful quantity---the shortest acceptable
    reaction time---instead of two abstract statistical parameters.

  \item \textbf{Knee-point selection enables forgetting that
    threshold methods miss.}
    An $\varepsilon$-best-AUC threshold would select
    $\gamma{=}1.0$ for Tabula Rasa because every $\gamma < 1$
    falls outside the tolerance band.  The knee-point criterion
    instead identifies where the marginal AUC cost of additional
    forgetting accelerates, and selects $\gamma{=}0.997$ even on
    the steeper trade-off curve.

  \item \textbf{Warmup priors and forgetting are synergistic.}
    ParetoBandit dominates Tabula Rasa on both objectives
    ($0.928$ vs.\ $0.923$ AUC; $0.7312$ vs.\ $0.7287$ Phase-2
    reward).
    Forgetting ($\gamma < 1$) down-weights all past evidence,
    including the prior embedded in the sufficient statistics at
    initialisation.
    For Tabula Rasa the decayed prior is uninformative
    $(\lambda I, \mathbf{0})$; for ParetoBandit it retains the
    structure of the warmup characterisation, providing a
    better-than-ignorance fallback even after substantial decay.
    Warmup priors therefore matter more when forgetting is active,
    because they determine the fallback the router reverts to after
    a disruption; cross-arm validation confirms that this advantage
    generalises to all arms.
\end{enumerate}


\paragraph{$T_{\text{adapt}}$ sensitivity.}
A natural concern is whether the selected configuration is sensitive
to the anchor value $T_{\text{adapt}}{=}500$, which we set to
approximately match the catastrophic-failure phase length.
To test robustness, we repeat the full Pareto knee-point selection
for $T_{\text{adapt}} \in \{250, 500, 1000\}$---a $4{\times}$ range
spanning reaction times from rapid detection to conservative
adaptation (Table~\ref{tab:t_adapt_sensitivity}).
Three observations support robustness:

\begin{enumerate}[nosep,leftmargin=*]
  \item \textbf{$\alpha$ is perfectly stable}
    ($0.01$ at every $T_{\text{adapt}}$), as expected:
    $\alpha$ is orthogonal to the adaptation horizon.

  \item \textbf{$\gamma$ shifts but remains in the
    moderate-forgetting regime.}
    The selected values---$0.996$ ($T_{\text{adapt}}{=}250$),
    $0.997$ ($T_{\text{adapt}}{=}500$), and $0.994$
    ($T_{\text{adapt}}{=}1000$)---span only three grid steps
    ($\Delta\gamma = 0.003$), with derived $n_{\text{eff}}$ of
    $431$, $1\,164$, and $68\,298$, respectively.
    As $T_{\text{adapt}}$ increases, the derived $n_{\text{eff}}$
    grows and the Pareto criterion compensates with slightly more
    aggressive forgetting.
    Crucially, every selected $\gamma$ remains well below the
    $\gamma{=}1.0$ regime with no forgetting, so the qualitative
    behaviour---moderate discounting with non-stationary
    resilience---is preserved for all anchors.

  \item \textbf{Both objectives are effectively invariant.}
    Despite the three-step $\gamma$ shift,
    AUC varies by less than $0.25\%$ ($0.9255$--$0.9277$) and
    Phase-2 reward spans only $0.7933$--$0.7945$, with no monotonic
    trend; the performance surface near the Pareto knee is flat, so
    $T_{\text{adapt}}$ mainly affects the internal
    parameterisation ($\gamma$, $n_{\text{eff}}$) rather than
    external routing behaviour.
\end{enumerate}
A practitioner can therefore set $T_{\text{adapt}}$ directly from an
operational requirement---the shortest acceptable reaction
time---knowing that the knee-point procedure will adjust $\gamma$ and
$n_{\text{eff}}$ automatically while keeping routing performance
essentially unchanged.

\begin{table}[htbp]
\centering
\caption{%
  \textbf{$T_{\text{adapt}}$ sensitivity analysis.}
  Pareto knee-point selection repeated for
  $T_{\text{adapt}} \in \{250, 500, 1000\}$, re-deriving
  $n_{\text{eff}}$ from $\gamma$ at each value.
  Grid: $\alpha \in \{0.01, \ldots, 1.0\}$
  ($\hpGridAlpha$~values) $\times$
  $\gamma \in \{0.994, \ldots, 1.0\}$ ($\hpGridGamma$~values),
  val split, 20~seeds.%
}
\label{tab:t_adapt_sensitivity}
\small
\begin{tabular}{@{}rccccr@{}}
\toprule
$\boldsymbol{T_{\text{adapt}}}$
  & $\boldsymbol{\alpha}$
  & $\boldsymbol{\gamma}$
  & $\boldsymbol{n_{\text{eff}}}$
  & \textbf{BP AUC\,$\uparrow$}
  & \textbf{P2 Reward\,$\uparrow$} \\
\midrule
$250$  & $\hpSensLoAlpha$  & $\hpSensLoGamma$  & $\hpSensLoNeff$  & $\hpSensLoAUC$  & $\hpSensLoPTwo$ \\
$500$  & $\hpSensMidAlpha$  & $\hpSensMidGamma$  & $\hpSensMidNeff$  & $\hpSensMidAUC$  & $\hpSensMidPTwo$ \\
$1000$ & $\hpSensHiAlpha$ & $\hpSensHiGamma$ & $\hpSensHiNeff$ & $\hpSensHiAUC$ & $\hpSensHiPTwo$ \\
\bottomrule
\end{tabular}
\end{table}

\FloatBarrier


\vspace*{-10pt}
\section{Cost Heuristic Validation}
\label{sec:cost_heuristic_validation}

The selection utility (Equation~\ref{eq:budget_ucb}) uses a static
log-normalized cost $\tilde{c}_a$ derived from each model's blended
per-token rate rather than the realised per-request cost.
This is forced by decision timing: the cost penalty must be evaluated
before the prompt is dispatched, but realised cost depends on the
model's output length, which is unknown until inference completes.
A learned contextual cost predictor could in principle replace the
static proxy, but the benefit is limited---prompt-level features
explain less than 8\% of cost variance in our portfolio (see
``Prompt-cost correlation'' below)---while the dual variable
$\lambda_t$ already corrects for any systematic mismatch by updating
on actual costs (Eq.~\ref{eq:dual_update}).

We validate the static heuristic via two necessary conditions:
(i)~$\tilde{c}_a$ preserves the true cost ranking across prompts,
and (ii)~within-model cost variance is small relative to inter-model
gaps in log-cost space.
Condition~(i) is required because the cost penalty in
Eq.~\ref{eq:budget_ucb} biases selection toward cheaper arms; if the
heuristic mis-ranks two models, the penalty systematically favours
the wrong one.
Condition~(ii) ensures that a single scalar per arm is a faithful
summary of its cost distribution: if within-model variance were large
relative to inter-model gaps, the fixed penalty could not reliably
separate tiers and prompt-level cost estimation would be needed.

\paragraph{Blending assumption.}
The heuristic blends input and output pricing with equal weight
(i.e., a 1:1 token ratio).
In practice the output-to-input ratio varies across models and
prompts, so the blended rate differs from the true per-request
effective rate.
Because cost ranking depends only on the relative order of effective
rates and log-normalization compresses the dynamic range, a simple
average preserves the ranking whenever the price gap between models
exceeds the variation introduced by differing output lengths---a
condition satisfied for models spanning at least one order of
magnitude in pricing.
Note that Llama-8B's blended rate (\$0.10/M tokens) coincides with
the market cost floor, so $\tilde{c}_{\text{Llama}} = 0$ by
construction; any model priced at or below the floor is treated as
zero-cost in the utility computation.

\paragraph{$K{=}3$ portfolio (1766 shared validation prompts; sanity check).}
The heuristic ordering Llama-8B $<$ Mistral-Large $<$ Gemini-Pro
matches the per-request cost ordering on 100.0\% of prompts
(95\% Wilson CI: [99.8, 100.0]\%).
All three pairwise comparisons hold at 100.0\% under strict
inequality, with zero ties observed.
This result is unsurprising given the ${\sim}500{\times}$ cost span
between the cheapest and most expensive arm (per-model CVs in
0.63--0.92), and serves primarily as a sanity check that the dataset
is internally consistent
(Figure~\ref{fig:cost_heuristic_distributions}a;
Figure~\ref{fig:cost_heuristic_ranking}a).

\paragraph{$K{=}4$ portfolio (1766 shared validation prompts, with Gemini-Flash).}
Adding a fourth arm whose pricing overlaps an existing arm provides a
harder test of the heuristic.
For a clean comparison, we restrict both portfolios to the shared
prompt subset: 19 prompts from the original $K{=}3$ validation split
are excluded because the Flash data-collection pipeline failed to
produce a successful response or judge score for them, so they are
absent from the $K{=}4$ data file.
Gemini-Flash ($\tilde{c} = 0.382$) is placed between Mistral-Large
($\tilde{c} = 0.333$) and Gemini-Pro ($\tilde{c} = 0.583$).
The full ordering matches on 79.7\%
(CI: [77.7, 81.5]\%) of prompts.
The closest pair (Mistral-Large vs.\ Gemini-Flash) preserves ranking
on 79.7\% (CI: [77.7, 81.5]\%; 0 exact ties), reflecting Flash's
high cost variance (CV $= 1.56$) and the narrow heuristic gap
($\Delta\tilde{c} = 0.049$)
(Figure~\ref{fig:cost_heuristic_distributions}b;
Figure~\ref{fig:cost_heuristic_ranking}b).

\paragraph{Log-cost separation.}
To quantify tier separation without referencing the heuristic itself,
we examine per-model distributions in
$\log(\text{cost}_{\text{USD}})$ space.
Within-model standard deviations are 8--11\% of the total inter-model
log-cost range, confirming that prompt-level cost noise does not
overwhelm tier separation.
We report Cohen's $d$ between adjacent tiers as a scale-free measure
of how many within-model standard deviations separate each pair;
$d \gg 1$ implies that the two distributions rarely overlap, so a
single scalar per arm faithfully distinguishes them (condition~(ii)
above).
In the $K{=}3$ portfolio, Cohen's $d$ for adjacent pairs ranges from
4.39 (Llama-8B $\to$ Mistral-Large) to 5.92 (Mistral-Large $\to$
Gemini-Pro)---well above $d{=}1$, consistent with the 100\% ranking
preservation reported above.
In the $K{=}4$ portfolio, the Mistral-Large $\to$ Gemini-Flash pair
has $d = 0.68$; this weakest separation corresponds to the only pair
where the ranking inverts on ${\sim}20\%$ of prompts.

\paragraph{Prompt-cost correlation.}
Per-model cost distributions are right-skewed
(Figure~\ref{fig:cost_heuristic_distributions}), so we use Spearman
(rank-based) correlations between prompt word count and per-request
cost.
Correlations are statistically significant but modest in magnitude
($\rho = 0.12$ to $0.27$; all $p < 10^{-4}$).
Because $\rho^2 < 8\%$ (proportion of rank variance explained),
prompt-level features capture only a small fraction of cost
variation, limiting the practical benefit of a contextual cost model
in this portfolio.

\paragraph{Cross-model cost correlation.}
If per-request costs co-vary across models (e.g.\ because a long
prompt elicits long outputs from every model), then prompt-level cost
fluctuations shift all arms together, preserving relative ordering
even when absolute costs vary.
We again use Spearman correlations given the skewed cost
distributions.
Per-request costs are moderately to strongly correlated across models
($\rho = 0.56$ to $0.68$), confirming this shared output-length
factor and explaining why the static heuristic's ranking holds
despite high within-model CVs (0.63--1.56).

\paragraph{Takeaway.}
The static $\tilde{c}_a$ heuristic is empirically well-calibrated for
portfolios where models span at least one order of magnitude in
pricing.
For denser price clusters (e.g., Mistral-Large vs.\ Gemini-Flash,
$d = 0.68$), the ranking holds for ${\sim}79.7\%$ of prompts but
inverts when Flash generates unusually short responses.
Three structural properties limit downstream impact:
(i)~$\tilde{c}_a$ enters only the soft penalty
(Eq.~\ref{eq:budget_ucb}); the hard ceiling and the pacer's EMA cost
signal operate on actual per-request costs, so heuristic errors
cannot cause budget violations.
(ii)~For closely priced arms the absolute penalty gap is small
(${\sim}0.015$ for Mistral--Flash at $\lambda_t{=}0$)---negligible
relative to the $[0,1]$ reward signal.
(iii)~The dual variable $\lambda_t$ updates on actual costs
(Eq.~\ref{eq:dual_update}), self-correcting any persistent
over-selection of a costly arm.

\begin{figure}[!htbp]
\centering
\includegraphics[width=\columnwidth]{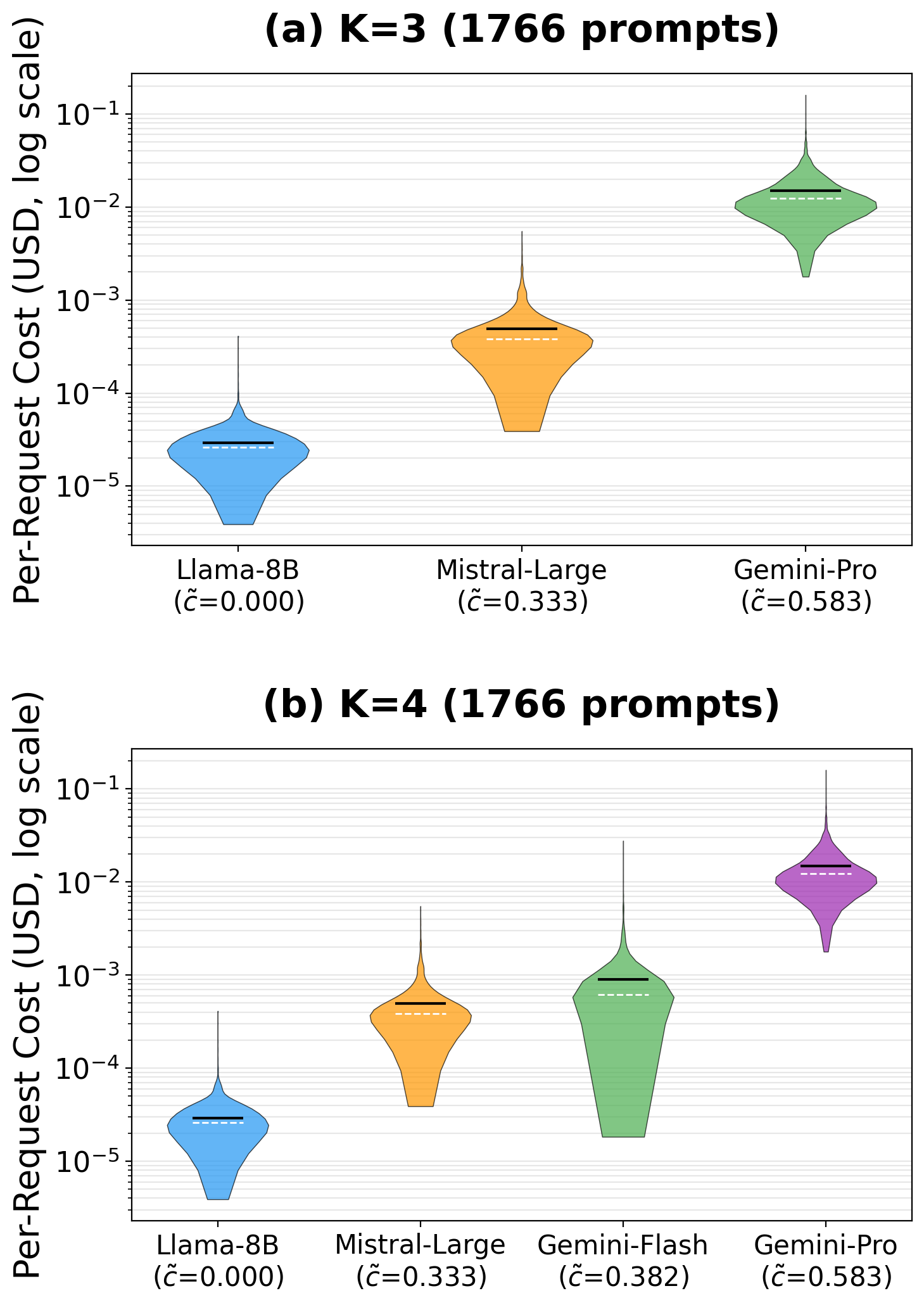}
\caption{Cost heuristic validation: realised cost distributions. Panel (a) shows per-request realised cost distributions for the $K{=}3$ portfolio on the shared validation subset, and panel (b) shows the same diagnostic for the $K{=}4$ portfolio. In both cases the violin plots are ordered by the static log-normalized heuristic $\tilde{c}_a$, illustrating that the major cost tiers are cleanly separated, with only the Mistral-Large and Gemini-Flash distributions showing substantial overlap.}
\label{fig:cost_heuristic_distributions}
\end{figure}

\begin{figure}[!htbp]
\centering
\includegraphics[width=\columnwidth]{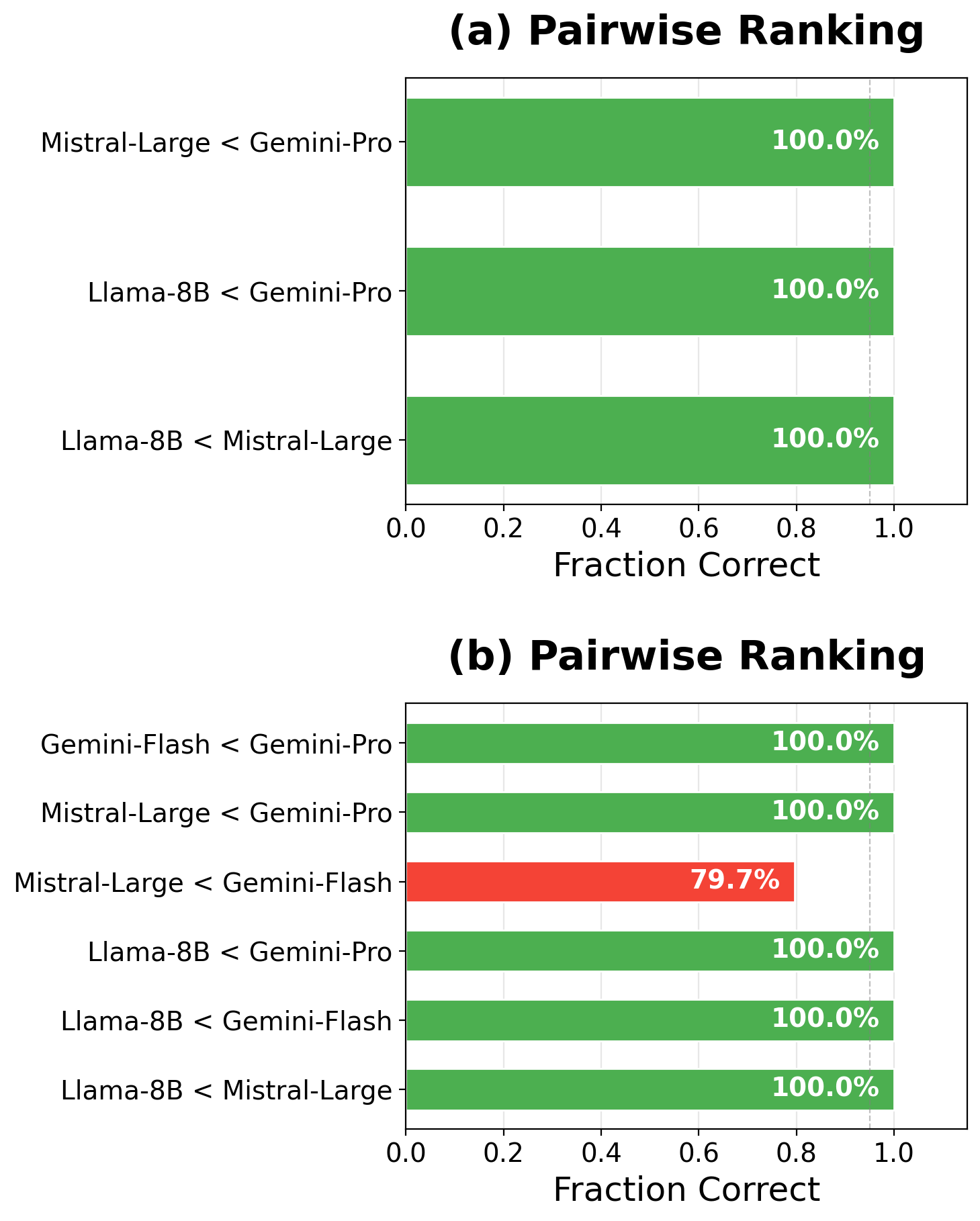}
\caption{Cost heuristic validation: ranking preservation. Panel~(a) shows pairwise ranking preservation for the $K{=}3$ portfolio on the shared validation subset, and panel~(b) shows the same diagnostic for the $K{=}4$ portfolio. The heuristic preserves the full per-request cost ordering on \chFullOrderK\% of shared-subset $K{=}3$ prompts and ${\sim}$\chFullOrderKFour\% of shared-subset $K{=}4$ prompts; the Mistral-Large vs.\ Gemini-Flash comparison is the only source of inversions.}
\label{fig:cost_heuristic_ranking}
\end{figure}


\section{Cold-Start vs Warmup Priors}
\label{sec:warmup_ablation}

Section~\ref{sec:warmup} introduces offline-to-online warmup priors,
which let the router begin with informed beliefs about each model's
strengths by loading sufficient statistics fitted on historical data.
The natural question is whether warmup priors reduce cold-start
regret in practice.
To answer this, we compare ParetoBandit with warmup priors
($\alpha{=}0.01$, $n_{\text{eff}}{=}1164$,
$\gamma{=}0.997$) against Tabula Rasa
($\alpha{=}0.05$, $n_{\text{eff}}{=}1$,
$\gamma{=}0.997$), each running under its own
Pareto-optimal hyperparameters tuned independently via the
Pareto-knee procedure (Appendix~\ref{sec:hparam_sweep}).
Notably, the sweep independently selected the same
$\gamma{=}0.997$ for both conditions, so they share identical
effective-memory windows (${\sim}333$ steps) and the only remaining
differences are prior strength ($n_{\text{eff}}{=}1164$ vs.\ $1$)
and exploration rate ($\alpha{=}0.01$ vs.\ $0.05$).
Any regret difference is therefore directly attributable to the
informative priors, not to a memory-length confound.

The comparison uses the $K{=}3$ stationary portfolio under four
budget regimes: unconstrained (no pacer), tight
($B{=}3{\times}10^{-4}$/request), moderate
($B{=}6.6{\times}10^{-4}$/request), and loose
($B{=}1.9{\times}10^{-3}$/request).

\paragraph{Statistical methodology.}
We assess each comparison along two dimensions.
The first is typical-case regret: does warmup tend to outperform
seed by seed?
The second is tail risk: does warmup avoid outlier runs where regret
is disproportionately high?
We call a seed a catastrophic failure when its cumulative regret
exceeds $2{\times}$ the pooled median across all conditions in that
budget regime; pooling ensures no single condition anchors the
threshold.
These two dimensions require different tests because a method can
improve the median while still leaving heavy tails, or eliminate
outliers without shifting the centre.
\begin{itemize}[nosep,leftmargin=*]
  \item \textbf{Sign test} (exact binomial): tests whether warmup
    has lower regret seed by seed, capturing the typical case
    ($H_0\colon P(\text{warmup wins}) = 0.5$).
  \item \textbf{Fisher exact test} on a $2{\times}2$ table of
    catastrophic vs.\ non-catastrophic seeds: tests whether warmup
    has a lower catastrophic-failure rate, capturing tail risk
    ($H_0\colon$ equal failure rates).
\end{itemize}
All $p$-values are corrected for multiplicity via the
Holm--Bonferroni procedure across the full family of tests
(4 sign tests and 4 Fisher tests, corrected separately per family).
Bootstrap CIs for paired differences are likewise
Bonferroni-corrected (4 simultaneous comparisons) to maintain 95\%
family-wise coverage.

\begin{table*}[t]
\centering
\caption{Warmup-prior ablation results across budget regimes (\waNseeds~seeds, held-out test split, $n{=}\waNPrompts$). Warmup priors are compared against Tabula Rasa and a random baseline (unconstrained only). Regret is cumulative over all \waNPrompts{} test steps; R@\waEarlyStep{} is cumulative regret at test step~\waEarlyStep. CIs are \waCILevel\% percentile-bootstrap (\waBootstrapResamples{} resamples). $^*$Holm--Bonferroni corrected. \textbf{Cat.}\ = catastrophic failures (regret $> \waCatMultiplier{\times}$ pooled median of all conditions in each budget regime).}
\label{tab:warmup_ablation_results}
\small
\resizebox{\textwidth}{!}{%
\begin{tabular}{@{}llrrrrrcc@{}}
\toprule
\textbf{Budget} & \textbf{Condition} & \textbf{Regret} (\waCILevel\% CI) & \textbf{Std} & \textbf{R@\waEarlyStep} (\waCILevel\% CI) & \textbf{Rwd} & \textbf{Cat.} & $p_\text{sign}^*$ & $p_\text{Fisher}^*$ \\
\midrule
None     & Warmup           & $\mathbf{\waUncWarmupRegret}$\,[\waUncWarmupRegretCILo, \waUncWarmupRegretCIHi]    & $\waUncWarmupRegretStd$   & $\mathbf{\waUncWarmupRAtTwoHundred}$\,[\waUncWarmupRAtTwoHundredCILo, \waUncWarmupRAtTwoHundredCIHi]    & $\mathbf{\waUncWarmupReward}$ & \waUncVsTRWarmupCat/\waNseeds & --- & ---  \\
         & Tabula Rasa      & $\waUncTRRegret$\,[\waUncTRRegretCILo, \waUncTRRegretCIHi]            & $\waUncTRRegretStd$  & $\waUncTRRAtTwoHundred$\,[\waUncTRRAtTwoHundredCILo, \waUncTRRAtTwoHundredCIHi]           & $\waUncTRReward$          & \waUncVsTRBaselineCat/\waNseeds & $\waUncVsTRSignP$ & $\waUncVsTRFisherP$ \\
         & Random           & $\waUncRandomRegret$\,[\waUncRandomRegretCILo, \waUncRandomRegretCIHi]           & $\waUncRandomRegretStd$   & $\waUncRandomRAtTwoHundred$\,[\waUncRandomRAtTwoHundredCILo, \waUncRandomRAtTwoHundredCIHi]           & $\waUncRandomReward$          & ---  & --- & --- \\
\midrule
Tight    & Warmup           & $\mathbf{\waTightWarmupRegret}$\,[\waTightWarmupRegretCILo, \waTightWarmupRegretCIHi]  & $\waTightWarmupRegretStd$   & $\mathbf{\waTightWarmupRAtTwoHundred}$\,[\waTightWarmupRAtTwoHundredCILo, \waTightWarmupRAtTwoHundredCIHi] & $\mathbf{\waTightWarmupReward}$ & \waTightVsTRWarmupCat/\waNseeds & --- & --- \\
         & Tabula Rasa      & $\waTightTRRegret$\,[\waTightTRRegretCILo, \waTightTRRegretCIHi]            & $\waTightTRRegretStd$  & $\waTightTRRAtTwoHundred$\,[\waTightTRRAtTwoHundredCILo, \waTightTRRAtTwoHundredCIHi]           & $\waTightTRReward$          & \waTightVsTRBaselineCat/\waNseeds & $\waTightVsTRSignP$ & $\waTightVsTRFisherP$ \\
\midrule
Moderate & Warmup           & $\mathbf{\waModWarmupRegret}$\,[\waModWarmupRegretCILo, \waModWarmupRegretCIHi]  & $\waModWarmupRegretStd$  & $\mathbf{\waModWarmupRAtTwoHundred}$\,[\waModWarmupRAtTwoHundredCILo, \waModWarmupRAtTwoHundredCIHi] & $\mathbf{\waModWarmupReward}$ & \waModVsTRWarmupCat/\waNseeds & --- & --- \\
         & Tabula Rasa      & $\waModTRRegret$\,[\waModTRRegretCILo, \waModTRRegretCIHi]            & $\waModTRRegretStd$  & $\waModTRRAtTwoHundred$\,[\waModTRRAtTwoHundredCILo, \waModTRRAtTwoHundredCIHi]           & $\waModTRReward$          & \waModVsTRBaselineCat/\waNseeds & $\waModVsTRSignP$ & $\waModVsTRFisherP$ \\
\midrule
Loose    & Warmup           & $\mathbf{\waLooseWarmupRegret}$\,[\waLooseWarmupRegretCILo, \waLooseWarmupRegretCIHi]     & $\waLooseWarmupRegretStd$   & $\mathbf{\waLooseWarmupRAtTwoHundred}$\,[\waLooseWarmupRAtTwoHundredCILo, \waLooseWarmupRAtTwoHundredCIHi]    & $\mathbf{\waLooseWarmupReward}$ & \waLooseVsTRWarmupCat/\waNseeds & --- & --- \\
         & Tabula Rasa      & $\waLooseTRRegret$\,[\waLooseTRRegretCILo, \waLooseTRRegretCIHi]            & $\waLooseTRRegretStd$  & $\waLooseTRRAtTwoHundred$\,[\waLooseTRRAtTwoHundredCILo, \waLooseTRRAtTwoHundredCIHi]           & $\waLooseTRReward$          & \waLooseVsTRBaselineCat/\waNseeds & $\waLooseVsTRSignP$ & $\waLooseVsTRFisherP$ \\
\bottomrule
\end{tabular}
}
\end{table*}
\vspace{8pt}

Table~\ref{tab:warmup_ablation_results} reports per-condition
means, bootstrap CIs, standard deviations, early-learning regret
(R@200), mean reward, catastrophic-failure counts, and
Holm-corrected $p$-values; below we highlight the main findings.

\paragraph{Result: warmup priors reduce cold-start regret.}
Relative to Tabula Rasa, warmup priors reduce mean cumulative regret
across all budget regimes.
We report the paired difference
$\Delta = \text{Tabula Rasa} - \text{Warmup}$ (positive favours
warmup) with Bonferroni-corrected 95\% bootstrap CIs (4 simultaneous
comparisons):
unconstrained $\Delta{=}32.0$\,[19.8, 48.4],
tight $\Delta{=}52.2$\,[29.6, 76.9],
moderate $\Delta{=}13.7$\,[$-$10.9, 38.5],
and loose $\Delta{=}30.4$\,[13.9, 49.2].
The CI excludes zero for unconstrained, tight, and loose regimes;
the moderate regime is inconclusive ($\Delta{=}13.7$, CI spans zero).
These reductions correspond to 9--37\% of Tabula Rasa's total regret
(largest in the unconstrained regime, smallest in the moderate regime
where the CI spans zero), reflecting that the cold-start baseline
converges within the ${\sim}333$-step effective-memory window
afforded by $\gamma{=}0.997$.
Much of this advantage is concentrated in the early learning phase.
The R@200 paired differences
($\Delta = \text{TR} - \text{Warmup}$, same sign convention)
are significant in every regime:
unconstrained $\Delta{=}10.0$\,[4.7, 15.8],
tight $\Delta{=}9.7$\,[5.5, 13.7],
moderate $\Delta{=}8.8$\,[3.6, 13.8],
and loose $\Delta{=}13.6$\,[8.2, 18.8]
(see also per-condition R@200 CIs in
Table~\ref{tab:warmup_ablation_results}).
The advantage narrows as online evidence accumulates beyond the
effective-memory window.

\paragraph{Cold-start variability.}
Warmup priors produce tighter per-seed regret distributions
(Figure~\ref{fig:warmup_ablation}).
Under tight budgets, the per-seed regret standard deviation for
warmup is $4.6$ compared to $41.8$ for Tabula Rasa
(${\sim}9{\times}$ lower).
Warmup records zero catastrophic failures
(regret $> 2{\times}$ pooled median) across all regimes.
Tabula Rasa produces 2/20 catastrophic failures in the
unconstrained regime, with none under budget constraints
(0/20 tight, 0/20 moderate, 0/20 loose); Fisher tests are
non-significant after Holm correction
($p^*_{\text{Fisher}} = 0.974$).
The occasional cold-start outliers confirm that starting without
priors introduces tail risk, even when the effective-memory window
is the same ($\gamma{=}0.997$ for both conditions).

\paragraph{Discussion.}
Warmup priors are best understood as a deployment-reliability
mechanism: they reduce cold-start regret and tighten the per-seed
distribution.
Warmup is optional---ParetoBandit is fully functional without it,
and the Tabula Rasa results in
Table~\ref{tab:warmup_ablation_results} confirm that the bandit
converges to comparable steady-state quality once sufficient online
evidence accumulates.
The warmup benefit is transient by design: geometric forgetting
($\gamma{=}0.997$) replaces priors within ${\sim}333$
effective-memory steps, after which warmup and tabula-rasa
conditions converge.
Enabling warmup priors reduces regret during this early learning
window and tightens run-to-run variability, but practitioners who
lack historical data or prefer a simpler deployment can omit them
without sacrificing long-run performance.
When warmup priors are used, these gains assume the prior is
directionally correct; a dedicated prior-mismatch sensitivity
analysis (Appendix~\ref{appendix:prior_mismatch}) quantifies
trade-offs across five quality levels---from well-calibrated through
domain-restricted to actively inverted priors.

\vspace{-6pt}
\begin{figure}[t]
\centering
\setlength{\abovecaptionskip}{1.5pt}
\includegraphics[width=\columnwidth]{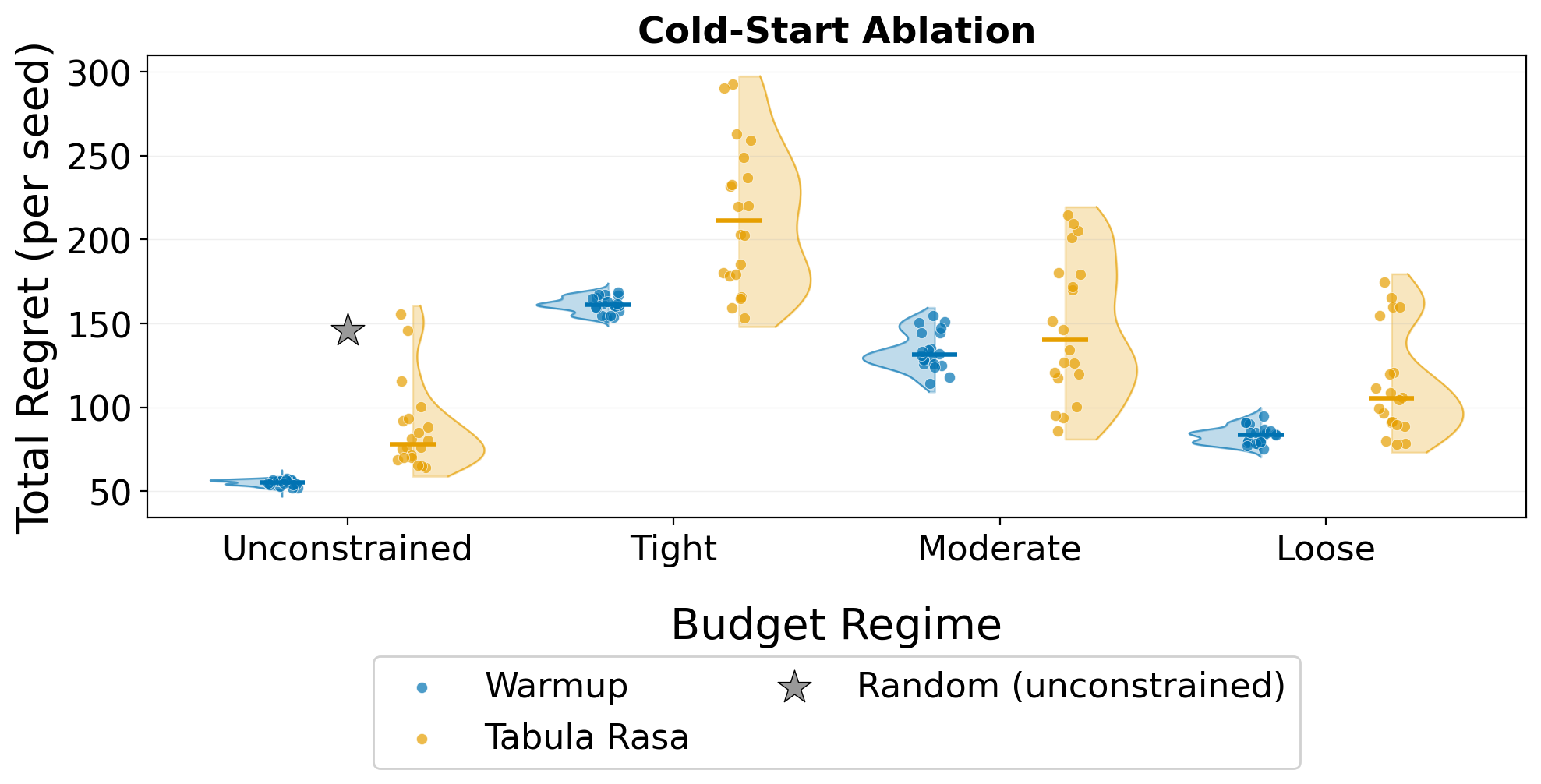}
\caption{Per-seed cumulative regret distributions for warmup priors (blue) vs.\ Tabula Rasa (orange) across four budget regimes ($K{=}\waK$, \waNseeds~seeds, held-out test split). Both conditions share $\gamma{=}\waGamma$ but differ in prior strength and exploration rate (deployment-level comparison). Gray star: Random baseline median (unconstrained only). Warmup priors yield tight, unimodal distributions; Tabula Rasa exhibits heavier tails under budget constraints. Statistical tests in Table~\ref{tab:warmup_ablation_results}.}
\label{fig:warmup_ablation}
\end{figure}


\section{Prior Mismatch Sensitivity}
\label{appendix:prior_mismatch}

The warmup ablation (Appendix~\ref{sec:warmup_ablation}) demonstrated
that well-calibrated priors substantially reduce regret and eliminate
catastrophic cold-start failures under stationary conditions.
All experiments in this paper use priors trained on a representative
training set drawn from the same distribution as the test traffic.
In practice, however, prior quality degrades after deployment:
the prompt distribution shifts as users adopt new workflows,
model providers silently update weights, and the roster itself
changes as models are added or retired.
A practitioner deploying warmup priors needs to know how wrong the
prior can be before cold start becomes the safer option, and whether
a conservative prior strength can limit the damage when calibration
quality is uncertain.

\paragraph{Why not re-optimise?}
The production hyperparameters (Appendix~\ref{sec:hparam_sweep})
are tuned once on a representative training set via the
$T_\text{adapt}$-constrained Pareto-knee procedure, which
couples $n_\mathit{eff}$ and $\gamma$ through the adaptation horizon.
The selected prior strength ($n_\mathit{eff} \approx 1{,}164$,
Eq.~\ref{eq:prior_scale}) implicitly assumes the prior is
directionally correct: high $n_\mathit{eff}$ tells the bandit to
trust the offline beliefs for many observation-equivalents before
online evidence takes over.
If the prior is wrong, high $n_\mathit{eff}$ amplifies the harm by
delaying the point at which online data overrides the prior.
Re-running the Pareto-knee sweep for every possible prior-quality
level is impractical in production, so this experiment instead
asks whether a single tunable parameter---$n_\mathit{eff}$---can
serve as a safety knob that limits damage from degraded priors
while preserving the variance-reduction benefit for good ones.

\paragraph{Design: mismatch gradient.}
We construct five prior-quality levels forming a mismatch gradient,
ordered from ideal to adversarial:
\begin{enumerate}[nosep,leftmargin=*]
  \item \textbf{Well-calibrated:} priors from the full training set
    ($n{=}8{,}373$)---the shipped default.
  \item \textbf{Random-1680:} a random subsample of 1{,}680 training
    prompts, matching the GSM8K-only count while preserving the full
    distribution.  This is a sample-size control: any gap between
    Random-1680 and Well-calibrated isolates covariance-estimation
    noise from domain effects.
  \item \textbf{MMLU-only:} priors from the 1{,}855 MMLU prompts
    only.  The prior preserves the correct model ranking
    (Gemini~$>$~Mistral~$\gg$~Llama) but with
    knowledge-domain-specific magnitudes.
  \item \textbf{GSM8K-only:} priors from the 1{,}680 GSM8K (math)
    prompts only.  All models score ${\sim}0.86{+}$ on math, so the
    prior encodes near-zero arm differentiation.
  \item \textbf{Inverted:} priors from the full training set with
    Llama and Gemini rewards swapped, so the prior believes the
    cheapest model is best and vice versa.
\end{enumerate}
These levels represent failure modes that arise naturally in
deployment: (1--2)~an operator who collected representative data;
(3)~an operator who collected data from a single knowledge domain;
(4)~an operator whose data happens to come from a task where all
models perform similarly; and (5)~the worst case, where an update
inverts the relative model ranking (e.g., because a provider
silently degrades a model that was previously the best).

\paragraph{Design: $n_\mathit{eff}$ as a safety knob.}
Each quality level is tested at three $n_\mathit{eff}$ values
(10, 100, 1000), spanning three orders of magnitude.
These values bracket a key trade-off.
Low $n_\mathit{eff}$ (${\approx}10$) treats the prior as a gentle
nudge: the bandit overrides it within tens of observations,
limiting the damage from a bad prior but also limiting the
cold-start benefit from a good one.
High $n_\mathit{eff}$ (${\approx}1{,}000$) trusts the prior for
hundreds of effective observations: beneficial when the prior is
correct, but potentially harmful when it confidently encodes the
wrong ranking.
Together with the quality gradient, the $5 \times 3$ grid reveals
whether there exists an $n_\mathit{eff}$ regime where any
prior---even a domain-mismatched one---helps more than it hurts.

\paragraph{Design: baseline.}
The no-prior baseline is the independently optimised Tabula Rasa
($\alpha{=}0.05$, $\gamma{=}0.997$).
As established in the warmup ablation
(Appendix~\ref{sec:warmup_ablation}), the Pareto-knee sweep selected
the same $\gamma{=}0.997$ for both warmup and Tabula Rasa, so
both share the same effective memory (${\sim}333$ steps)
and the comparison is not driven by different forgetting horizons.
However, because $\alpha$ also differs between conditions
($\alpha{=}0.01$ for warmup vs.\ $0.05$ for Tabula Rasa),
the appendix should be interpreted as a system-level deployment
comparison against the best no-prior baseline rather than a fully
controlled ablation that isolates prior information while holding
every learning hyperparameter fixed.
This yields $5 \times 3 + 1 = 16$ conditions.
The warmup hyperparameters were not re-optimised per
prior-quality level---matching the production scenario where prior
quality degrades after deployment without hyperparameter adjustment.

\paragraph{Design: unconstrained regime.}
Evaluation follows the same cumulative-regret protocol as the warmup
ablation: held-out test split ($n{=}1824$), 20 seeds
(offset${=}9{,}000$, aligned for paired comparisons), unconstrained
(no budget pacer).
We deliberately restrict this experiment to the unconstrained regime
for two reasons.
First, the budget pacer constrains arm selection, which limits the
online observations the bandit receives to correct a bad prior;
budget-induced failures and prior-induced failures would be
confounded.
Second, the warmup ablation (Appendix~\ref{sec:warmup_ablation})
already characterises the pacer interaction for well-calibrated
priors vs.\ tabula rasa.
Adding budget constraints here would conflate two failure modes
without additional insight, since the prior-quality threshold for
harm is determined by the bandit's learning dynamics, not the pacer.

\paragraph{Statistical methodology.}
We use the same paired-comparison protocol as
Appendix~\ref{sec:warmup_ablation}: exact binomial sign test for
location shift and Fisher exact test on a $2{\times}2$
catastrophic-failure table for tail risk, with Holm--Bonferroni
correction across all 15 pairwise comparisons per baseline.
All confidence intervals are 95\% percentile-bootstrap
(10{,}000 resamples, seed-level resampling);
median CIs resample the median directly rather than relying on
normality, which is inappropriate for the heavy-tailed baseline
distributions observed here.
A seed is classified as catastrophic if its regret exceeds
$2{\times}$ the Tabula Rasa median.

\vspace*{-12pt}
\paragraph{Results.}
Figure~\ref{fig:prior_mismatch_heatmap} summarises total regret
across the quality--strength grid.

\begin{figure*}[t]
\centering
\includegraphics[width=\textwidth]{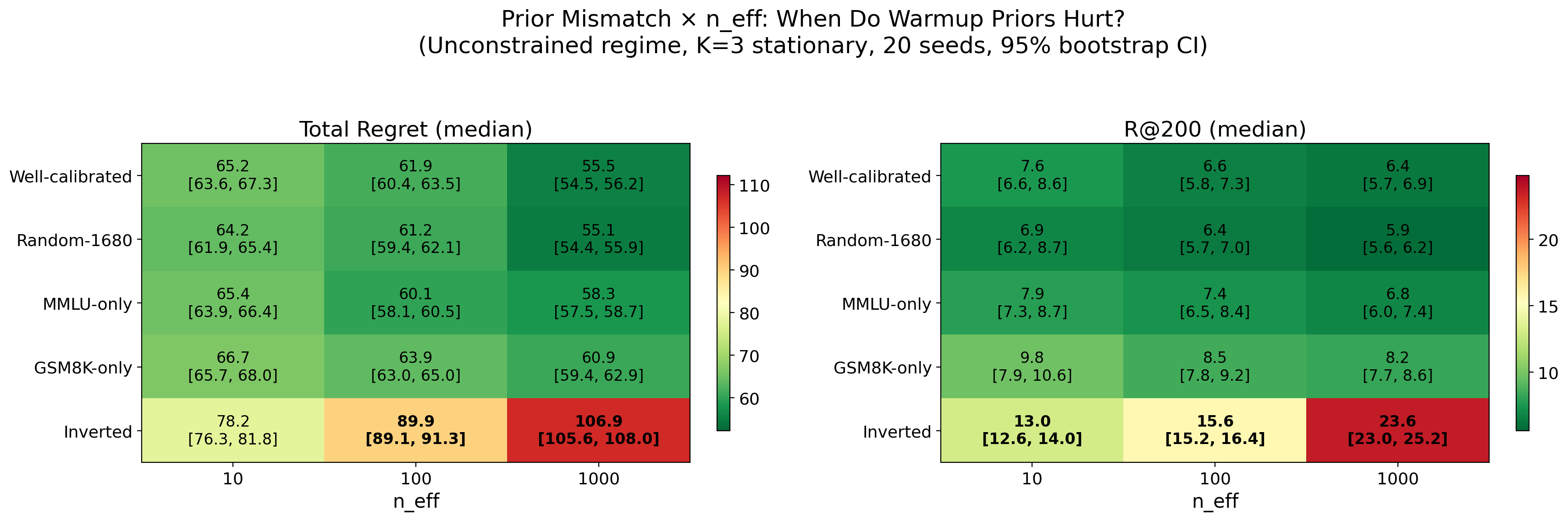}
\caption{Total cumulative regret for ParetoBandit across the prior-quality vs.\ prior-strength grid (20~seeds, unconstrained, test split).  All conditions use the same ParetoBandit router; only the warmup prior varies.  Tabula Rasa is the independently optimised no-prior baseline.  \textbf{Bold} cells exceed the Tabula Rasa median; normal-weight cells fall below it.}
\label{fig:prior_mismatch_heatmap}
\end{figure*}

\emph{Directionally correct priors help at every strength.}
Well-calibrated priors reduce median total regret monotonically
with $n_\mathit{eff}$, relative to the Tabula Rasa baseline
(median $78.2$\,[70.7, 90.1]):
\begin{itemize}[nosep,leftmargin=*]
  \item $n_\mathit{eff}{=}10$: median
    $65.2$\,[63.6, 67.3]
    ($16.6\%$\,[6.5, 28.5] reduction).
  \item $n_\mathit{eff}{=}100$: median
    $61.9$\,[60.4, 63.5]
    ($20.9\%$\,[11.9, 32.1] reduction).
  \item $n_\mathit{eff}{=}1{,}000$: median
    $55.5$\,[54.5, 56.2]
    ($29.1\%$\,[21.7, 38.8] reduction).
\end{itemize}
All well-calibrated and Random-1680 comparisons are significant after
Holm correction at every $n_\mathit{eff}$
($p \leq 0.0103$ even at $n_\mathit{eff}{=}10$),
with the strongest effect at $n_\mathit{eff}{=}1{,}000$:
Well-calibrated wins 20 of 20 seeds ($p < 10^{-4}$);
Random-1680 wins 20 of 20 ($p < 10^{-4}$).

\emph{Sample size does not matter---distributional match does.}
A practitioner might worry that the Well-calibrated prior benefits
from a larger training set rather than better calibration.
The Random-1680 control rules this out: it uses only 1{,}680
prompts (matching the GSM8K-only count) drawn from the full
distribution and performs comparably to Well-calibrated at every
$n_\mathit{eff}$ (e.g., median
$55.1$\,[54.4, 55.9]
vs.\ $55.5$\,[54.5, 56.2]
at $n_\mathit{eff}{=}1{,}000$).
This confirms that covariance estimation saturates
quickly---1{,}680 representative prompts are sufficient for
$d{=}26$ features---and that the domain-mismatch findings
below are not confounded by sample-count effects.
The practical implication is that operators need a
\emph{representative} training sample, not a large one.

\emph{Domain-mismatched priors still beat cold start.}
Even when the prior is trained on a single domain rather than
representative traffic, median regret stays below the Tabula Rasa
baseline at every $n_\mathit{eff}$---the prior never hurts.
MMLU-only priors reduce median regret to
$60.1$\,[58.1, 60.5]
at $n_\mathit{eff}{=}100$ and
$58.3$\,[57.5, 58.7]
at $1{,}000$
(vs.\ Tabula Rasa $78.2$\,[70.7, 90.1]),
despite being trained on a single knowledge domain.
GSM8K-only priors---which encode near-zero arm
differentiation---lower median regret to
$66.7$\,[65.7, 68.0]
through
$60.9$\,[59.4, 62.9]
across $n_\mathit{eff}$ values.
These reductions are statistically significant after Holm correction
(Holm-corrected $p < 0.004$ for all domain-mismatched conditions
at every $n_\mathit{eff}$).
Crucially, no domain-mismatched prior \emph{increases} median regret
at any $n_\mathit{eff}$: partial knowledge is consistently at least
as good as no knowledge.

\emph{Inverted priors cause harm that scales with $n_\mathit{eff}$.}
This is the only condition where priors hurt.
When the prior confidently encodes the wrong model ranking
(cheapest = best), the router initially routes traffic to the worst
model and must accumulate online evidence to override the prior.
Higher $n_\mathit{eff}$ delays that override, so damage grows with
prior strength (vs.\ Tabula Rasa median
$78.2$\,[70.7, 90.1]):
\begin{itemize}[nosep,leftmargin=*]
  \item $n_\mathit{eff}{=}10$: median
    $78.2$\,[76.3, 81.8]
    (modest; the prior is overridden quickly).
  \item $n_\mathit{eff}{=}100$: median
    $89.9$\,[89.1, 91.3].
  \item $n_\mathit{eff}{=}1{,}000$: median
    $106.9$\,[105.6, 108.0]---a
    $37\%$ increase over the baseline
    ($+29$~regret units\,[16.3, 36.4]).
\end{itemize}
\noindent
Tabula Rasa wins 17 of 20 seeds
at $n_\mathit{eff}{=}1{,}000$
(Holm $p{=}0.0103$), confirming that the
degradation is both statistically significant and operationally
meaningful.

\paragraph{Reliability: the more important finding.}
For a practitioner, run-to-run consistency matters as much as average
regret.
A router that usually improves but occasionally catastrophically
misroutes is harder to trust than one that reliably delivers moderate
gains.
Figure~\ref{fig:prior_mismatch_distribution} shows that warmup priors
deliver exactly this consistency: \emph{every} warmup
condition---including domain-mismatched and even inverted
priors---produces a tight, unimodal per-seed regret distribution,
while Tabula Rasa exhibits a wide spread
($\mathit{std}{=}24.8$).
Concretely, non-inverted warmup conditions have
$\mathit{std} \leq 3.2$ (range: $1.6$--$3.2$); even the worst
condition (Inverted at $n_\mathit{eff}{=}10$) achieves
$\mathit{std}{=}4.9$---still far tighter than Tabula Rasa's $24.8$.
All 15 warmup conditions record zero catastrophic failures
(regret $> 2{\times}$ the Tabula Rasa median, i.e., $> 156.5$).
The warmup ablation (Appendix~\ref{sec:warmup_ablation}) shows that
Tabula Rasa exhibits substantially higher per-seed variance
than warmup priors across all budget regimes, with
${\sim}9{\times}$ larger standard deviation under tight budgets;
the present experiment extends that finding by showing that the
reliability benefit is robust to prior degradation.

\paragraph{Practical implications.}
The prior-quality threshold for harm is high: priors need to be
actively wrong (inverting model rankings) before they degrade
performance.
The reliability benefit---tight, unimodal distributions vs.\ Tabula
Rasa's heavy tail---is robust across the entire quality gradient.
The $n_\mathit{eff}$ parameter serves as a safety knob:
$1{,}000$ for representative priors
($29.1\%$\,[21.7, 38.8] regret reduction),
$100$ as a safe default for domain-shifted priors
($20.9\%$\,[11.9, 32.1] reduction),
and $10$ when quality is entirely uncertain
(worst-case damage limited to ${\sim}0\%$, while non-adversarial
priors still reduce regret by $14.8$--$17.9\%$).
Tabula rasa is appropriate only when the prior is known to be
systematically inverted.

\begin{figure*}[t]
\centering
\includegraphics[width=\textwidth]{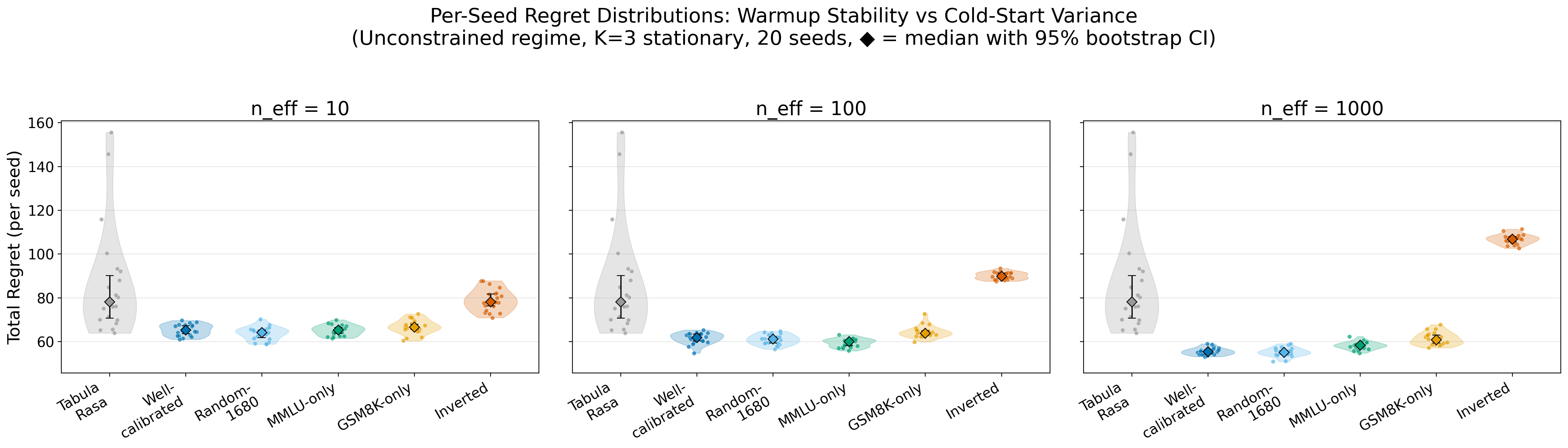}
\caption{Per-seed regret distributions across prior-quality levels at each $n_\mathit{eff}$ value (20~seeds, unconstrained).  Tabula Rasa is the independently optimised no-prior baseline.}
\label{fig:prior_mismatch_distribution}
\end{figure*}


\section{Reward Signal Robustness (Three-Judge Validation)}
\label{sec:judge_robustness}

All main-paper experiments use DeepSeek-R1 as the sole evaluator.
A natural concern is whether the paper's conclusions---the relative
ordering of methods and the qualitative behaviour of the router---are
artefacts of this particular judge.  We address this with a
three-judge validation aimed at a single question: would the paper's
conclusions change with a different evaluator?

\subsection{Setup}

We draw a stratified random sample of $n = 2{,}000$ prompts from the
12{,}000-prompt benchmark, preserving proportional representation of
all source benchmarks.  Each prompt has existing responses from all
three candidate models (Llama-3.1-8B, Mistral-Large-2512,
Gemini-2.5-Pro), yielding $6{,}000$ (prompt, model) evaluation pairs.
We re-evaluate these fixed responses---no new model outputs---with two
supplementary judges chosen for provider diversity: GPT-4.1-mini
(OpenAI) and Claude-3.7-Sonnet (Anthropic).

All three judges receive the same structured chain-of-thought
evaluation prompt (Figure~\ref{fig:judge_prompt}); DeepSeek-R1
additionally receives a brevity instruction to limit its CoT
verbosity.  The composite reward is a weighted average
$r = 0.4\,s_{\text{reason}} + 0.3\,s_{\text{instr}} +
0.3\,s_{\text{comm}}$.  The judge panel spans three independent
providers (DeepSeek, OpenAI, Anthropic), none of which overlaps with
the routed model providers (Meta, Mistral, Google).

\begin{figure*}[t]
\centering
\begingroup
\fontsize{5.6}{6.2}\selectfont
\begin{verbatim}
SYSTEM:
You are a Discriminative Router Judge. Your goal is to evaluate
how well an LLM response addresses the given prompt.

Score on three continuous dimensions (0.0-1.0). Use the FULL
range; do NOT default to 0 or 1.

1. Reasoning Quality (40%) - How sound is the reasoning?
   0.9-1.0  Flawless; every step correct and clearly justified.
   0.7-0.8  Sound overall; minor inefficiency or a trivial
            error that does not change the conclusion.
   0.5-0.6  Partially correct; approach is reasonable but
            important steps are wrong or missing.
   0.3-0.4  Weak; only fragments of correct logic.
   0.0-0.2  No coherent reasoning, or completely wrong approach.
   If the prompt needs no multi-step reasoning, score factual
   accuracy and depth of explanation.

2. Instruction Following (30%) - Were all explicit and implicit
   constraints satisfied?
   0.9-1.0  Every constraint followed precisely.
   0.7-0.8  All major constraints met; one minor instruction
            partially missed.
   0.5-0.6  Some important instructions missed or only
            partially addressed.
   0.3-0.4  Multiple instructions ignored or misinterpreted.
   0.0-0.2  Response largely ignores the prompt's requirements.

3. Communication Quality (30%) - How clear, well-structured,
   and useful is the response?
   0.9-1.0  Exceptionally clear, well-organized, appropriate
            detail.
   0.7-0.8  Clear and competent; minor improvements possible.
   0.5-0.6  Adequate but noticeably unclear, verbose, or poorly
            organized.
   0.3-0.4  Hard to follow; significant clarity issues.
   0.0-0.2  Unintelligible, unhelpful, or inappropriate tone.

Format your response EXACTLY as follows:

## Reasoning
<Concise chain-of-thought analysis>

## Reasoning Quality
<0.0 to 1.0>

## Instruction Following
<0.0 to 1.0>

## Communication Quality
<0.0 to 1.0>

USER:
PROMPT: {prompt}

RESPONSE: {response}
\end{verbatim}
\endgroup
\caption{Structured CoT evaluation prompt sent to all three judges.
The composite reward is
$r = 0.4\,s_{\text{reason}} + 0.3\,s_{\text{instr}} + 0.3\,s_{\text{comm}}$.
For DeepSeek-R1 only, the system message appends: ``\texttt{IMPORTANT:
Keep the \#\# Reasoning section to 3--5 sentences.  Be direct---identify
errors or confirm correctness, then move to scoring.}''
Temperature is set to 0 for all judges.}
\label{fig:judge_prompt}
\end{figure*}

\subsection{Population-Level Agreement}

The bandit's converged policy depends on the expected reward ordering
$\mathbb{E}[r(x, a)]$ across arms, not on individual prompt scores.
If this ordering is preserved across judges, the converged policy is
invariant to evaluator choice.

\begin{table}[h]
\centering
\caption{Expected reward ordering under each judge ($n = 2{,}000$
prompts; Bonferroni-corrected bootstrap CIs, 9 comparisons,
$B = 10{,}000$).}
\label{tab:expected_ordering}
\small
\resizebox{\columnwidth}{!}{%
\begin{tabular}{lccc}
\toprule
\textbf{Judge} & \textbf{Gemini-Pro} & \textbf{Mistral-Large} & \textbf{Llama-8B} \\
\midrule
DeepSeek-R1       & \jrROneGeminiMean{} & \jrROneMistralMean{} & \jrROneLlamaMean{} \\
GPT-4.1-mini      & \jrGPTGeminiMean{} & \jrGPTMistralMean{} & \jrGPTLlamaMean{} \\
Claude-3.7-Sonnet & \jrClaudeGeminiMean{} & \jrClaudeMistralMean{} & \jrClaudeLlamaMean{} \\
\bottomrule
\end{tabular}
}
\end{table}

Table~\ref{tab:expected_ordering} shows that all three judges rank
the models identically: Gemini-Pro $>$ Mistral-Large $>$ Llama-8B.
Every pairwise expected-reward gap has the same sign, with all nine
Bonferroni-corrected bootstrap CIs excluding zero (narrowest lower
bound 0.0046; 9 comparisons, $B = 10{,}000$).
Gap-conditioned Kendall's~$W$ (Table~\ref{tab:routing_diagnostics})
reaches 0.71 on prompts with large inter-model gaps, indicating that
the conditional ordering $\mathbb{E}[r(x,a) \mid x]$ is also largely
preserved where routing decisions matter most.

Because all main-paper comparisons are within-judge (same R1 rewards,
same R1 oracle), judge choice can change the magnitude of regret
differences but not their sign.  R1's oracle lift (0.031 per step) is
the largest among the three judges (0.016--0.020 for the supplementary
judges), so main-paper effect sizes correspond to the most favourable
case for adaptive routing; all qualitative conclusions hold under every
judge tested.

\paragraph{Cross-judge routing regret.}
Even where per-prompt routing decisions differ, the practical cost is
small.  Following R1's oracle policy and evaluating with the other
judges' scores yields ${\geq}97.4\%$ of their own oracle reward
(Table~\ref{tab:cross_regret}).  The reverse is worse: following
GPT-4.1-mini's or Claude's oracle and evaluating with R1 captures only
$95.8\%$ of R1's oracle reward.  The asymmetry is partly mechanical:
judges with smaller inter-model gaps inflate capture fractions for any
routing policy, while R1's larger gaps penalise misrouting more
heavily.

\begin{table}[h]
\centering
\caption{Cross-judge routing evaluation.  Each cell reports the mean
reward achieved by following the row judge's oracle, evaluated by the
column judge's scores.  Parenthetical values show the fraction of the
column judge's own oracle reward captured ($B = 10{,}000$ bootstrap
resamples).}
\label{tab:cross_regret}
\small
\resizebox{\columnwidth}{!}{%
\begin{tabular}{lccc}
\toprule
\textbf{Train $\downarrow$ / Eval $\rightarrow$}
    & \textbf{R1} & \textbf{GPT-4.1-mini} & \textbf{Claude-3.7} \\
\midrule
R1
    & \jrCrossROneROneMean{} (100\%) & \jrCrossROneGPTMean{} (\textbf{\jrCrossROneGPTCapture\%}) & \jrCrossROneClaudeMean{} (\textbf{\jrCrossROneClaudeCapture\%}) \\
GPT-4.1-mini
    & \jrCrossGPTROneMean{} (\jrCrossGPTROneCapture\%) & \jrCrossGPTGPTMean{} (100\%) & \jrCrossGPTClaudeMean{} (\jrCrossGPTClaudeCapture\%) \\
Claude-3.7-Sonnet
    & \jrCrossClaudeROneMean{} (\jrCrossClaudeROneCapture\%) & \jrCrossClaudeGPTMean{} (\jrCrossClaudeGPTCapture\%) & \jrCrossClaudeClaudeMean{} (100\%) \\
\bottomrule
\end{tabular}
}
\end{table}
\vspace{6pt}

Panel averaging~\cite{zheng2023judging} compresses routing margins:
averaging the three judges shrinks the mean inter-model gap from
$0.0485$ (R1 alone) to $0.0338$ (30\% compression), halving the
fraction of actionable prompts (31\% vs.\ 17\% at a 0.05 threshold).
Z-score calibration~\cite{hedges1985statistical} and multi-judge
aggregation methods such as PoLL~\cite{verga2024poll} and
CARE~\cite{zhao2025care} recover $< 3$ percentage points of this
compression, indicating genuine inter-judge disagreement rather than
pure scale mismatch.  R1 alone provides the largest margins and the
lowest per-evaluation cost.

\subsection{Per-Prompt Agreement}

Table~\ref{tab:judge_agreement} reports per-response agreement using
distribution-independent metrics robust to ceiling-compressed rewards
(most scores in $[0.8, 1.0]$).  Spearman's $\rho$ (0.633--0.658) and
Kendall's $\tau_b$ (0.528--0.547) indicate moderate rank agreement;
mean absolute deviation is ${\approx}0.075$ and Bland--Altman limits
of agreement remain within roughly $\pm 0.3$.  This level of
LLM-as-judge noise is expected, and the key point is that it preserves
population-level ordering and routing policies.

\begin{table}[h]
\centering
\caption{Per-response agreement with DeepSeek-R1 ($6{,}000$ scored pairs
per judge).  All measures are rank-based or distribution-independent to avoid
attenuation from ceiling-compressed rewards.}
\label{tab:judge_agreement}
\small
\resizebox{\columnwidth}{!}{%
\begin{tabular}{lcc}
\toprule
\textbf{Metric} & \textbf{GPT-4.1-mini} & \textbf{Claude-3.7-Sonnet} \\
\midrule
Spearman's $\rho$            & \textbf{\jrGPTSpearman} & \jrClaudeSpearman{} \\
Kendall's $\tau_b$           & \textbf{\jrGPTKendall} & \jrClaudeKendall{} \\
MAD                          & \textbf{\jrGPTMAD} & \jrClaudeMAD{} \\
Mean bias (judge $-$ R1)     & $+\jrGPTBias$ & $\jrClaudeBias$ \\
95\% Bland--Altman LoA       & [$\jrGPTBALo$, $+\jrGPTBAHi$] & [$\jrClaudeBALo$, $+\jrClaudeBAHi$] \\
\bottomrule
\end{tabular}
}
\end{table}

\paragraph{Gap-conditioned concordance.}
Per-prompt best-model agreement with R1 is ${\sim}50\%$ overall, but
disagreements cluster on prompts where all models are nearly tied.
Table~\ref{tab:routing_diagnostics} shows that when R1's inter-model
gap is $< 0.05$ (30\% of prompts), Kendall's $W = 0.17$ and
best-model agreement is 48--58\%; when the gap is $\geq 0.20$ (37\%
of prompts), $W$ rises to 0.57--0.71 and best-model agreement to
50--58\%.  In the low-gap region, choosing the ``wrong'' model costs
at most 0.05 reward---less than per-response judge noise
(MAD ${\approx} 0.075$)---so disagreement is concentrated where it is
cheap.

\begin{table}[h]
\centering
\caption{Routing agreement conditioned on R1's inter-model gap.
Kendall's~$W$ measures three-judge concordance on the full model ranking.
Conditioning on R1's gap bins when validating
R1 can introduce selection bias; we verify in the reproducibility code
that the same monotonic trend holds when conditioning on the consensus
(panel-median) gap.}
\label{tab:routing_diagnostics}
\small
\resizebox{\columnwidth}{!}{%
\begin{tabular}{lrccc}
\toprule
& & & \multicolumn{2}{c}{\textbf{Best-model agr.\ vs.\ R1}} \\
\cmidrule(lr){4-5}
\textbf{R1 gap range} & $n$ & \textbf{Kendall $W$}
    & \textbf{GPT-mini} & \textbf{Claude} \\
\midrule
$[0.00, 0.05)$ & 603  & 0.17 & 57.5\% & 48.3\% \\
$[0.05, 0.10)$ & 367  & 0.29 & 37.6\% & 37.9\% \\
$[0.10, 0.20)$ & 294  & 0.43 & 44.9\% & 39.5\% \\
$[0.20, 0.30)$ & 184  & 0.57 & 53.3\% & 49.5\% \\
$[0.30, 1.00)$ & 552  & 0.71 & 56.3\% & 58.0\% \\
\midrule
\emph{Overall} & 2{,}000 & 0.42 & 51.3\% & 47.9\% \\
\bottomrule
\end{tabular}
}
\end{table}

\subsection{End-to-End Regret Under Alternative Judges}
\label{sec:cross_judge_regret}

We now test whether the bandit's learning dynamics---convergence shape,
sample complexity, and relative method ordering---are preserved when
the reward signal comes from a different judge, by re-running the
bandit simulation under GPT-4.1-mini and Claude-3.7-Sonnet.

\paragraph{Protocol.}
We split the 2{,}000 judge-robustness prompts into a stratified
$\frac{1}{3}$~validation / $\frac{2}{3}$~test partition
(672 / 1{,}328 prompts).  The bandit learns online during the
validation burn-in (no metrics), then cumulative regret is evaluated
on the held-out test split.  To avoid any data-leakage concern, we use
cold start (Tabula Rasa) only---no warmup priors---so this is the
harder test; if the bandit converges from cold start, warm-start
robustness follows.  We compare Tabula Rasa against a Random baseline
(uniform $1/K$) under four budget regimes (unconstrained, tight,
moderate, loose), with 20 seeds per condition.
All three judges share the same hyperparameters
($\alpha = 0.05$, $\gamma = 0.997$), tuned originally on R1;
judge-specific tuning would only improve the supplementary judges'
results, so this choice is conservative.

\paragraph{Results.}
The bandit converges under all three judges.  Under unconstrained
routing (Figure~\ref{fig:cross_judge_regret}), Tabula Rasa achieves
final cumulative regret of $51.0$\,[47.7, 54.4] vs.\
$106.2$\,[104.6, 107.8] for Random under R1 (52\% reduction);
$32.8$ vs.\ $71.9$ under GPT-4.1-mini (54\% reduction); and
$37.6$ vs.\ $97.6$ under Claude (61\% reduction).  All curves are
sublinear with decreasing slope, consistent with progressive learning
of the reward structure regardless of evaluator.

Absolute regret scales with inter-model margins: GPT-4.1-mini's
Random baseline ($71.9$) is 32\% lower than R1's ($106.2$), with
Claude intermediate at $97.6$.  Supplementary/R1 regret ratios are
stable across budget regimes
(GPT: $0.64$\,[0.60, 0.68] to $0.89$\,[0.82, 0.97];
Claude: $0.74$\,[0.70, 0.79] to $0.92$\,[0.87, 1.00]),
confirming proportional compression rather than a structural change in
learning dynamics.  Tabula Rasa confidence intervals are wider than
Random's, reflecting higher run-to-run variance from cold-start
exploration, but the paired difference (Random $-$ Tabula Rasa) is
significant for all three judges.  Under budget constraints, the
qualitative pattern is identical: tighter budgets increase regret, and
loose-budget regret approaches the unconstrained level.

\begin{figure*}[t]
    \centering
    \includegraphics[width=0.92\textwidth]{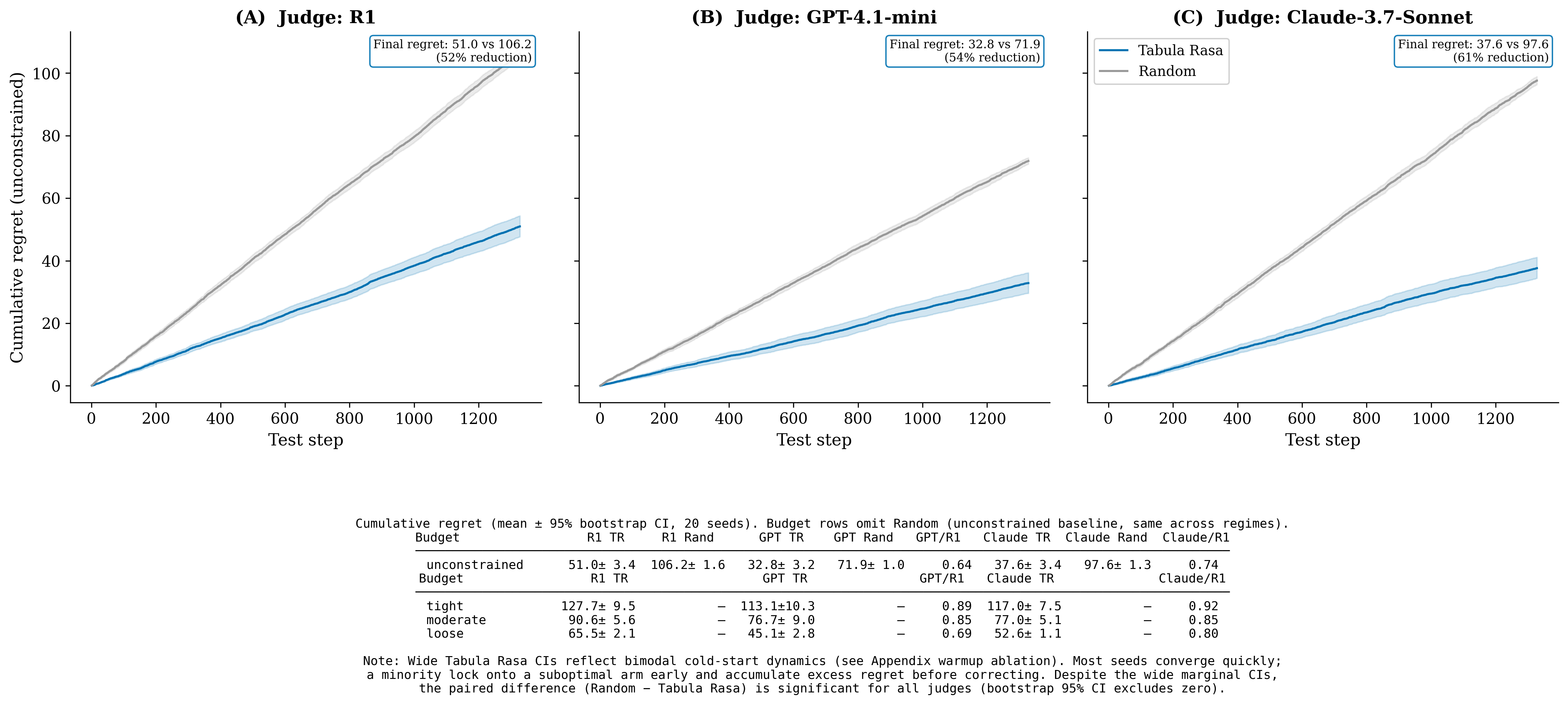}
    \caption{Cold-start bandit regret under three judges (\jrNTest{}
    test prompts drawn from the \jrNSubset{}-prompt judge-robustness sample).
    \textbf{(A)}~R1:
    \jrROneUncReductionPct\% regret reduction.
    \textbf{(B)}~GPT-4.1-mini: \jrGPTUncReductionPct\% reduction.
    \textbf{(C)}~Claude-3.7-Sonnet: \jrClaudeUncReductionPct\% reduction.
    Shaded bands show 95\% bootstrap CIs over \jrNseeds{} seeds.
    Tabula~Rasa bands are wider than the Random baseline's due to higher
    per-seed variance from cold-start exploration.  The paired difference
    (Random~$-$~Tabula~Rasa) is significant for all three judges (bootstrap
    95\% CI excludes zero).  The summary table reports Tabula~Rasa
    cumulative regret (mean $\pm$ 95\% bootstrap CI) across all four budget
    regimes with the supplementary/R1 ratio.}
    \label{fig:cross_judge_regret}
\end{figure*}

\subsection{Summary}

Across three independent judges, the expected reward ordering is
identical (Gemini-Pro $>$ Mistral-Large $>$ Llama-8B;
Table~\ref{tab:expected_ordering}), R1's oracle captures
${\geq}97.4\%$ of other judges' oracle reward
(Table~\ref{tab:cross_regret}), and the bandit's learning dynamics
replicate under both supplementary judges, with regret reductions of
52\%/54\%/61\% (R1/GPT/Claude) under unconstrained routing
(Figure~\ref{fig:cross_judge_regret}).  R1 is therefore well-justified
as the sole evaluator: it provides the largest inter-model margins and
the lowest per-evaluation cost.  Correlating LLM-as-judge scores with
human preferences remains important future work, but it is orthogonal
to the judge-consistency and routing-robustness validated here.


\section{Routing Latency Microbenchmark}
\label{sec:latency_benchmark}

We profile ParetoBandit's per-request routing overhead to validate the
computational-efficiency claims in Section~\ref{sec:efficiency}.
Eight configurations isolate three factors: Sherman--Morrison (SM) vs.\
full inversion (at the same abstraction level), production overhead
(locks, budget pacing, forgetting), and PCA dimensionality
($d{=}26$ vs.\ $d{=}385$).
All configurations use $K{=}3$ arms, synthetic whitened context
vectors, and 4{,}500 measured route\,+\,update cycles (500-round
warmup excluded).
Contexts are drawn i.i.d.\ from a standard normal,
$\ell_2$-normalised, and augmented with a bias term, matching the
dimensionality and unit-norm structure of real PCA-whitened embeddings.
Synthetic inputs are appropriate here because \texttt{route()} and
\texttt{update()} cost depends only on $d$ and $K$, not on semantic
content; using synthetic vectors removes dataset I/O, encoder
inference, and PCA from the timing loop.
The full end-to-end pipeline including embedding and PCA is profiled
separately below (Table~\ref{tab:e2e_latency_breakdown}).

\begin{table}[h]
\centering
\caption{Per-request routing latency ($\mu$s) and throughput ($K{=}3$ arms, 4{,}500 measured cycles).
Configuration groups are described in the text.}
\label{tab:latency}
\small
\resizebox{\columnwidth}{!}{%
\begin{tabular}{@{}l r r r r r@{}}
\toprule
 & \multicolumn{2}{c}{\textbf{Route ($\mu$s)}}
 & \multicolumn{2}{c}{\textbf{Update ($\mu$s)}}
 & \textbf{Throughput} \\
\cmidrule(lr){2-3}\cmidrule(lr){4-5}
\textbf{Configuration}
 & p50 & p95 & p50 & p95 & (req/s) \\
\midrule
\multicolumn{6}{@{}l}{\emph{Production (full router with locks, pacing, forgetting)}} \\
ParetoBandit ($d{=}26$)        & \latPBdTwentySixRouteMedian & \latPBdTwentySixRoutePNineFive & \latPBdTwentySixUpdateMedian & \latPBdTwentySixUpdatePNineFive & \latPBdTwentySixThroughput \\
ParetoBandit ($d{=}385$)       & \latPBdFullRouteMedian & \latPBdFullRoutePNineFive & \latPBdFullUpdateMedian & \latPBdFullUpdatePNineFive & \latPBdFullThroughput \\
\addlinespace
\multicolumn{6}{@{}l}{\emph{Algorithmic isolation (identical \texttt{route()}, only \texttt{update()} differs)}} \\
Bare SM ($d{=}26$)             & \latSMdTwentySixRouteMedian & \latSMdTwentySixRoutePNineFive & \latSMdTwentySixUpdateMedian & \latSMdTwentySixUpdatePNineFive & \latSMdTwentySixThroughput \\
Bare SM ($d{=}385$)            & \latSMdFullRouteMedian & \latSMdFullRoutePNineFive & \latSMdFullUpdateMedian & \latSMdFullUpdatePNineFive & \latSMdFullThroughput \\
Cached Inv.\ ($d{=}26$)       & \latCachedDTwentySixRouteMedian & \latCachedDTwentySixRoutePNineFive & \latCachedDTwentySixUpdateMedian & \latCachedDTwentySixUpdatePNineFive & \latCachedDTwentySixThroughput \\
Cached Inv.\ ($d{=}385$)      & \latCachedDFullRouteMedian & \latCachedDFullRoutePNineFive & \latCachedDFullUpdateMedian & \latCachedDFullUpdatePNineFive & \latCachedDFullThroughput \\
\addlinespace
\multicolumn{6}{@{}l}{\emph{Worst-case baseline (never caches $A^{-1}$)}} \\
Per-Route Inv.\ ($d{=}26$)    & \latPerRouteDTwentySixRouteMedian & \latPerRouteDTwentySixRoutePNineFive & \latPerRouteDTwentySixUpdateMedian & \latPerRouteDTwentySixUpdatePNineFive & \latPerRouteDTwentySixThroughput \\
Per-Route Inv.\ ($d{=}385$)   & \latPerRouteDFullRouteMedian & \latPerRouteDFullRoutePNineFive & \latPerRouteDFullUpdateMedian & \latPerRouteDFullUpdatePNineFive & \latPerRouteDFullThroughput \\
\bottomrule
\end{tabular}
}
\end{table}

Table~\ref{tab:latency} reports per-request latency and throughput for
all configurations (p50, p95, 4{,}500 cycles).  Bare~SM and
Cached~Inv.\ share the same \texttt{route()} code path, isolating the
$O(d^2)$ SM vs.\ $O(d^3)$ full-inversion effect in \texttt{update()}
only, while ParetoBandit adds production features; Per-Route~Inv.\
never caches $A^{-1}$ and serves as a worst-case baseline.
Figure~\ref{fig:latency_benchmark} visualises these comparisons on a
log scale.

\begin{figure*}[t]
\centering
\includegraphics[width=\textwidth]{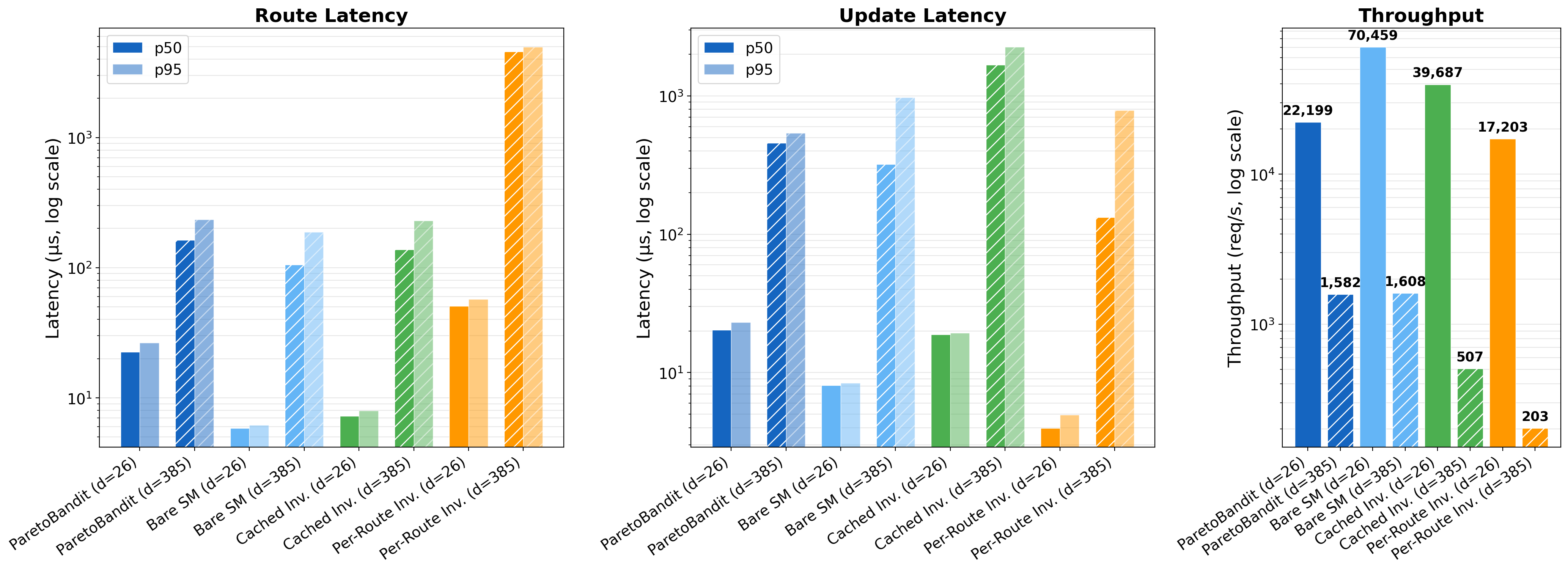}
\caption{Route latency (left), update latency (centre), and throughput
(right) for eight router configurations on a log scale.
Bare~SM and Cached~Inv.\ share identical \texttt{route()} code, so the
route panel confirms they are equivalent; the update panel exposes
Sherman-Morrison's actual contribution ($O(d^2)$ vs.\ $O(d^3)$).
ParetoBandit bars quantify the additional cost of production features.}
\label{fig:latency_benchmark}
\end{figure*}

\paragraph{Algorithmic isolation: SM vs.\ full inversion.}
Bare~SM is a stripped-down LinUCB that applies only the $O(d^2)$
Sherman--Morrison rank-1 update, with no locks, pacing, or forgetting.
Cached~Inv.\ is identical except that \texttt{update()} calls
$O(d^3)$ \texttt{np.linalg.inv} instead of the SM correction; both
inherit from the same Python base class and execute literally the same
\texttt{route()} method (UCB scoring via cached $A_a^{-1}$ and
$\hat\theta_a$), so any route-latency difference is measurement noise.

Route latency is near-identical between Bare~SM and Cached~Inv.\ at
each dimension ($\latSMdTwentySixRouteMedian$ vs.\
$\latCachedDTwentySixRouteMedian\,\mu$s at $d{=}26$;
$\latSMdFullRouteMedian$ vs.\ $\latCachedDFullRouteMedian\,\mu$s at
$d{=}385$), consistent with their shared \texttt{route()} code path,
with the small residual gap likely reflecting Python object-layout
differences between the two subclasses.
Update latency is where SM helps: at $d{=}385$, the $O(d^2)$ SM
update is $5.0{\times}$ faster than full inversion
($346.6\,\mu$s vs.\ $1735.8\,\mu$s, p50); at $d{=}26$, the gap
narrows to $2.3{\times}$ ($8.1\,\mu$s vs.\ $18.6\,\mu$s) because
$d^3 = 17{,}576$ is already fast for LAPACK.
Per-Route~Inv.\ never caches $A_a^{-1}$, paying $K$ full inversions
on every \texttt{route()} call; no reasonable implementation would do
this, but it bounds worst-case behaviour.
\vspace{-1ex}
\paragraph{Production overhead.}
ParetoBandit is the full production router.  It uses the same SM
update as Bare~SM but adds a threading lock around
\texttt{select\_arm()} and \texttt{update()}, budget pacing,
forgetting-factor decay, staleness tracking, and constraint filtering.
Comparing ParetoBandit with Bare~SM isolates the cost of these
production features.

\paragraph{Key findings.}

\textbf{Production overhead is negligible.}
At the production dimension ($d{=}26$), ParetoBandit completes a full
route\,+\,update cycle in $43.0\,\mu$s (p50), adding $<$\,0.05\%
latency even to the fastest LLM call (${\sim}100$\,ms
time-to-first-token for Llama-3.1-8B).
The router sustains ${\approx}22{,}000$\,req/s on a single CPU
core, sufficient for all but the most extreme throughput tiers.

\textbf{SM's benefit is in update cost, not routing.}
Bare~SM and Cached~Inv.\ have near-identical route latency; the improvement
appears in update: at $d{=}385$, SM is $5.0{\times}$ faster
($346.6\,\mu$s vs.\ $1735.8\,\mu$s, p50), and $2.3{\times}$ faster
at $d{=}26$.

\textbf{Production features add bounded overhead.}
At $d{=}26$, ParetoBandit's route is $3.9{\times}$ and update
$2.5{\times}$ slower than Bare~SM due to locking and pacing; at
$d{=}385$, these ratios shrink ($1.7{\times}$ route,
$1.4{\times}$ update) because linear algebra dominates fixed overhead.

\textbf{PCA yields a ${\approx}14.8{\times}$ throughput gain.}
Within production configs, reducing $d$ from 385 to 26 improves
throughput from 1{,}487 to 21{,}983\,req/s (${\approx}14.8{\times}$),
consistent with $O(d^2)$ scaling plus fixed costs.

\paragraph{End-to-end latency and comparisons.}
Table~\ref{tab:e2e_latency_breakdown} breaks down ParetoBandit's
production CPU pipeline (MiniLM-L6-v2 embedding + PCA +
\texttt{route()}) on a single Apple M-series core (200 iterations,
50-iteration warmup).

\begin{table}[t]
\centering
\caption{End-to-end latency breakdown for ParetoBandit's production
CPU pipeline.  Reported values are p50 and p95 over 200 measured
iterations after a 50-iteration warmup; percentages are computed
relative to the p50 total.}
\label{tab:e2e_latency_breakdown}
\footnotesize
\setlength{\tabcolsep}{4pt}
\resizebox{0.98\columnwidth}{!}{%
\begin{tabular}{@{}l r r r@{}}
\toprule
\textbf{Stage} & \textbf{p50} & \textbf{p95} & \textbf{\% of total} \\
\midrule
Embed (MiniLM-L6-v2, CPU) & $\latEteEmbedMedianMs$\,ms & $\latEteEmbedPNineFiveMs$\,ms & $\latEteEmbedPctOfTotal\%$ \\
PCA + whitening            & $\latEtePcaMedianMs$\,ms & $\latEtePcaPNineFiveMs$\,ms & --- \\
\texttt{route()}           & $\latEteRouteMedianMs$\,ms & $\latEteRoutePNineFiveMs$\,ms & $\latEteRoutePctOfTotal\%$ \\
\midrule
\textbf{Total E2E (CPU)}   & $\mathbf{\latEteTotalMedianMs}$\,\textbf{ms} & $\mathbf{\latEteTotalPNineFiveMs}$\,\textbf{ms} & $100\%$ \\
\bottomrule
\end{tabular}
}
\end{table}

\noindent
The embedding step dominates at $97.9\%$ of total time ($9.6$\,ms
p50), PCA\,+\,whitening adds $0.18$\,ms, and \texttt{route()}
contributes $0.080$\,ms ($0.8\%$ of total), for a $9.8$\,ms
end-to-end latency (p50).
The routing algorithm is not the bottleneck; prompt embedding is, and
every contextual router (e.g., RouteLLM, CSCR, LLM Bandit) faces the
same floor.

Among published routers, PROTEUS~\cite{bhatti2026proteus} reports
$8.7$\,ms p50 per-query latency on an A100 GPU, dominated by a
DeBERTa-v3-small encoder forward pass.
ParetoBandit matches this class of latency on CPU only ($9.8$\,ms),
with the routing decision itself ($22.5\,\mu$s at $d{=}26$)
${\approx}390{\times}$ faster than PROTEUS's per-query forward pass.
Other routing systems---LLM
Bandit~\cite{li2025llmbandit}, BaRP~\cite{wang2025barp},
MixLLM~\cite{wang2025mixllm}, RouteLLM~\cite{ong2024routellm},
HybridLLM~\cite{ding2024hybridllm},
OmniRouter~\cite{mei2025omnirouter},
Lookahead~\cite{sun2025lookahead}---do not report per-decision
latency.

The vLLM Semantic Router~\cite{liu2026vllmsr}, a system-level
domain/safety classifier, reports $9$--$22$\,ms routing on MI300X GPUs
depending on signal type, with embedding-based signals dominating
latency.
CSCR~\cite{shirkavand2025cscr} claims ``microsecond'' $k$-NN lookups
but excludes the encoder forward pass.
Infrastructure-layer proxies such as Portkey
Gateway~\cite{portkey2025} (${<}\,1$\,ms) and Kong AI
Gateway~\cite{konghq2025benchmark} ($24$\,ms p95) provide a lower
bound on non-routing overhead.
Figure~\ref{fig:latency_spectrum_e2e} places these systems and
ParetoBandit on a common log-scale alongside typical LLM inference
latencies, showing that all routing overheads are orders of magnitude
below model serving time.

\begin{figure*}[t]
\centering
\includegraphics[width=0.75\textwidth]{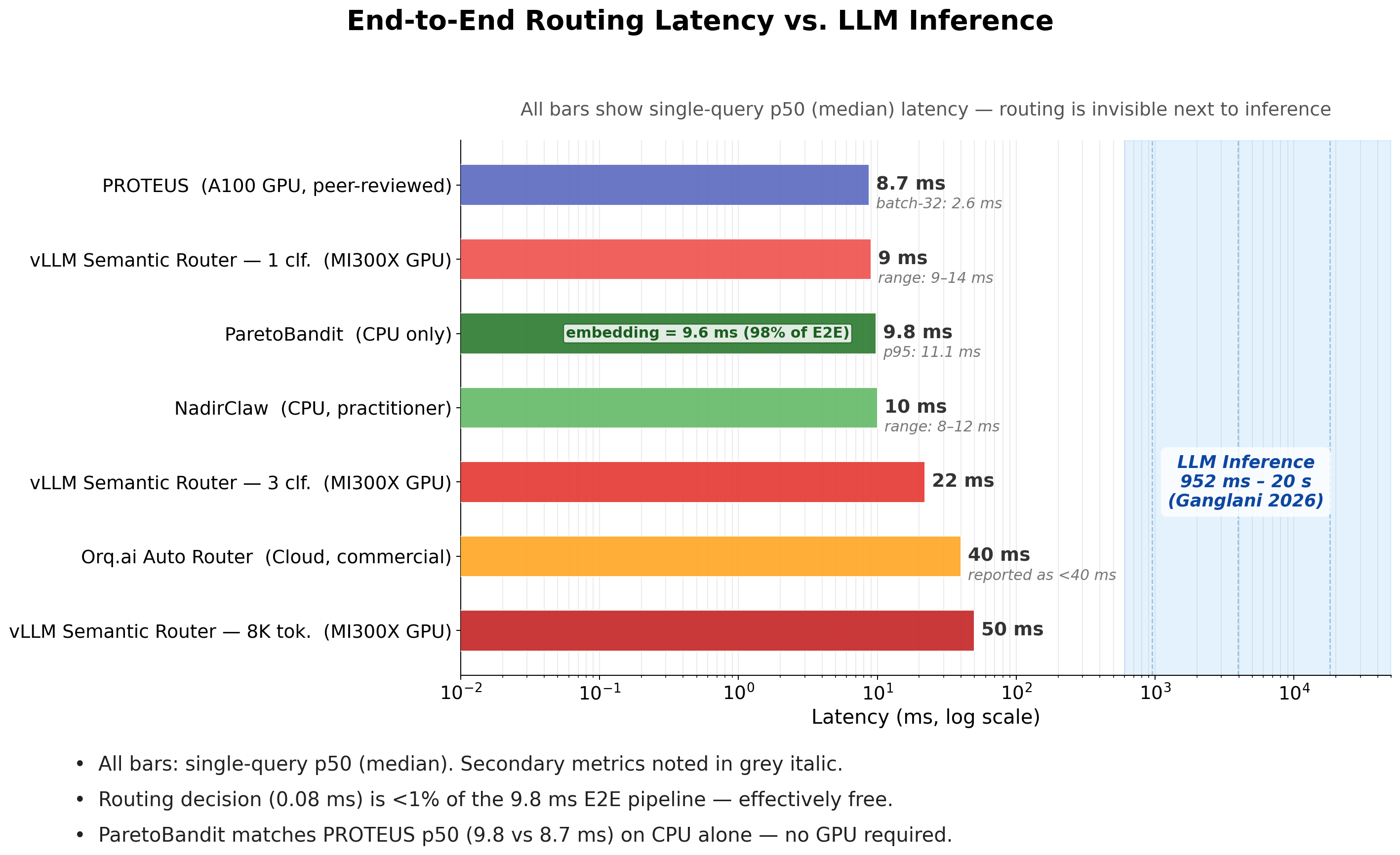}
\caption{End-to-end routing latency (p50, single query) for published
routing systems vs.\ LLM inference time (shaded region).}
\label{fig:latency_spectrum_e2e}
\end{figure*}

\paragraph{Routing overhead relative to LLM inference.}
Table~\ref{tab:overhead_vs_inference} contextualises ParetoBandit's
$9.8$\,ms E2E overhead against the inference latency of the paper's
own $K{=}4$ model portfolio, measured via 20 streaming API calls per
configuration through OpenRouter.

\begin{table}[h]
\centering
\caption{Routing overhead as a fraction of LLM inference latency
($K{=}4$ portfolio, \latInfTrials{} streaming API calls via OpenRouter).
\emph{TTFT} = time to first token.
Routing overhead is the $\latEteTotalMedianMs$\,ms end-to-end CPU
critical path (embedding + PCA + \texttt{route()}).}
\label{tab:overhead_vs_inference}
\small
\resizebox{\columnwidth}{!}{%
\begin{tabular}{@{}l l r r r@{}}
\toprule
\textbf{LLM} & \textbf{Prompt} & \textbf{TTFT (ms)} & \textbf{Total (ms)}
  & \textbf{Routing / Total} \\
\midrule
Llama-3.1-8B        & short  & $\latInfLlamaShortTtft$    & $\latInfLlamaShortTotal$    & $\latInfLlamaShortRoutingPct\%$ \\
Llama-3.1-8B        & medium & $\latInfLlamaMedTtft$      & $\latInfLlamaMedTotal$      & $\latInfLlamaMedRoutingPct\%$ \\
Llama-3.1-8B        & long   & $\latInfLlamaLongTtft$     & $\latInfLlamaLongTotal$     & $\latInfLlamaLongRoutingPct\%$ \\
\addlinespace
Mistral-Large-2512  & short  & $\latInfMistralShortTtft$  & $\latInfMistralShortTotal$  & $\latInfMistralShortRoutingPct\%$ \\
Mistral-Large-2512  & medium & $\latInfMistralMedTtft$    & $\latInfMistralMedTotal$    & $\latInfMistralMedRoutingPct\%$ \\
Mistral-Large-2512  & long   & $\latInfMistralLongTtft$   & $\latInfMistralLongTotal$   & $\latInfMistralLongRoutingPct\%$ \\
\addlinespace
Gemini 2.5 Flash    & short  & $\latInfGemFlashShortTtft$ & $\latInfGemFlashShortTotal$ & $\latInfGemFlashShortRoutingPct\%$ \\
Gemini 2.5 Flash    & medium & $\latInfGemFlashMedTtft$   & $\latInfGemFlashMedTotal$   & $\latInfGemFlashMedRoutingPct\%$ \\
Gemini 2.5 Flash    & long   & $\latInfGemFlashLongTtft$  & $\latInfGemFlashLongTotal$  & $\latInfGemFlashLongRoutingPct\%$ \\
\addlinespace
Gemini 2.5 Pro      & short  & $\latInfGemProShortTtft$   & $\latInfGemProShortTotal$   & $\latInfGemProShortRoutingPct\%$ \\
Gemini 2.5 Pro      & medium & $\latInfGemProMedTtft$     & $\latInfGemProMedTotal$     & $\latInfGemProMedRoutingPct\%$ \\
Gemini 2.5 Pro      & long   & $\latInfGemProLongTtft$    & $\latInfGemProLongTotal$    & $\latInfGemProLongRoutingPct\%$ \\
\bottomrule
\end{tabular}
}
\end{table}

\noindent
Even against the fastest model and shortest prompt (Gemini 2.5 Flash,
$2{,}574$\,ms total latency), the full routing pipeline adds
${\approx}0.38\%$ overhead; for the slowest model (Gemini 2.5 Pro,
long prompt, $8{,}638$\,ms), routing contributes
${\approx}0.11\%$.
The routing decision is effectively invisible to the end user.

\section{Recovery Limit under Quality Degradation}
\label{appendix:recovery_limit}

Experiment~03 (\S\ref{sec:eval_catastrophic}) showed that the system
recovers from a silent quality regression (Mistral-Large reward
dropping to $0.75$) within a 608-prompt Phase~3 horizon.  Here we
extend that result across a range of severities.  The system trends
toward recovery at all tested levels---including near-total
degradation---but deeper corruption requires more Phase~3 prompts
under normal conditions to fully converge, and our evaluation horizon
is too short to observe complete recovery at the most extreme levels.
We therefore characterise a finite-horizon recovery envelope: for a
given Phase~3 length, the degradation severity up to which the system
reaches near-Phase~1 performance, and how extending the horizon shifts
that boundary.

\vspace{-0.5ex}
\paragraph{Setup.}
We sweep Mistral-Large's degraded reward from $0.05$ to $0.85$
(normal reward ${\approx}0.92$).  Degradation severity, the x-axis in
Figure~\ref{fig:recovery_limit}a, is the fractional gap between the
degraded reward and the Phase~1 system baseline (${\approx}0.89$),
ranging from ${\approx}4$\% (mild) to ${\approx}94$\% (near-total).
Degradation is modelled as a mean shift: per-prompt rewards are
shifted so the degraded arm's mean equals the target level while
retaining prompt-dependent variation (clipped to $[0,1]$), which is
more realistic than constant-reward replacement.  Cost is held fixed
during Phase~2 so only the reward signal reveals the regression.  All
runs use the moderate budget
(\rlBudgetTarget{} per prompt), 20~seeds,
and the same hyperparameters as the main experiment.  We define full
recovery as a Phase~3/Phase~1 reward ratio ${\geq}97$\%.

For the extended-horizon evaluation, Phase~3 uses all non-Phase-2
holdout prompts (1{,}216 prompts, $2{\times}$ the base horizon)
without recycling, so the i.i.d.\ arrival assumption is preserved and
extended-horizon results reflect recovery on fresh prompts.

\paragraph{Recovery mechanisms.}
Recovery operates through two mechanisms.  Geometric forgetting
($\gamma{<}1$) exponentially down-weights corrupted Phase-2
observations so that fresh Phase-3 data dominates the mean estimate;
this is the primary driver.  Staleness-driven exploration inflates UCB
variance for non-selected arms and can trigger re-exploration, but
under budget-constrained routing the cost penalty can exceed the
exploration bonus, limiting this effect on its own.  Both mechanisms
act jointly, but the forgetting half-life (${\approx}231$ steps at
$\gamma{=}0.997$) sets the characteristic timescale.

\paragraph{Finite-horizon recovery envelope (Figure~\ref{fig:recovery_limit}a).}
Figure~\ref{fig:recovery_limit}a plots the P3/P1 reward ratio against
degradation severity for two Phase~3 lengths.  At the 608-prompt
horizon (solid blue), full recovery (${\geq}97$\%) is achieved for
degradations up to ${\approx}17$\%, after which the ratio declines
smoothly to a floor of ${\approx}90$\%.  Doubling Phase~3 to 1{,}216
unique prompts (dashed orange) uniformly lifts the curve: the 97\%
boundary shifts to ${\approx}30$\% degradation, and the
severe-degradation floor rises to ${\approx}93$\%.

This uniform lift is consistent with the floor being a finite-horizon
artefact rather than a fundamental limit on recoverable severity:
geometric forgetting continues erasing Phase-2 corruption, so more
Phase-3 prompts yield more recovery.  Within any fixed horizon,
severity beyond ${\approx}50$\% has diminishing marginal impact
because the forgetting dynamics are unchanged; only the initial bias
differs.  At the other extreme, mild degradation (${\leq}10$\%) can
produce P3/P1 ratios above $100$\%: Phase~2 drives the policy away
from the degraded arm, and for some prompts the alternative routing
discovered during degradation is genuinely better than the original
policy.

\vspace{0.5ex}
\begin{figure*}[!t]
\centering
\includegraphics[width=0.88\textwidth]{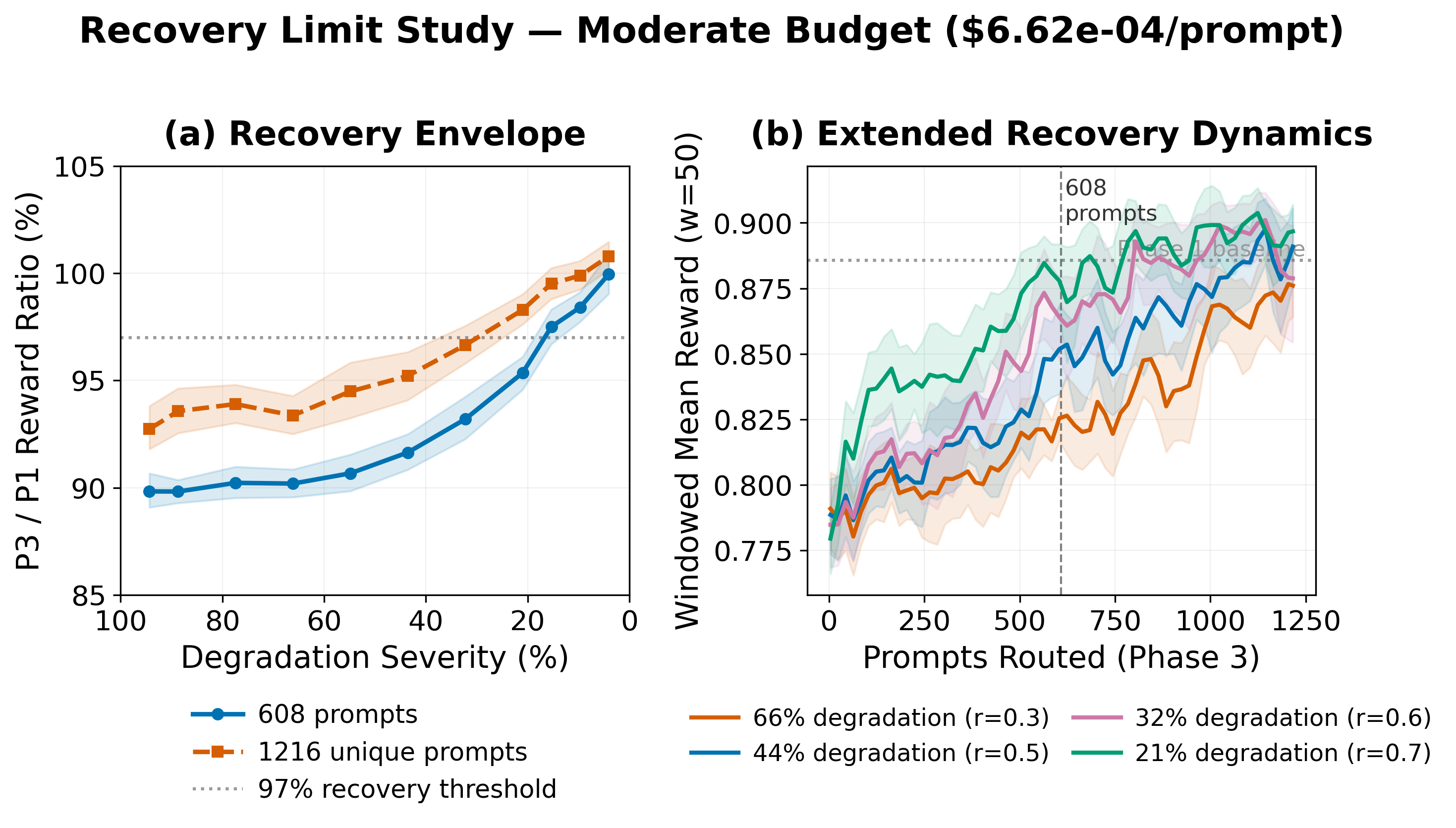}
\caption{Recovery limit under quality degradation (moderate budget,
\rlNseeds~seeds, 95\% bootstrap CI).
\textbf{(a)}~P3/P1 reward ratio vs.\ degradation severity; dotted
line = 97\% full-recovery threshold.
\textbf{(b)}~Phase-3 recovery dynamics (50-prompt rolling window) at
the extended horizon for four severity levels.}
\label{fig:recovery_limit}
\end{figure*}

\paragraph{Convergence dynamics (Figure~\ref{fig:recovery_limit}b).}
The extended-horizon time series (50-prompt rolling window) confirm
this interpretation.  All trajectories rise monotonically toward the
Phase~1 baseline with no sign of plateauing, but convergence speed is
severity-dependent: milder degradations reach baseline well before
1{,}216 prompts, while more severe degradations are still improving at
the horizon.  In all cases the confidence bands narrow over time,
indicating stable convergence rather than high-variance oscillation.
The monotonic, non-plateauing trajectories are consistent with
recovery being rate-limited by Phase-3 length rather than
fundamentally bounded by degradation severity, though we cannot
confirm full convergence at the most extreme levels within our
evaluation budget.

\paragraph{Limitations and deployment implications.}
This study evaluates a single budget level and a single-arm,
quality-only degradation; other budgets or simultaneous multi-arm
regressions may behave differently.  For typical silent regressions
(${\leq}20$\%), ParetoBandit recovers automatically within the
608-prompt observation window.  For more severe regressions, recovery
still proceeds but requires a longer Phase~3 horizon; operators can
either allow additional routing time or intervene with an explicit
model demotion to accelerate convergence.

\endgroup

\end{document}